# An Agent Architecture with Tools and Memory for Selective Correction of Wrist PPG Heart Rate Estimates

Xu Wang[1,2], Shuhan Yi[2], and Zhen Cheng[2*]

[1] School of Engineering, The Hong Kong University of Science and Technology, Hong Kong 999077, China

[2] School of Biomedical Engineering, Tsinghua University, Beijing 100084, China

*Corresponding author:

Zhen Cheng, School of Biomedical Engineering, Tsinghua Medicine, Tsinghua University, Beijing 100084, P. R. China. Email: zcheng@mail.tsinghua.edu.cn

# Abstract

Wrist photoplethysmography (PPG) heart rate estimates are vulnerable to motion artifacts, and corrections that reduce mean error can worsen individual windows. We propose an agent architecture with tools and memory around a fixed heart rate estimator. Pulse, motion, and limited historical evidence generate candidate estimates, whose predicted benefits and harms guide selection. An independent risk gate determines whether to retain, correct, or reject each output. We trained the architecture on PPG-DaLiA and tested it on Wrist Exercise, GalaxyPPG, and WildPPG with frozen weights. Across these four public datasets, mean absolute error (MAE), macro-averaged over ten estimators, decreased by up to 44.5%, with the largest reduction on GalaxyPPG. Beneficial and harmful corrections decreased and increased absolute error by more than 1 bpm, respectively. Their ten-estimator macro-averaged proportions ranged from 16.06% to 36.64% and from 1.68% to 6.77%, respectively, across the four datasets. At 60% and 80% target coverage, numerical correction on the same accepted windows contributed more to total error reduction than window selection. We implemented real-time wearable acquisition, Bluetooth reception, and smartphone deployment of a selected estimator and the framework. In backend analysis of four prototype recordings from one participant, ten-estimator macro-averaged MAE decreased from 9.09 to 4.45 bpm without additional rejection. Under the evaluated conditions, the architecture improved average accuracy while quantifying correction benefits and harms, supporting wearable heart rate analysis that considers error, correction risk, and coverage.

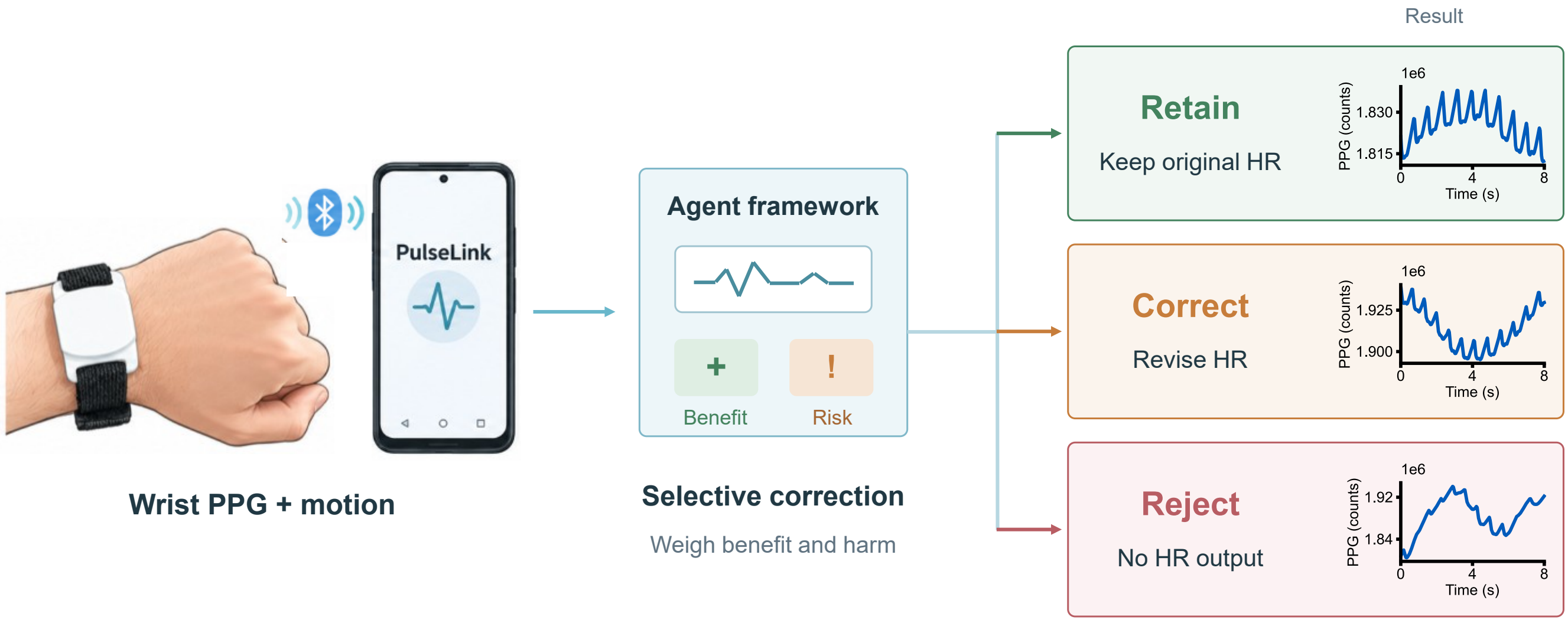

**Abstract figure**

## I. INTRODUCTION

Wearable heart rate monitoring, predominantly enabled by photoplethysmography (PPG), has become central to wearable health sensing, enabling continuous characterization of daily activity and physiological status in real-world settings. PPG measures rhythmic changes in peripheral blood volume using a compact, noninvasive optical sensor [1], [2]. Motion can introduce interference that overlaps with pulse components and reduces estimation accuracy [1], [3]. This interference varies with activity and wearing conditions, and substantial errors can remain after artifact suppression. Continuous monitoring therefore requires both accurate estimates and a way to identify unreliable outputs before they enter subsequent analyses.

Signal processing and data-driven approaches have been developed to address motion interference. TROIKA [4], JOSS [5], SpaMA [6], and WFPV [7] use signal decomposition, joint spectral reconstruction, analysis of motion-related frequencies, and filtering or tracking to improve heart rate estimation under interference. Deep PPG [3] and TimePPG [8] learn relationships between sensor signals and heart rate using time–frequency representations and temporal convolution. Output reliability has also received attention. Signal quality indices can identify contaminated segments; BeliefPPG models heart rate dynamics and probabilistic inference to provide estimates and their uncertainty [9], whereas KID-PPG combines knowledge-guided inference with identification of unreliable outputs [10]. Beyond improving estimates and detecting unreliable results, post-calibration methods use device outputs, activity, and historical information to adjust existing heart rate estimates [11]. Such additional information can support correction, but it also raises the question of which estimates should be changed and which should be retained.

These methods do not fully resolve when an existing estimate should be changed. Mean error measures overall accuracy but does not establish whether a correction benefits an individual window. BeliefPPG and some post-calibration methods assess window-level reliability or adjust unreliable outputs [9], [11]. However, explicit assessment of correction gains and harm to originally accurate estimates remains limited. Average improvements can conceal deterioration in individual windows. If a system also rejects outputs, lower error may partly reflect the exclusion of difficult windows [12], [13]. Evaluation must therefore distinguish correction benefits, correction harms, and output coverage.

Here, we propose a selective correction framework built around a fixed heart rate estimator. Its organization links observations, internal states, and action selection, following an agent architecture [14]. Signal evidence and recent history inform whether to retain, correct, or reject an estimate without changing the underlying estimator. Decisions account for predicted gains and harm relative to the original estimate, including deterioration of initially accurate outputs. We evaluate numerical correction separately from window rejection to identify the source of error reductions.

This study makes three contributions. First, we introduce a framework with signal analysis tools and memory that evaluates candidate gains and harms before replacing a fixed estimator's output. A shared base-error prediction and geometric constraints give candidates a consistent comparison against the original estimate. The original estimate remains available, and candidate selection precedes a separate output risk gate. Second, we evaluate ten prespecified estimators on four public datasets and prototype recordings, separating correction gains from selection effects and reporting correction harm and coverage. Third, we implement real-time wearable acquisition and smartphone deployment of a selected estimator and the framework. The smartphone receives and analyzes the signals, while the computer backend supports comparisons across methods.

## II. MATERIALS AND METHODS

### *A. Task Definition and Overall Architecture*

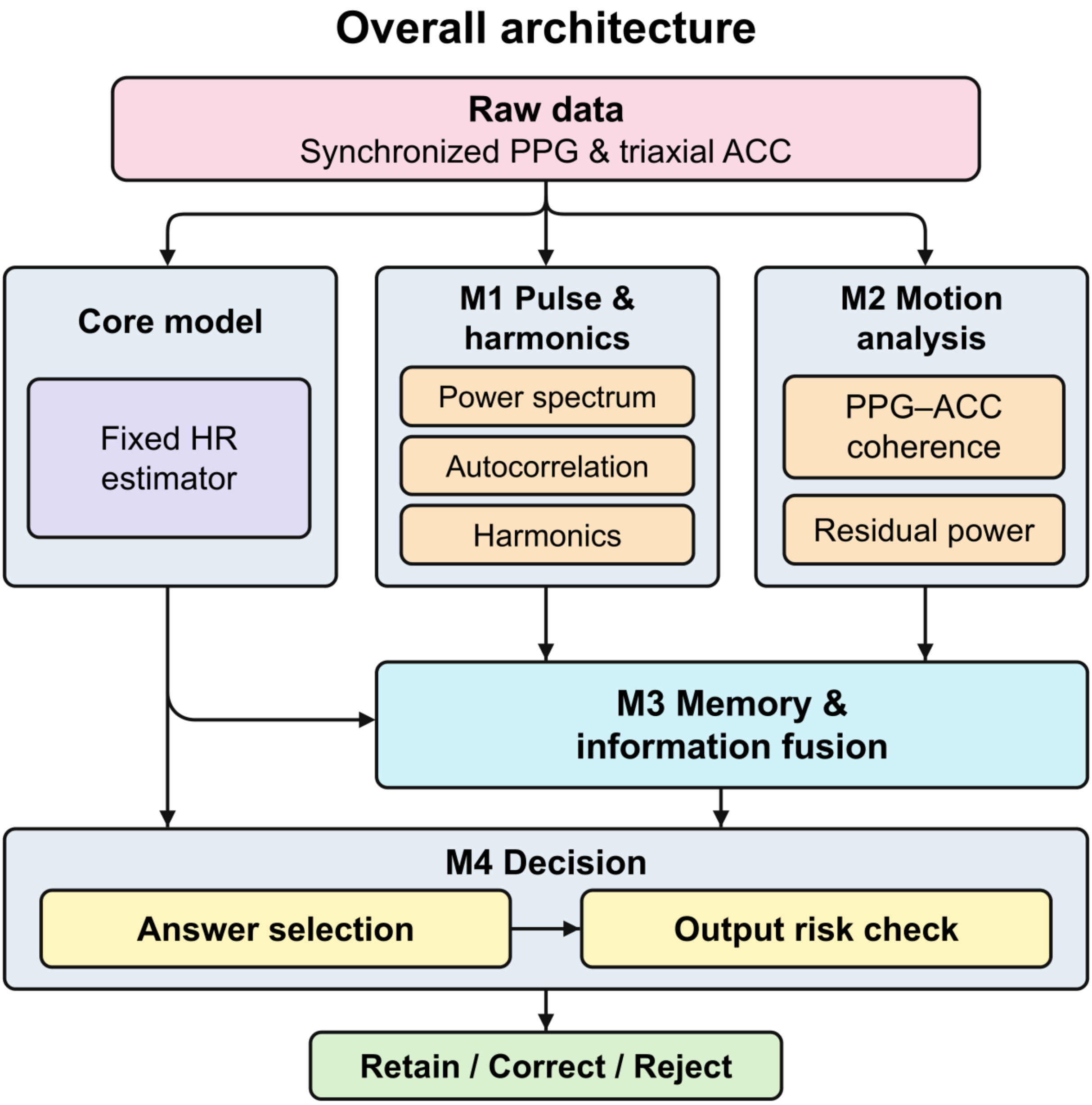

Fig. 1. Agent architecture with tools and memory for selective heart rate correction. Information flow among the fixed core estimator and M1–M4, with retain, correct, and reject outputs. M1/M2 extract pulse, harmonic, and motion evidence from PPG and triaxial ACC to form a 186 × 30 evidence tensor. M3 integrates evidence from five temporal positions (the current position and four preceding positions, spaced 4 s apart) into memory and combines it with current evidence and the core estimator prior to generate a heart rate posterior. M4 evaluates gain, harm, and error from 106-dimensional features for nine candidates, selects an estimate, and then checks output risk. Arrows indicate information flow. PPG: photoplethysmography; ACC: acceleration; HR: heart rate.

As illustrated in Fig. 1, the framework comprises four modules, including M1 and M2 which extract pulse and motion evidence, respectively. M3 combines this evidence with recent history and the core estimator prior to generate heart rate candidates. M4 predicts each candidate’s gain and harm relative to the original estimate, selects an estimate, and applies a separate risk gate. The architecture is shared across core estimators, but M3 and M4 are fitted independently for each estimator. The evaluated framework therefore uses estimator-specific auxiliary weights rather than a single set of weights shared by all ten estimators.

The framework corrects the window-level outputs of a fixed heart rate estimator. For analysis window $t$, the raw sensor input is a synchronized $N \times 4$ matrix comprising $N$ samples from one PPG channel and three acceleration (ACC) channels. In addition, the framework receives three estimator-related quantities: the original heart rate estimate $h_{0,t}$, an adapted heart rate prior $p_t$, and a baseline risk score $r_{0,t}$. These are window-level quantities rather than additional sensor channels. The baseline risk score is an estimator-specific uncertainty indicator associated with the original estimate. For BeliefPPG and KID-PPG, it is the standard deviation of the predictive distribution; for the other estimators, it is an adapter-derived entropy measure of spectral evidence. Larger scores indicate greater uncertainty. The score is computed without reference heart rate at inference and is not the observed absolute error. Estimator-specific definitions are provided in Supplementary Methods S6. The output is a retained estimate, a corrected estimate, or a missing-value indicator. Candidate heart rates are denoted by $a_{t,j}$, with $j = 0$ corresponding to the original estimate, i.e., $a_{t,0} = h_{0,t}$. Given the reference heart rate $y_t$, candidate absolute error, net gain relative to the original estimate, and added error are defined as follows:

$$\begin{aligned} e_{t,j} &= |a_{t,j} - y_t|, \\ G^*_{t,j} &= e_{t,0} - e_{t,j}, \\ H^*_{t,j} &= max(-G^*_{t,j}, 0). \end{aligned} \tag{1}$$

Here, $e_{t,j}$ is the candidate absolute error, $G^*_{t,j}$ is the net gain, and $H^*_{t,j}$ is the added error, all measured in beats per minute (bpm). Positive net gain indicates error reduction, whereas negative net gain indicates error increase. The original estimate always remains in the candidate set and has zero relative net gain and added error. Reference heart rate is used only for source-domain supervision, calibration, and offline evaluation; it is not used for candidate generation or action selection at test time.

M1 and M2 are deterministic signal processing modules; M3 and M4 are learned modules. The fixed workflow links observation analysis, internal-state updates, and action selection [14]. It does not use a language model, dynamically schedule tools, or combine votes from multiple core estimators. Fig. 1 summarizes evidence extraction, memory fusion, candidate selection, and output risk checking. Supplementary Methods S18 specifies the module interfaces.

*B. Pulse and Motion Evidence Extraction*

M1 and M2 transform raw signals into evidence on a common heart rate grid to compare pulse and motion-related components. Analysis windows are 8 s long and updated every 4 s. At each update, M1 and M2 use both the current 8 s signal segment and a continuous 24 s signal segment ending at the same time, with a common feature-computation sampling rate of 64 Hz. The 24 s segment spans five consecutive, overlapping 8 s windows with a 4 s stride (8 + 4 × 4 = 24 s). It provides a longer raw-signal context for evidence extraction and is distinct from the five-position feature memory used by M3. The heart rate grid spans 35–220 bpm in 1 bpm increments, giving 186 positions. Each position contains 30 features, forming the 186 × 30 evidence matrix $X_t$. M1 extracts PPG power, autocorrelation, and spectral features at the fundamental, half, and double frequencies. M2 characterizes motion coupling using magnitude-squared coherence between PPG and triaxial ACC and computes residual power that downweights motion-related frequencies.

Pulse and motion frequencies can overlap, so the representation retains the original spectrum, coherence, and residual power. Its 30 channels contain spectra and autocorrelation at two time scales, together with interaction features at the fundamental, half, and double frequencies. Supplementary Methods S11 and S12 define spectral estimation, coherence, and residual power; Table S1 lists the channel order.

*C. Limited-History Fusion and Heart Rate Candidate Generation*

M3 uses recent evidence to constrain the heart rate distribution of the current window. Its input comprises evidence matrices from five temporal positions: the current position and four preceding positions, spaced 4 s apart. Each feature channel c has a learnable decay coefficient $\alpha_c$, obtained by applying a sigmoid to parameter $d_c$ and shared across all 186 heart rate grid positions. The state is initialized using the evidence matrix at the earliest position and then updated chronologically:

$$S_t^{(i)} = \alpha \odot S_t^{(i-1)} + (1-\alpha) \odot X_{t-4+i}, i = 1, \ldots, 4. \tag{2}$$

In (2), $\odot$ denotes elementwise multiplication, and i indexes updates within the five-position sequence. The final historical state $S_t$ is the result of the fourth update. The state is recomputed over the most recent five positions at every inference step; missing history at the beginning of a recording is padded with the earliest available features. The five endpoints span 16 s, and each position uses a 24 s evidence window, so the earliest signal used precedes the current endpoint by 40 s. No future data are accessed. Initialization and the expanded recurrence are given in Supplementary Methods S13.2.

At heart rate grid point g, current evidence, the historical state, and their difference are concatenated into a 90-dimensional vector. A shared two-layer perceptron with 24 hidden units and sigmoid linear unit (SiLU) activation produces an energy correction $E_t$(g). This correction is bounded between −6 and 6 by a hyperbolic tangent, added to the temperature-scaled log prior, and normalized over the heart rate grid $\mathcal{G}$:

$$q_t(g) = \frac{\exp\{\log p_t(g)/T + E_t(g)\}}{\sum_{u \in \mathcal{G}} \exp\{\log p_t(u)/T + E_t(u)\}}, T = 0.5 + 3\sigma(\theta). \tag{3}$$

Here, T is a learnable temperature parameterized by θ and bounded between 0.5 and 3.5. Prior probabilities are floored at $10^{-10}$ before taking logarithms. $p_t$ is adapted from the core estimator output distribution or corresponding evidence. $q_t$ is the fused heart rate distribution, referred to below as the posterior; this terminology does not imply that its probabilities have been calibrated in the target domain. Full expressions for feature concatenation and the energy network are given in Supplementary Methods S13.3.

Nine candidates are derived from the posterior: the original estimate, mean, mode, five percentiles, and a midpoint candidate. The percentiles are 10th, 25th, 50th, 75th, and 90th. The midpoint between the original estimate and posterior median offers a smaller correction. Candidate order is fixed, and duplicates are allowed. The original estimate occupies the first position and retains its continuous value, even outside the heart rate grid. Supplementary Methods S13.4 defines the candidate set.

*D. Modeling Candidate Gains and Harms*

M4 evaluates candidates generated by M3 by learning how error changes when the original estimate is replaced. It first computes the expected candidate absolute error $e_{t,j}^{q}$ and expected added error $h_{t,j}^{q}$ under posterior $q_t$. These quantities are posterior-weighted averages, not the actual errors against reference heart rate. Their expressions are provided in Supplementary Methods S14.1.

To learn the discrepancy between posterior expectations and actual errors, M4 constructs 106 features for each candidate: 16 candidate and distribution statistics, plus 30-dimensional current evidence, historical evidence, and current-evidence differences relative to the original estimate at the candidate position. Evidence is sampled at the nearest heart rate grid position, while the candidate heart rate itself is unchanged. Features are standardized using source-training statistics and clipped to [−5, 5]. A shared linear branch produces three allocation coefficients for each candidate; a second branch reads only the original-estimate features to produce a nonnegative base error $b_t$ shared within the window. Feature definitions, standardization, and network expressions are given in Supplementary Methods S14.2–S14.4.

The three allocation coefficients $\pi_{t,j}$ partition the magnitude of the candidate change relative to the original estimate; they are not action probabilities for retain, correct, and reject. Joint decoding with the base error yields a beneficial component $\hat{B}_{t,j}$, added error $\hat{H}_{t,j}$, net gain $\hat{G}_{t,j}$, and candidate absolute error $\hat{E}_{t,j}$. Net gain equals the beneficial component minus added error, and candidate absolute error equals base error minus net gain. The complete decoding expressions are given in (S22)–(S25) of Supplementary Methods S14.4.

The shared base error gives all candidates the same comparison reference within a window. Decoding keeps predicted absolute error nonnegative and limits predicted net gain to the magnitude of the candidate change. It also enforces the triangle inequality between the original estimate, candidate estimate, and reference. These constraints make the predictions geometrically consistent but do not guarantee actual error bounds. Supplementary Methods S16 and S17 provide the training objectives and proofs.

*E. Candidate Selection and Risk Control*

Candidate selection precedes output rejection. A replacement candidate must have predicted net gain of at least 1 bpm and predicted added error of at most 2 bpm. Its posterior expected absolute error must also be lower than that of the original estimate. Eligible candidates receive utility equal to net gain minus twice the added error; the original estimate has zero utility:

$$U_{t,0} = 0,$$
$$U_{t,j} = \begin{cases} \hat{G}_{t,j} - 2\hat{H}_{t,j}, & \hat{G}_{t,j} \geq 1,\ \hat{H}_{t,j} \leq 2,\ e^q_{t,0} > e^q_{t,j}, \\ -\infty, & \text{otherwise.} \end{cases} \tag{4}$$

$$j^* = argmax_j U_{t,j}, \hat{h}_t = a_{t,j^*}. \tag{5}$$

$j^*$ is the index of the candidate with maximum utility, and $\hat{h}_t$ is the selected estimate before risk filtering. A tie at zero utility favors the original estimate, which occupies the first candidate position. The proposed method imposes no 20 bpm cap on the magnitude of correction; its utility can also be written as $\hat{B}_{t,j} - 3\hat{H}_{t,j}$. The 1 bpm and 2 bpm thresholds apply to predicted quantities and must not be interpreted as guarantees of actual improvement or harm.

The risk gate then determines whether the selected estimate can be output. Risk score $r_t$ is either the baseline risk $r_{0,t}$ from the core estimator or its adapter (Supplementary Methods S6), or the absolute error $\hat{E}_{t,j*}$ predicted by M4 for the selected candidate. Both the risk type and threshold τ are determined exclusively from source-domain calibration data. A window is accepted only if the selected estimate and risk score are finite and $r_t \leq \tau$:

$$A_t = \mathbf{1}(\hat{h}_t \text{ finite})\mathbf{1}(r_t \text{ finite})\mathbf{1}(r_t \leq \tau). \tag{6}$$

Here, $A_t$ is the acceptance indicator. An accepted window outputs the selected estimate; a rejected window is marked as missing. The complete output expression is given in (S41) of Supplementary Methods S19. Retention and correction describe estimate selection, whereas rejection describes the subsequent output status, so these operations should be evaluated separately. M3 and M4 contain 2,240 and 428 learnable parameters, respectively, totaling 2,668 auxiliary parameters per core estimator.

*F. Data Splits, Model Training, and Source-Domain Calibration*

We used PPG-DaLiA to train and calibrate the auxiliary models and evaluated transfer to Wrist Exercise, Galaxy-PPG, and WildPPG [3], [15]–[17]. These datasets contained 15, 8, 20, and 14 participants with eligible reference measurements, respectively. Table I lists the recording and window counts. Activity recordings were pooled within each participant, and windows without eligible reference measurements were excluded from error evaluation. All datasets used 8 s windows and a 4 s stride. Supplementary Methods S1 specifies reference eligibility.

TABLE I
DATASETS AND ANALYSIS SCOPE

| Dataset | Participants included | Participants with eligible reference | Recordings | Reference-eligible windows |
|---|---|---|---|---|
| PPG-DaLiA | 15 | 15 | 15 | 32308 |
| Wrist Exercise | 8 | 8 | 19 | 1671 |
| GalaxyPPG | 23 | 20 | 23 | 16239 |
| WildPPG | 16 | 14 | 16 | 166132 |

Source-domain participants were divided into disjoint auxiliary-training, calibration, and evaluation groups. M3 first generated out-of-fold posteriors within the training group. M4 was fitted using these posteriors and historical features from the corresponding M3 models. The final M3 was then fitted on the full auxiliary-training group for calibration and evaluation. This procedure prevents M4 from training on M3 predictions fitted to the same participants. Participant assignments, folds, and random seeds are detailed in Supplementary Methods S1 and S5.

M3 represents reference heart rate as a Gaussian soft target with a standard deviation of 3 bpm. Its loss combines cross-entropy and cumulative-distribution discrepancy. M4 uses the true net gain, added error, and candidate absolute error in (1) as supervision. It minimizes SmoothL1 loss with an additional penalty for underestimating added error. The stages are trained sequentially; M4 gradients do not propagate to M3 or the core estimator. Both use participant-balanced sampling and AdamW. M3 uses a learning rate of 0.002, weight decay of 0.03, and batch size of 96. The corresponding M4 settings are 0.003, 0.1, and 256. Each stage runs for 2,000 steps, using the final weights without checkpoint selection on the evaluation set. Supplementary Methods S5 and S16 specify loss weights, initialization, and other settings.

Source-domain calibration was performed separately from training. We selected either original risk or learned error according to macro-averaged mean absolute error (MAE) at 80% coverage in the calibration participants; ties favored original risk. This comparison used uncapped correction without the posterior-consistency condition. The selected risk type was then applied to the corresponding decision configurations. Two thresholds were calibrated separately: one for 80% source coverage and one for source MAE no greater than 5 bpm. The error-based rule additionally required at least 10% coverage for each calibration participant. Thresholds remained fixed during external evaluation. No feasible output was assumed when calibration constraints could not be met. Supplementary Methods S6 describes the search rules.

The ten core estimators were Deep PPG, KID-PPG, TimePPG, BeliefPPG, Welch, ACC-NLMS, TROIKA, JOSS, SpaMA, and WFPV [3]–[10]. Supplementary Methods S6 describes their implementations. Core estimators remained fixed during auxiliary training, with M3 and M4 fitted independently for each estimator. External evaluation froze auxiliary weights, standardization statistics, decision rules, and source thresholds. Target references were not used to select source models or refit auxiliary modules. Each external participant was evaluated with each source configuration, and metrics were averaged without fusing predictions. Supplementary Methods S1 and S5 report the folds, seeds, and training procedure. KID-PPG retained its per-recording adaptation without heart rate labels [10]. Training-participant provenance was incomplete for some pretrained weights, and all four datasets had been used in earlier development. The comparisons therefore assess additional correction gains for the specified implementations, rather than reproduce a ranking of the original methods. Original and corrected errors are compared within each implementation under the same evaluation conditions; headline accuracies from the original publications are not directly comparable across different datasets, splits, and processing protocols.

### *G. Evaluation Metrics and Statistical Analysis*

Evaluation considers estimation error, correction harm, and output coverage. Let $\mathscr{R}$ denote all windows with eligible reference measurements, $\mathscr{A}$ the subset accepted with finite outputs, and $\mathscr{A}_0$ the subset of $\mathscr{A}$ with finite original estimates. The set $\mathscr{A}_0$ supports paired comparisons before and after correction within the same windows. Mean absolute error (MAE), measured in bpm, is the mean absolute difference between estimates and reference heart rates in accepted windows. Coverage is the fraction of all reference-eligible windows that are accepted.

Paired accepted windows are classified by the change in absolute error relative to the original estimate. A decrease exceeding 1 bpm is a beneficial correction; an increase exceeding 1 bpm is a harmful correction. Changes within ±1 bpm form the middle category. The three proportions share the same denominator and sum to 100%. The middle category includes unchanged estimates and small error changes, so it differs from the retention-action rate. The severe added-error proportion counts accepted windows with an error increase exceeding 10 bpm, divided by all reference-eligible windows. MAE and paired correction proportions are undefined when no valid outputs remain. Supplementary Methods S8 and S19 give the definitions. Comparators include original estimates, rejection-only strategies, candidate-selection variants, and ablations of harmonic evidence, motion evidence, or memory.

Both rejection-only comparators preserve the original heart rate values. Ridge rejection uses a ridge regression model trained on source data to predict each window's original estimation error. It ranks test windows by predicted error and retains those with the lowest scores. This comparator measures what a simple error predictor can achieve through selection alone. In risk-ranked comparisons, same-risk rejection uses exactly the framework's risk scores and accepted windows but outputs the original heart rates. Its MAE difference from the full framework therefore isolates numerical correction on the same windows. Neither comparator uses test-time reference errors to select windows. For thresholds targeting source MAE of 5 bpm, each strategy is calibrated independently, so accepted sets can differ. Supplementary Methods S6 describes their features, training, and threshold rules.

Risk–coverage curves are evaluated from 5% to 100% target coverage in steps of 5 percentage points. We selected 80% target coverage as a common operating point to retain most eligible outputs while allowing rejection of a limited fraction of high-risk windows. This choice represents a practical trade-off between output availability and error reduction, rather than an established standard or an optimal coverage level. At the prespecified 80% operating point, error changes are decomposed into numerical correction and window selection. Let $\mathcal{A}$ be the accepted set obtained by ranking the proposed method risk scores and $\mathcal{B}$ the accepted set obtained using the original estimator risk scores. Then

$$\Delta_{\text{total}} = \Delta_{\text{correction}} + \Delta_{\text{selection}}. \quad (7)$$

The total change is MAE($\hat{h}$, $\mathcal{A}$) − MAE($h_0$, $\mathcal{B}$). The correction term is MAE($\hat{h}$, $\mathcal{A}$) − MAE($h_0$, $\mathcal{A}$), comparing estimates on a fixed accepted set. The selection term is MAE($h_0$, $\mathcal{A}$) − MAE($h_0$, $\mathcal{B}$), comparing window sets while keeping original estimates fixed. Fig. 4 reports the negatives of these terms as MAE reductions. This decomposition is a descriptive algebraic comparison. Target-coverage evaluation ranks risk scores within target recordings and retains at most the specified fraction of eligible windows without using reference heart rate values to select estimates. It does not imply that a fixed source-domain threshold maintains the same coverage in the target domain.

Participants were the statistical units for public-dataset evaluation. Source-configuration metrics were first averaged within each participant and estimator. Cross-estimator macro-averages then weighted the ten implementations equally, followed by equal weighting of participants. We calculated 95% percentile confidence intervals from 10,000 participant-bootstrap resamples. Paired effects were calculated before resampling; Supplementary Methods S8 specifies the random seed and aggregation details. These intervals are conditional on the fitted models and exclude retraining variability. Each ablation retrained the affected auxiliary modules on source data and froze their weights for external evaluation. Table S6 reports MAE changes and confidence intervals.

### *H. Wearable Acquisition, Smartphone Deployment, and Prototype Evaluation*

The wearable system uses a Seeed Studio XIAO nRF52840 Sense controller, a MAX30102 optical sensor, and the Android application PulseLink. It acquires red and infrared PPG and triaxial acceleration, then transmits signals and frame counts over Bluetooth Low Energy (BLE). A selected core model and the framework run on the smartphone to analyze received signals, display heart rate, and save recordings. The computer backend compares methods and calculates paired errors, correction harm, and risk–coverage curves. Fig. 5(a)–(c) show the measurement setup, hardware, and smartphone workflow. Supplementary Methods S20.3 specifies the smartphone and backend configurations. The quantitative results reported here were calculated from saved recordings.

One participant wore the wrist device and an independent electrocardiography (ECG) sensor on the upper sternum during static, walking, running, and recovery-rest tests. ECG provided the heart rate reference. Wrist signals were sent to the smartphone for real-time analysis and recording, followed by backend comparison of original and corrected estimates. Supplementary Methods S20 describes timing alignment, preprocessing, and evaluation settings.

## III. RESULTS

### *A. Examples of Retention, Correction, and Rejection*

Fig. 2 illustrates retention, correction, and rejection using GalaxyPPG wrist signals and synchronized ECG, with Deep PPG as the fixed core estimator. All examples use the configuration specified in Supplementary Methods S9. In the retained window, the original and final estimates were both 77.35 bpm, versus a reference of 75.73 bpm. In the corrected window, the estimate changed from 76.53 to 91.00 bpm, closer to the 88.64 bpm reference. Absolute error decreased from 12.10 to 2.36 bpm. In the rejected window, the estimate before risk filtering was 76.81 bpm, versus a reference of 86.19 bpm; the framework withheld the output. Corrections could also cause harm: Table S8 shows an example in which absolute error increased from 1.51 to 4.00 bpm.

The three illustrated windows start at 1412 s for P21, 420 s for P15, and 2508 s for P12. Each column aligns raw PPG, ECG, and window-end heart rates from 8 s before to 16 s after the central window start. Cases were classified retrospectively using reference heart rate; selection rules are specified in Supplementary Methods S9. They illustrate output behavior and cannot be used to estimate the overall frequencies of the three behaviors.

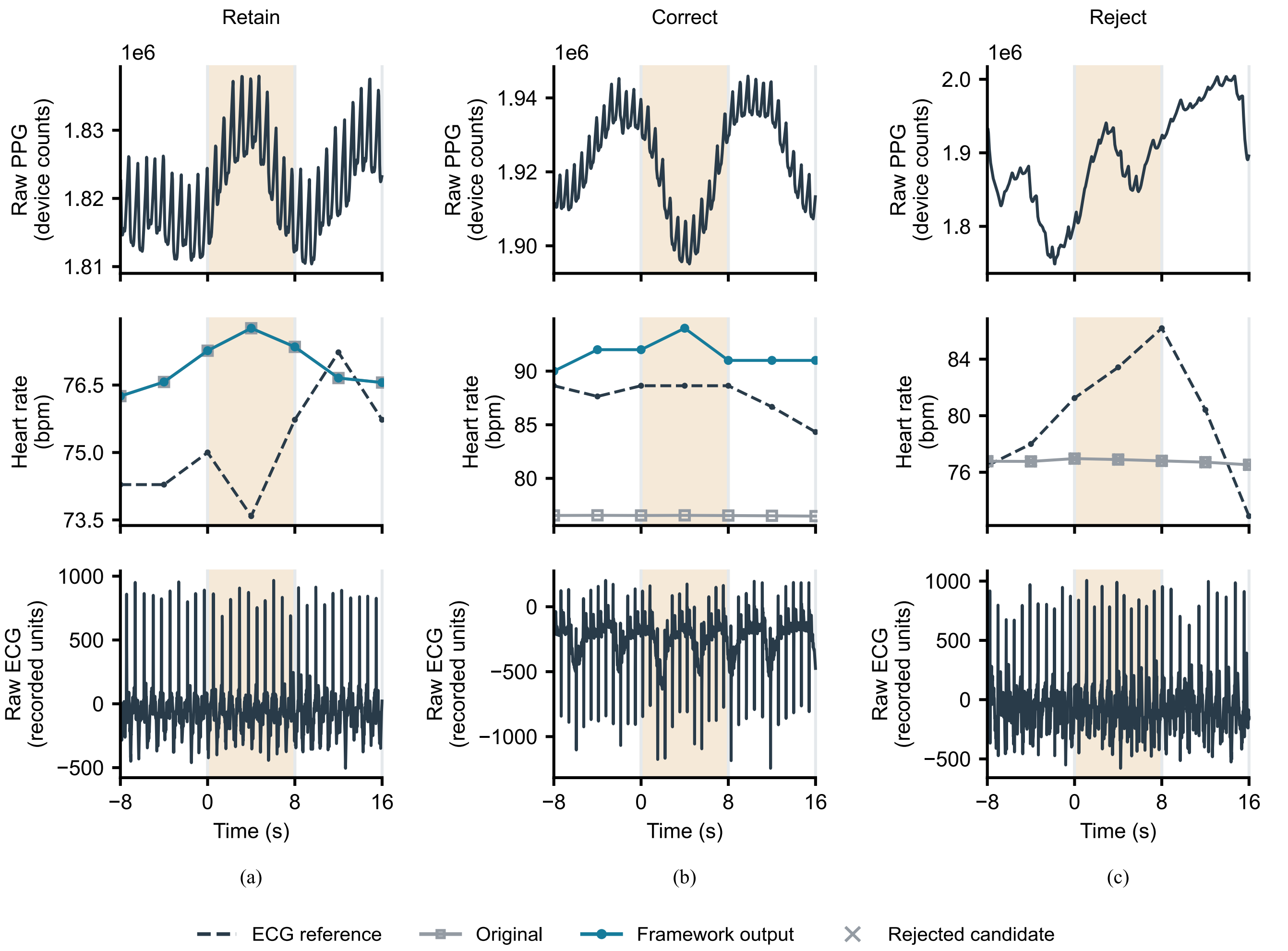

Fig. 2. Selective correction examples from GalaxyPPG with Deep PPG as the fixed core estimator. (a) Retention. (b) Correction. (c) Rejection. From top to bottom, each column shows raw PPG, overlaid reference/original/framework heart rates, and raw ECG. Time is relative to the central window start; shading marks the 0–8 s central window, and heart rate points are located at window endpoints. Black dashed lines indicate the ECG reference, gray open squares the original estimates, blue circles framework outputs, and crosses rejected internal candidates only. Overlapping curves retain their actual values without artificial offsets. Signal amplitudes use the original recording units. Fold and seed are specified in S9.

*B. Error, Correction Benefits, and Harms With All Available Outputs*

Without additional rejection, the framework reduced macro-averaged MAE on all four datasets [Fig. 3(a)]. MAE decreased from 13.68 to 9.23 bpm on DaLiA and from 45.41 to 34.61 bpm on Wrist. It decreased from 38.50 to 21.38 bpm on Galaxy and from 39.97 to 26.08 bpm on Wild. The Galaxy reduction from 38.50 to 21.38 bpm corresponds to the maximum relative reduction of 44.5% reported in the abstract. This percentage is calculated from the macro-averaged MAEs, rather than by averaging estimator-specific percentage reductions. Table S3 reports the confidence intervals. Shared coverage was 100.00%, 98.29%, 97.81%, and 99.34%, respectively; the first value is rounded (Table S10). These reductions therefore did not depend on rejecting additional windows. Results were averaged equally across estimators within participants and then across participants. In Fig. 3(a), each dot represents the participant-averaged MAE of one core estimator under the Original or Framework condition. Open diamonds indicate equal-weight means across the ten estimators; these are summaries of errors, not ensemble predictions. Figs. S5, S7, and S8 show participant-level and estimator-level results, and Table S9 reports MAE reductions for selected estimators. These comparisons distinguish the final error after correction from the size of the correction gain; neither alone establishes a universally preferred core estimator.

Beneficial corrections were more frequent than harmful corrections on every dataset [Fig. 3(b)]. Beneficial proportions were 16.06%, 30.74%, 36.64%, and 26.99% on DaLiA, Wrist, Galaxy, and Wild, respectively. Harmful proportions were 1.68%, 6.77%, 3.52%, and 3.12%; the remaining 82.26%, 62.49%, 59.84%, and 69.88% had error changes within ±1 bpm. All three categories use the same paired-window denominator. The middle category includes unchanged outputs and small error changes, so it does not equal the retention rate. Severe added error occurred in 0.37%, 3.89%, 1.61%, and 1.35% of all reference-eligible windows, respectively (Table S3). These proportions describe the frequency of error changes and should be read alongside final MAE.

The primary analysis retained all ten prespecified implementations. Leave-one-estimator-out analyses tested whether a single implementation dominated the macro-averages. Supplementary Methods S21 and Figs. S7 and S21 report these comparisons, implementation differences, and the rationale for separately excluding the local TROIKA variant.

*C. Contributions of Correction and Selection at Fixed Target Coverage*

At 80% target coverage, numerical correction accounted for most of the total MAE reduction [Fig. 4(e)]. Correction contributed 3.88, 11.24, 17.79, and 14.70 bpm on DaLiA, Wrist, Galaxy, and Wild, respectively. Window selection contributed 2.38, 0.95, 1.50, and 0.32 bpm (Table S5). Ratios of aggregate means attributed approximately 62%, 92%, 92%, and 98% of the reduction to correction. Same-risk rejection accepts the same windows as the framework and preserves original estimates, making this comparison independent of changes in the accepted set. The percentages summarize this operating point; they do not describe every participant or estimator.

Fig. 4(a)–(d) show error curves across target coverage levels; estimator-specific curves are provided in Figs. S1–S4. Reducing target coverage from 80% to 60% decreased macro-averaged MAE from 6.22 to 4.63 bpm on DaLiA, from 32.47 to 30.59 bpm on Wrist, from 18.39 to 15.70 bpm on Galaxy, and from 24.45 to 22.72 bpm on Wild. These reductions came at the cost of fewer retained outputs. Numerical correction contributed more than window selection to the total MAE reduction at both operating points. Accepted windows were determined by ranking risk scores within target recordings. Insufficient finite predictions and integer rounding of accepted counts could reduce actual coverage below the target. This evaluation differs from direct transfer of fixed source thresholds, which is reported separately.

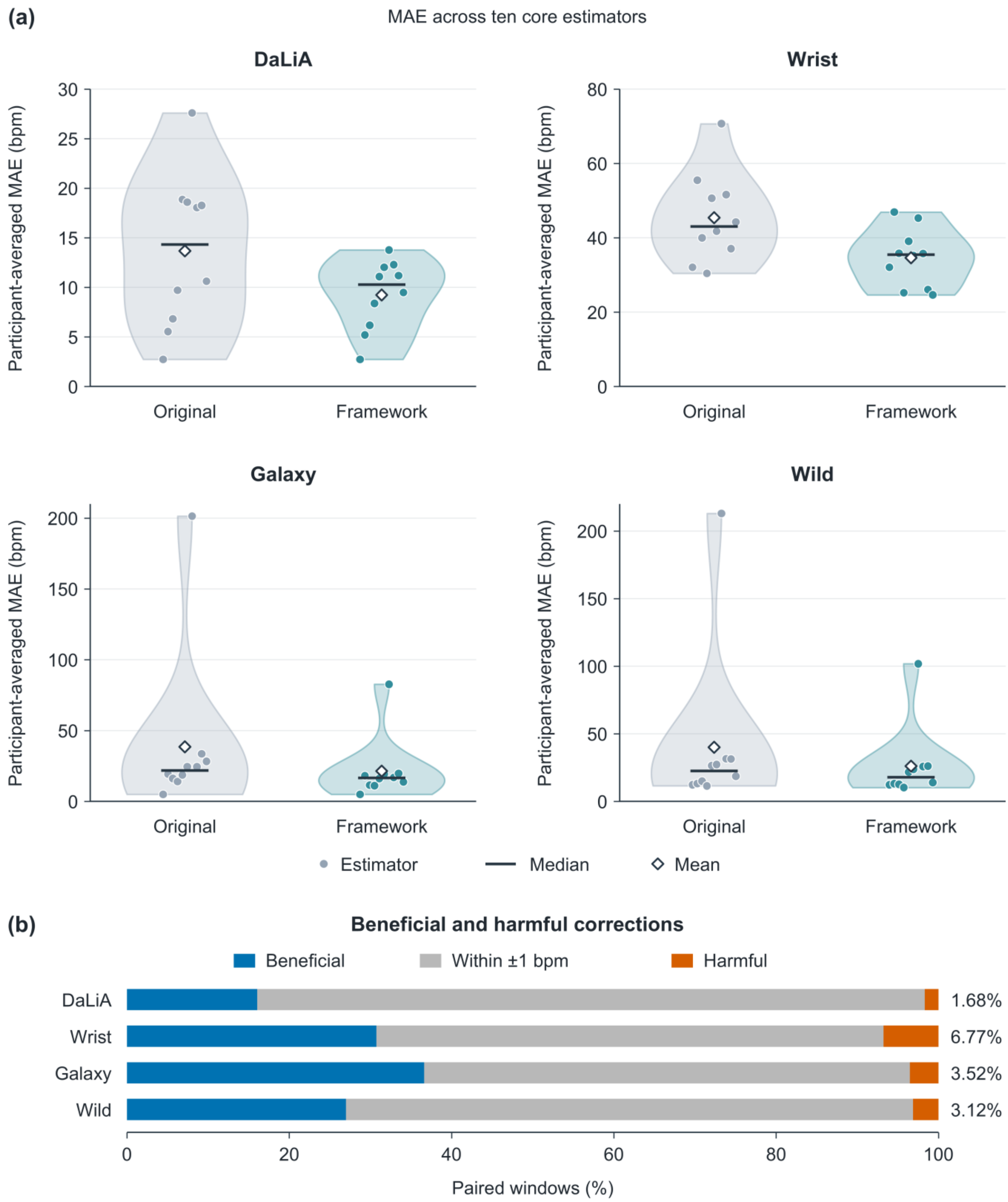

Fig. 3. Error and correction benefits and harms for ten prespecified core estimators using all available outputs. (a) Distributions of participant-averaged MAE across the ten estimators before and after correction. Each dot represents one estimator, with configurations averaged within each participant and then participants weighted equally. Violin widths show kernel-density estimates truncated to the observed range; short horizontal lines indicate medians, and open diamonds indicate equal-weight means across estimators. The violins describe between-estimator variation, not participant-bootstrap confidence intervals. The four dataset panels use different vertical-axis ranges, and all ten estimators are retained. (b) Macro-averaged proportions of beneficial corrections (blue), error changes within ±1 bpm (gray), and harmful corrections (orange) in the same paired window set. Beneficial and harmful corrections decrease and increase absolute error, respectively, by more than 1 bpm relative to the original estimate. Gray includes unchanged outputs and small error changes; orange proportions are labeled to the right. Each estimator is compared on the same available windows without additional risk rejection. Source configurations are averaged within each participant and estimator, followed by equal weighting of estimators and participants; heart rate predictions are not fused. Panel (b) shows composition point estimates. Table S3 reports these proportions and macro-averaged MAEs, together with participant-bootstrap 95% confidence intervals from 10,000 resamples. Participant counts and coverage are provided in Table S10, and estimator-composition sensitivity in Fig. S21.

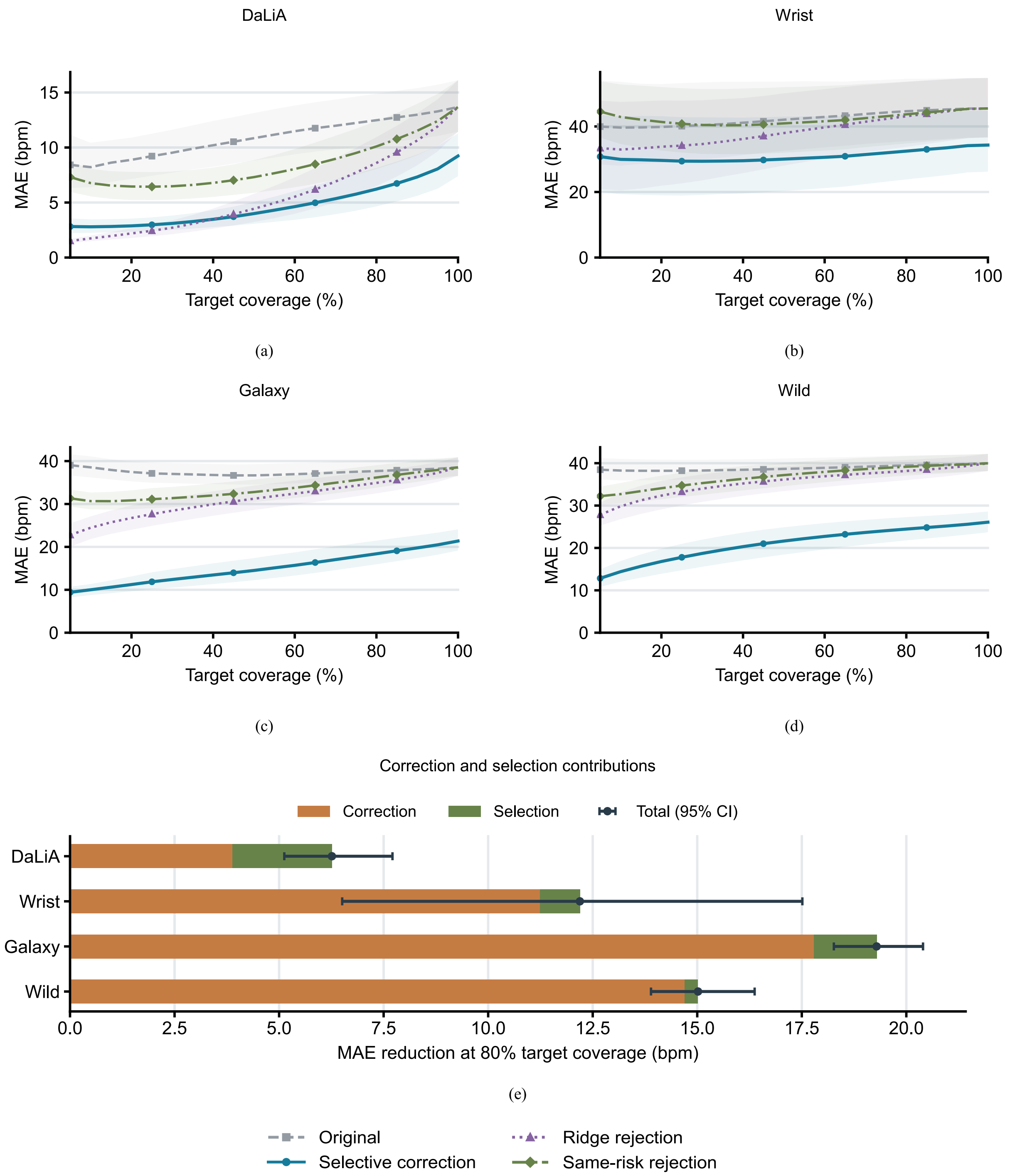

Fig. 4. Risk–coverage relationships and paired decomposition of MAE reduction. (a)–(d) DaLiA, Wrist, Galaxy, and Wild. Ridge rejection selects windows using original-error predictions from source-domain ridge regression; same-risk rejection uses the framework risk scores and accepted set. Both retain original heart rates. (e) At 80% target coverage, orange denotes correction gains on the same accepted set, green denotes selection gains with original heart rates fixed, and points indicate their sum. Positive values indicate error reduction. Curve shading and error bars for total reduction indicate participant-bootstrap 95% confidence intervals; total intervals are calculated from paired total effects.

### *D. Effects of Module Ablations on Error and Correction Harm*

Removing motion evidence or historical memory increased macro-averaged MAE on every dataset (Table II). Motion ablation increased MAE by 0.97, 5.98, 6.08, and 7.08 bpm on DaLiA, Wrist, Galaxy, and Wild, respectively. Memory ablation increased MAE by 0.87, 1.52, 1.17, and 1.35 bpm. Harmonic ablation increased MAE by 0.61 and 0.93 bpm on DaLiA and Wrist but reduced it by 1.62 and 0.94 bpm on Galaxy and Wild. The Wrist interval for harmonic ablation included zero (95% confidence interval: −0.23 to 2.10 bpm). Tables II and S6 report the remaining intervals.

TABLE II
EFFECTS OF MODULE ABLATIONS ON MEAN ABSOLUTE ERROR

| Dataset | No harmonic ΔMAE (bpm) | No motion ΔMAE (bpm) | No memory ΔMAE (bpm) |
|---|---|---|---|
| DaLiA | +0.61 [+0.36, +0.85] | +0.97 [+0.74, +1.24] | +0.87 [+0.71, +1.05] |
| Wrist | +0.93 [-0.23, +2.10] | +5.98 [+3.02, +8.67] | +1.52 [+0.52, +2.46] |
| Galaxy | -1.62 [-1.91, -1.31] | +6.08 [+5.51, +6.63] | +1.17 [+0.89, +1.48] |
| Wild | -0.94 [-1.35, -0.56] | +7.08 [+6.71, +7.43] | +1.35 [+1.10, +1.58] |

Note: ΔMAE is ablated-model MAE minus full-framework MAE, in bpm; positive values indicate higher mean error. Brackets denote participant-bootstrap 95% confidence intervals. DaLiA, Wrist, Galaxy, and Wild include 15, 8, 20, and 14 participants, respectively.

Average accuracy and severe added error did not respond consistently to ablation (Tables II and III). For example, removing motion evidence increased MAE by 6.08 and 7.08 bpm on Galaxy and Wild but decreased the severe added-error proportion by 1.25 and 0.94 percentage points, respectively. Removing historical memory also increased MAE while reducing the severe added-error proportion on these datasets. Thus, a module that improves average accuracy does not necessarily reduce every measure of correction harm.

TABLE III
EFFECTS OF MODULE ABLATIONS ON THE SEVERE ADDED-ERROR PROPORTION

| Dataset | No harmonic Δ severe added error (percentage points) | No motion Δ severe added error (percentage points) | No memory Δ severe added error (percentage points) |
|---|---|---|---|
| DaLiA | -0.12 [-0.24, -0.03] | -0.07 [-0.19, +0.02] | -0.07 [-0.19, +0.02] |
| Wrist | -0.40 [-1.42, +0.51] | +0.31 [-1.41, +3.05] | -1.02 [-1.87, -0.22] |
| Galaxy | -1.40 [-1.99, -0.97] | -1.25 [-1.85, -0.84] | -0.84 [-1.19, -0.57] |
| Wild | -0.86 [-1.11, -0.64] | -0.94 [-1.09, -0.81] | -0.44 [-0.68, -0.21] |

Note: Values are the ablated-model severe added-error proportion minus the full-framework proportion, in percentage points. Positive values indicate more severe added error. Brackets denote participant-bootstrap 95% confidence intervals. Participant counts were 15, 8, 20, and 14, respectively.

### *E. Transfer of Fixed Source-Domain Thresholds*

Fixed source thresholds did not preserve their coverage or error targets on external datasets (Table IV). The nominal 80% coverage threshold yielded 79.07%, 42.29%, 45.48%, and 30.50% coverage on DaLiA, Wrist, Galaxy, and Wild, respectively. A separate rule targeted source MAE no greater than 5 bpm. After transfer, output MAE exceeded 5 bpm in 39.55%, 91.45%, 49.72%, and 73.47% of nonempty configurations, respectively. Zero-output configuration counts were 48/450, 183/1200, 365/3000, and 227/2100. These configurations were excluded from the exceedance denominator. The two rules test coverage transfer and error-budget transfer separately. Table S7 and Fig. S9 provide detailed results.

TABLE IV
TRANSFER OF COVERAGE AND ERROR TARGETS WITH FIXED SOURCE THRESHOLDS

| Dataset | 80% source-coverage threshold Actual coverage (%) | Source MAE 5 bpm threshold Exceeding / nonempty (%) | Source MAE 5 bpm threshold Zero-output / all |
|---|---|---|---|
| DaLiA | 79.07 [73.80, 84.16] | 159/402 (39.55%) | 48/450 |
| Wrist | 42.29 [38.46, 46.42] | 930/1017 (91.45%) | 183/1200 |
| Galaxy | 45.48 [40.50, 50.32] | 1310/2635 (49.72%) | 365/3000 |
| Wild | 30.50 [26.07, 35.73] | 1376/1873 (73.47%) | 227/2100 |

Note: Coverage is a participant macro-average; brackets denote bootstrap 95% confidence intervals. Exceedance means MAE > 5 bpm among retained outputs, with nonempty configurations as the denominator. Zero-output configurations are counted separately and excluded from that denominator. Configuration-count proportions are descriptive and have no reported confidence intervals. The two sets of results use different source-domain threshold-selection rules.

### *F. Prototype System and Evaluation of Acquired Data*

Fig. 5(d) compares ECG references, original Deep PPG estimates, and framework outputs during static, walking, and running recordings from one participant. Recovery-rest recordings were also included in the aggregate analysis. Without additional rejection, ten-estimator macro-averaged MAE decreased from 9.09 to 4.45 bpm [Fig. 5(e)]. Eight estimators improved, BeliefPPG was essentially unchanged, and WFPV increased slightly from 1.09 to 1.10 bpm. Restricting the comparison to 289 windows available for all ten estimators reduced MAE from 9.24 to 4.33 bpm. Table S12 and Fig. S12 report estimator-specific counts and errors; Supplementary Methods S21 examines sensitivity to estimator composition.

At 80% target coverage, each method accepted 259 of 324 reference-eligible windows, giving 79.94% actual coverage. Macro-averaged MAE was 8.60 bpm with original-risk selection, 7.49 bpm with same-risk rejection, and 3.30 bpm with the full framework. Correction contributed 4.19 bpm of the total 5.30 bpm reduction, while selection contributed 1.11 bpm, or approximately 79% and 21%, respectively [Fig. 5(f)]. Ridge rejection achieved 6.83 bpm, outperforming same-risk rejection at this operating point. The framework's risk ranking therefore did not improve on every selection baseline. Figs. S13 and S14 show the full curves and estimator-specific decompositions.

The prototype recordings also showed a trade-off between accuracy and conservative correction. Without additional rejection, adding M4 reduced the macro-averaged harmful correction rate from 2.18% to 0.42% but increased MAE from 2.57 to 4.45 bpm. Figs. S12 and S16 report these comparisons. Nominal 80% source-coverage thresholds yielded 58.21%–86.85% actual coverage across methods. The separate 5 bpm source-MAE thresholds produced both error exceedances and zero-output configurations (Table S13; Fig. S17). Macro-averaged MAE decreased during static, walking, and running recordings. However, estimator-specific gains and protection of originally accurate estimates were inconsistent (Figs. S19 and S20). Figs. S10, S15, and S18 show synchronized raw signals, red-channel sensitivity, and preprocessing examples, respectively.

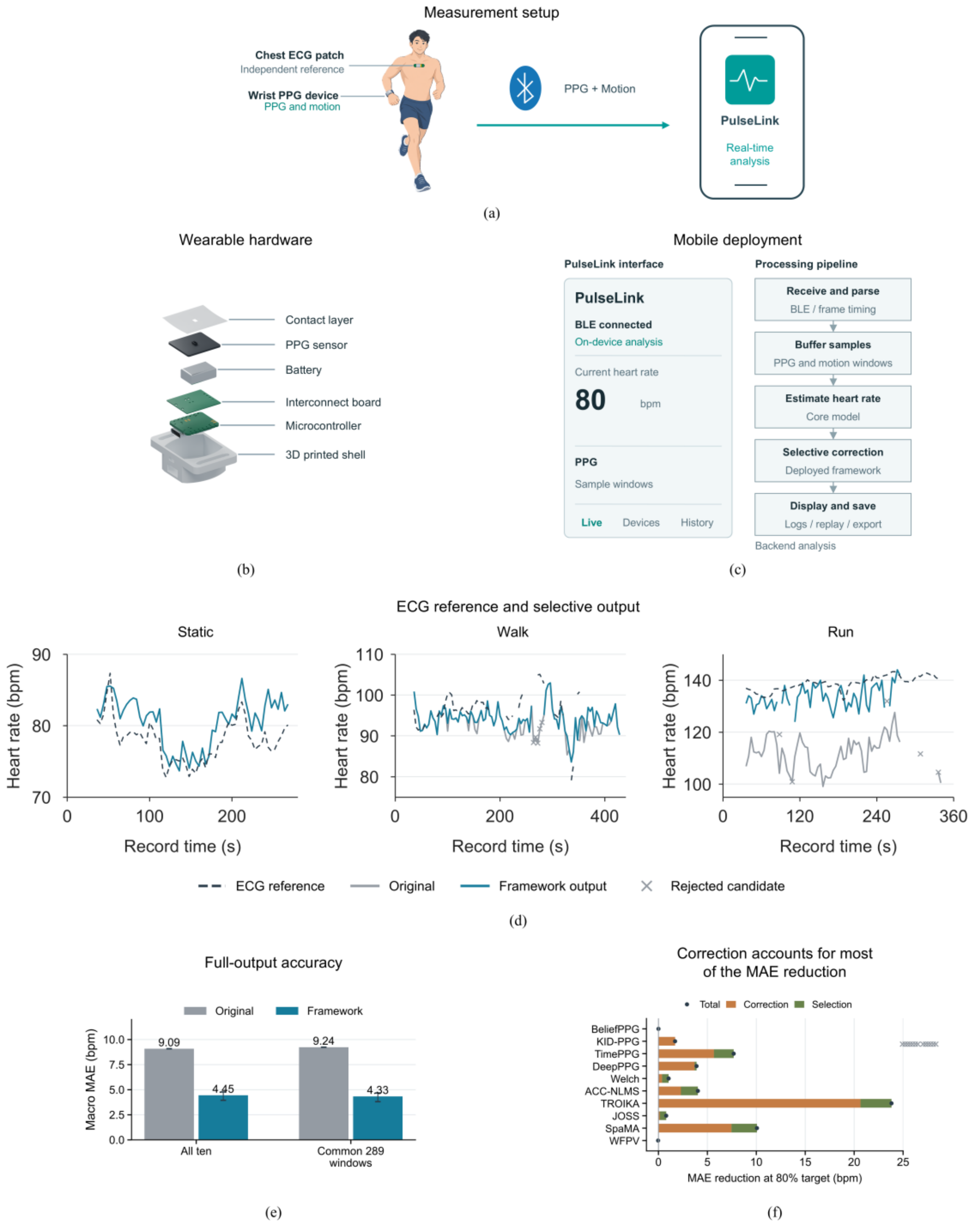

Fig. 5. Wearable acquisition, smartphone deployment, and prototype evaluation. (a) Measurement setup and independent ECG reference. (b) Hardware structure and appearance. (c) Smartphone interface and processing workflow for the deployed model and framework. (d) Original Deep PPG estimates, ECG references, and framework outputs in acquired static, walking, and running recordings using fixed source thresholds; crosses denote rejected internal candidates. The three states use different vertical-axis ranges to resolve within-state variation. Configurations are specified in S20.3. (e) Ten-estimator macro-averaged MAE pooled across four states without additional rejection, and results for the 289 common windows. (f) Method-specific correction and selection gains at 80% target coverage; points denote totals, and negative components are retained. Each strategy accepts 259/324 windows. Correction fixes the framework accepted set, whereas selection fixes original heart rates. Bars are means across frozen source configurations; error bars in (e) indicate configuration ranges. n = 1; statistics are from backend analysis of saved recordings. Panel (f) uses risk ranking, unlike (e).

## IV. DISCUSSION

### *A. Selective Correction Relative to Original Estimates*

Evaluating candidate gains and harms reduced average heart rate error while leaving the core estimators fixed. The all-output comparison separates this improvement from the effect of rejecting additional windows. Beneficial corrections outnumbered harmful corrections on all four datasets, although some windows still deteriorated [Fig. 3(b)]. These proportions measure frequency rather than the magnitude of error changes and must be interpreted alongside MAE. At 80% target coverage, correction on the same accepted windows contributed most of the average error reduction. Thus, the observed gains cannot be explained solely by withholding difficult outputs.

The architecture separates candidate generation from the decision to replace an estimate. M3 forms a distribution from current and historical evidence; M4 evaluates candidates relative to the original estimate. High posterior probability alone does not trigger replacement. A candidate must also satisfy the predicted-gain, predicted-harm, and posterior-error conditions. The original estimate remains available when no replacement qualifies, and a separate risk gate determines whether to release the selected value. The three outcomes therefore arise from sequential decisions rather than a single three-class network.

The term agent describes the organization of observations, a limited internal state, and constrained action selection [14]. Execution follows a fixed sequence. Historical states are updated from input evidence, without using previous output heart rates or online reference feedback. The results support this implementation of selective correction. They do not establish autonomous planning, collaboration among estimators, or one set of auxiliary weights that works for every estimator.

### *B. Relationship to Existing Estimation Methods*

The auxiliary modules reorganize existing signals for relative-error assessment rather than introduce new sensor measurements. TROIKA, JOSS, SpaMA, and WFPV already use spectral separation, motion processing, or tracking [4]–[7]. Deep PPG and BeliefPPG also use time–frequency information, and BeliefPPG additionally models heart rate dynamics and uncertainty [3], [9]. The observed gains therefore cannot be attributed simply to an absence of spectral or historical information in these methods.

The auxiliary modules may complement existing estimators through their representations and training objectives. M1 and M2 retain power, harmonic, and motion features at multiple time scales on a common heart rate grid. M3 summarizes recent history, and M4 learns paired-error differences for a specific estimator. Scalar heart rates or distribution summaries may omit information useful for these comparisons. Additional supervision may also help the auxiliary models learn estimator-specific biases. We did not independently vary representation, model capacity, and supervision, so their individual contributions remain unresolved.

Post-calibration is an established approach [11]. Device adaptation, temporal modeling, and multisite measurements offer other routes to improved estimation [18]–[20]. We did not compare all these approaches under matched inputs, training data, and computational budgets. Our focus is the assessment of relative gains and harms before replacement. If an estimator is already Bayes-optimal for absolute error, post-processing with identical information cannot universally reduce risk under the same target distribution [14]. Any additional benefit must therefore be demonstrated for the specific estimator and operating conditions.

*C. Core-Estimator Differences and Module Contributions*

Model-specific results constrain the interpretation of macro-averages. Participant-paired MAE reductions for KID-PPG were 0.35, 8.56, 1.09, and 0.12 bpm on DaLiA, Wrist, Galaxy, and Wild, respectively. The prespecified JOSS, SpaMA, and WFPV implementations also improved on average across all four datasets (Table S9; Fig. S8). Additional gains can therefore occur for estimators that already incorporate spectral or temporal processing, although their magnitudes depend on the model and dataset.

BeliefPPG changed by less than 0.001 bpm in absolute MAE on every dataset, indicating that the framework largely retained its outputs. Overlap between its representation and the auxiliary modules is one possible explanation. However, near-zero change establishes neither redundancy nor optimality. Original error and scope for correction also differ across implementations. Macro-averages should therefore be read with estimator-specific results and the composition analyses in Supplementary Methods S21, rather than used to rank the original published methods.

The ablations support a contribution from motion evidence and recent history to average accuracy. Harmonic evidence showed less consistent benefits across datasets. Pulse and motion rhythms, including their harmonics, can occupy the same frequency bands [2], [4]–[7]. Feature weights learned on source data may therefore transfer poorly to different activities or sensors. These are possible explanations for the Galaxy and Wild ablation results, rather than established causes. Each ablation also removes related interaction features, preventing an independent attribution of the total gain to individual modules.

*D. Correction Harm and Decision Traceability*

Mean error and correction harm capture different aspects of performance. Ablations with opposing effects in Tables II and III show that a module reducing overall MAE may still increase severe deterioration in a small number of windows. Model comparisons should therefore jointly report error changes on the same accepted set, harmful correction rates, and coverage. Uncertainty estimation and selective prediction provide a basis for identifying low-confidence outputs [9], [10], [12], [13]. When a system both modifies estimates and rejects outputs, the contributions of these operations must also be distinguished.

The framework exposes intermediate quantities that can be used to trace decisions, including candidate evidence, historical states, posterior distributions, predicted gains and harms, and utility rankings. Geometric constraints rule out incompatible error predictions but cannot guarantee accurate harm estimates. Fig. 2 aligns signals with actions without showing complete candidate-level trajectories. Similarly, high motion coherence alone does not prove that a spectral peak originates from motion. Tracing an algorithmic decision and explaining its physiological cause require different evidence.

The paired decomposition separates correction from selection but remains descriptive. Correction contributed more than selection at 80% target coverage (Fig. 4), yet some accepted windows still deteriorated. The harmful example in Table S8 illustrates why retention options and candidate constraints must be evaluated using actual outputs. Their presence alone does not establish protection against harmful corrections.

*E. Cross-Dataset Transfer and Deployment Limitations*

Lower average error after transfer does not establish sufficient absolute accuracy for deployment. External-dataset MAE remained 21.38–34.61 bpm. Transferring thresholds calibrated for 80% source coverage produced only 30.50%–45.48% coverage on external data. Error-based thresholds also failed to consistently maintain the 5 bpm target. Risk scores and thresholds therefore need independent validation in the intended operating conditions. The 2 bpm predicted-harm limit is a decision criterion, not a bound on actual error.

Data history and implementation differences further limit generalization claims. All four datasets were used during earlier development, and participant provenance is incomplete for some pretrained weights. The current participant splits and weight freezing do not create a confirmatory set untouched throughout the research process. Reference eligibility rules differ across datasets, and Galaxy uses a historical ECG cache checked for consistency. Participant-bootstrap intervals do not capture the uncertainty of full retraining. In addition, some preprocessing and core-estimator operations are retrospective, and KID-PPG retains per-recording adaptation without heart rate labels. Cross-method statistics therefore cannot be equated directly with the real-time accuracy of the deployed smartphone configuration.

The prototype combines wearable acquisition, smartphone analysis, and backend evaluation. Its recordings support paired-error comparisons after input adaptation, but they comprise only four recordings from one participant. Fig. 5(e) and Supplementary Methods S21 report common-window and estimator-composition analyses. The cross-estimator averages represent neither the deployed smartphone model alone nor population-level performance. M4 reduced the overall harmful correction rate but increased MAE, and some protection metrics worsened for individual methods (Fig. S20). Its usefulness therefore depends on the acceptable balance between error, harm, and coverage.

Smartphone analysis runs in real time; the multi-method backend comparisons use saved recordings and standardized preprocessing. Earlier exploration of the same participant, zero-phase filtering, and whole-block amplitude normalization limit these comparisons. Repeated source configurations do not remove those limitations. Reconstructed reception times do not measure hardware transmission latency, and the low effective reception rate limits recoverable bandwidth. Independent end-to-end tests should use a fixed deployed model, defined input units, causal preprocessing, and prespecified thresholds. Such tests should assess accuracy, correction harm, coverage, and latency against synchronized reference measurements.

## V. CONCLUSION

We presented a selective correction framework that combines signal evidence, recent history, and predicted gains and harms around a fixed heart rate estimator. Across ten estimators and four public datasets, the framework reduced macro-averaged MAE without additional rejection. At 80% target coverage, numerical correction on the same accepted windows accounted for most of the error reduction. We also implemented wearable acquisition and smartphone analysis, with backend comparisons on prototype recordings. Gains varied by estimator and dataset, harmful corrections remained, and source-calibrated thresholds did not transfer reliably. Further evaluation in independent participants should measure end-to-end accuracy, correction harm, output coverage, and latency.

## DECLARATIONS AND REPRODUCIBILITY

Data and code availability: Public datasets are described in [3] and [15]–[17]. The local study archive contains participant summaries, window-level predictions, configuration results, reference-eligibility records, analysis and plotting scripts, prototype recordings, timing reconstruction, source-model bindings, figure data, and editable figures. Sources and local indices for complete public recordings are retained separately. Public repository links for the study code, weights, and results are not provided in this version.

Author self-recordings and ethics status: The new prototype recordings were collected from the first author for deployment and technical feasibility evaluation. These recordings do not establish generalization across participants or clinical validity. No institutional ethics approval or exemption determination was obtained for these author self-recordings. No other participants were recruited for the prototype evaluation.

# Supplementary Information

# An Agent Architecture with Tools and Memory for Selective Correction of Wrist PPG Heart Rate Estimates

Xu Wang[1,2], Shuhan Yi[2], and Zhen Cheng[2*]

[1] School of Engineering, The Hong Kong University of Science and Technology, Hong Kong 999077, China

[2] School of Biomedical Engineering, Tsinghua University, Beijing 100084, China

*Corresponding author:

Zhen Cheng, School of Biomedical Engineering, Tsinghua Medicine, Tsinghua University, Beijing 100084, P. R. China. Email: zcheng@mail.tsinghua.edu.cn

This supplement provides the reference eligibility criteria, training and calibration procedures, module definitions, and proofs supporting the main manuscript. Sections S11–S14 specify the signal features, historical state, candidate construction, and gain–harm model. Section S19 defines the outputs and evaluation metrics. Section S20 describes prototype acquisition, preprocessing, and offline evaluation; Section S21 examines sensitivity to estimator composition. Tables S1–S13 and Figs. S1–S21 provide supporting results. Reference numbers are shared with the main manuscript.

# Table of contents

# SUPPLEMENTARY METHODS

## S1. Datasets, Reference Eligibility, and Participant Splits

The public datasets are PPG-DaLiA, Wrist Exercise, GalaxyPPG, and WildPPG, abbreviated here as DaLiA, Wrist, Galaxy, and Wild [3], [15]–[17]. Heart rate and mean absolute error (MAE) are reported in beats per minute (bpm). Signal inputs are wrist photoplethysmography (PPG) and triaxial acceleration (ACC). Electrocardiography (ECG)-derived heart rates or published heartbeat annotations are used only for training targets, source calibration, and evaluation, not as deployment inputs. The common evaluation grid uses 8 s windows with a 4 s stride. Multiple Wrist activity recordings are pooled within each participant; its 19 recordings are not treated as 19 independent participants. Wild uses wrist green-channel PPG and the corresponding acceleration, without combining body sites. Registered native outputs from each core estimator are retained and aligned to the common time grid.

Table I reports the actual frozen analysis scope, rather than the full populations described in the dataset papers. Galaxy originally reported 24 participants [16], whereas the current input manifest contains 23 participants, P02–P24; no P01 results are imputed. Galaxy participants P05, P07, and P22 and Wild participants fex and p5d have no eligible references under the current rules. Their inputs, predictions, and eligibility records are retained, but their mean errors are not calculated. These exclusions depend only on reference eligibility, not PPG estimation error. The four datasets include 62 participants, of whom 57 have eligible references. Analyses remain separate by dataset.

DaLiA uses the frozen ECG-derived window heart rates. Wrist uses published heartbeat events. Each 8 s window must contain at least three events, at least two RR intervals corresponding to 35–220 bpm, and at least 80% valid intervals. Heart rate is calculated from the median valid RR interval. Galaxy reuses a historical ECG reference cache whose bindings to raw inputs, processing code, and time grids were checked. Eligible windows require agreement within 5 bpm between XQRS and a second ECG estimator; the XQRS value is used. Cache consistency checks do not constitute independent validation of reference accuracy. Wild references are recomputed from the current raw ECG using segment-wise R peaks and stable RR intervals. Existing recording-level rules require at least 80% reference coverage, 90% physiological RR intervals, and 80% stable RR intervals. All windows are ineligible if a recording fails these rules. The two participants without eligible references were identified by the current rules, rather than by reusing an earlier 14-participant list. Reference rules differ across datasets, so cross-dataset error differences cannot be attributed solely to devices.

DaLiA uses five fixed groups: G0={S1,S6,S11}, G1={S2,S7,S12}, G2={S3,S8,S13}, G3={S4,S9,S14}, and G4={S5,S10,S15}. Outer fold k evaluates Gk, calibrates on G(k+1 mod 5), and trains on the remaining nine participants. Each configuration uses seeds 219, 220, and 221. A participant contributes to DaLiA results only in their evaluation fold. All participants had been accessed during earlier development. These splits isolate participants during auxiliary fitting but cannot make the evaluation set untouched throughout the research process.

## S2. Retention, Correction, and Rejection

The original estimate is the output of the fixed core estimator. The auxiliary modules generate alternative candidate heart rates. After selecting the original estimate or a candidate, a separate risk gate determines whether to release the selected value. Reference heart rates are not used to generate candidates during inference.

A candidate's true error is its absolute difference from the reference heart rate. Relative net gain is original-estimate error minus candidate error and may be negative. Added error is the positive part of candidate error minus original-estimate error. Retention outputs the original estimate; correction outputs another candidate; rejection withholds heart rate for the window. Rejection changes output status but does not remove the window from the reference-eligible set.

The harmful correction rate is the proportion of paired accepted windows in which absolute error increases by more than 1 bpm relative to the finite original estimate. This differs from a heart rate change exceeding 1 bpm and from the continuous added-error target used in training. We also record deterioration among windows whose original error is at most 5 bpm and tail measures for error or added error exceeding 10 bpm. These measures expose local failures that mean error can conceal.

### S3. Physiological Evidence and Candidate Construction

The auxiliary posterior combines an estimator-related prior with 30-dimensional physiological evidence. Features describe PPG structure around candidate frequencies, half- and double-frequency relationships, motion-related spectra, and their interactions. Estimators with native predictive distributions and those producing only scalar heart rates use separate adapters. Supplementary evidence-based risk for scalar estimators is not described as their native confidence. For KID-PPG, the actual continuous mean is retained as the original estimate, rather than replaced by a mean on a truncated grid.

The history module applies learnable channel-wise decay to five aligned feature positions, including the current window. Current evidence, historical state, and their difference at each candidate frequency are used to model the auxiliary posterior. The posterior head has 90 inputs, 24 hidden units, and sigmoid linear unit (SiLU) activation. It predicts a bounded energy adjustment, which is normalized with the temperature-scaled log prior to form a heart rate distribution. History indices never access future windows. This design compares current evidence with recent context, but some core-estimator preprocessing remains retrospective; it does not validate end-to-end real-time operation.

The auxiliary posterior provides nine candidates: the original estimate, mean, mode, 10th, 25th, 50th, 75th, and 90th percentiles, and the midpoint between the original estimate and posterior median. The original estimate always occupies the first position. Each candidate has a posterior expected absolute error, expected added error relative to the original estimate, and posterior support for error within 5 bpm. This finite candidate space allows the framework to retain an estimate instead of producing a new value in every window.

### S4. Consistent Gain–Harm Decisions

M4 uses 106 features per candidate. The first 16 describe candidate and original values, correction direction and magnitude, posterior error, added error, support, and distribution statistics. The remaining 90 comprise current evidence, historical state, and current-evidence differences between the candidate and original estimate. This candidate-relative difference serves a different purpose from the temporal difference in the posterior head.

M4 contains two linear branches. A branch mapping 106 features to three outputs produces allocation logits for every candidate; a branch mapping 106 features to one output produces a base-error parameter from the original-estimate features only. Softmax and softplus transformations are followed by consistent decoding to jointly predict gain, added error, and absolute error. Candidate distance limits achievable gain, while a nonnegative base error and triangle-inequality constraints maintain consistency among predictions. Training targets are the actual paired absolute errors, with an extra penalty for underestimating added error. These constraints govern relationships between predicted quantities. They do not guarantee calibration in unseen participants or safety in individual windows.

The primary configuration requires predicted net gain of at least 1 bpm, predicted added error no greater than 2 bpm, and lower posterior expected error than the original estimate. Eligible candidates are ranked by predicted net gain minus twice predicted added error. The original estimate remains available with utility zero. Ties favor earlier candidates, so equal utility does not force a correction. The primary configuration has no correction-magnitude cap. These thresholds apply to predictions and do not bound actual error increases to 2 bpm.

**S5. Auxiliary Training and Evaluation With Fixed Configurations**

All core estimators remain fixed during auxiliary training, and auxiliary weights are fitted separately for each estimator. Within the nine source-training participants of each outer fold, cross-fitting uses the three existing participant groups. M3 is trained on six participants and generates out-of-fold posteriors for the other three. M4 uses only these posteriors and history features from the corresponding M3 models. The final M3 is trained on all nine source participants before predicting for three calibration and three evaluation participants. This avoids using M3 in-sample posteriors to train M4. It does not remove participant exposure during original-estimator pretraining or earlier research.

Each posterior head uses AdamW with learning rate 0.002, weight decay 0.03, batch size 96, and 2000 training steps. References become Gaussian soft targets with a standard deviation of 3 bpm. The loss combines cross-entropy and cumulative-distribution discrepancy, with participant-balanced sampling, prior-channel perturbation, and gradient clipping. M4 uses participant-balanced sampling, batch size 256, learning rate 0.003, weight decay 0.1, and 2000 steps. Its features are standardized from source data and clipped. Final-step weights are used throughout; checkpoints are not selected on the evaluation set.

Ten estimators, five outer folds, three seeds, and four variants requiring retraining give 600 configurations. Each trains three cross-fitted posterior heads, one final posterior head, and one M4 module, totaling 2,400 posterior-head fits and 600 M4 fits. Calibration and evaluation predictions were archived before final scoring. The complete internal archive contains 81,000 participant-, strategy-, and coverage-mode metric records. This manuscript reports only the specified strategies, and the plotting package records row counts before and after filtering. Analyses use the archived results without retraining or configuration reselection.

**S6. Comparators, Risk Scores, and Thresholds**

The ten registered implementations are BeliefPPG, KID-PPG, TimePPG, Deep PPG, Welch, ACC-NLMS, TROIKA, JOSS, SpaMA, and WFPV. Names identify the implementations evaluated here, not the best results in their original papers. The local TROIKA implementation is a registered 25 Hz paper-guided variant, not the authors' code or a validated reproduction. It uses custom singular spectrum analysis (SSA) component pairing and spectral tracking, with outputs constrained to 24–300 bpm. Training-participant exposure could not be fully established for estimators using author-provided weights. Source-training and early-stopping splits for locally trained estimators also differ from the current auxiliary outer folds. The study therefore measures auxiliary gains around fixed estimators, rather than providing a fair from-scratch ranking of the ten published methods.

Baseline risk definition. The score $r_{0,t}$ is an estimator-specific indicator of uncertainty in the original estimate $h_{0,t}$, computed without reference heart rate at inference. It is not the observed absolute error. The risk score and the adapted heart rate prior $p_t$ are separate quantities: they need not be constructed from the same weights. Scores are used for within-estimator ranking and estimator-specific source calibration; their magnitudes are not compared across estimators.

BeliefPPG uses the standard deviation of its native predictive distribution, evaluated on the native heart rate bins before adaptation to the common grid: $r_{0,t} = [\Sigma_k \pi_{t,k}(b_k - \mu_t)^2]^{1/2}$, where $\pi_{t,k}$ are native predictive probabilities, $b_k$ are native bin values, and $\mu_t = \Sigma_k \pi_{t,k} b_k$. KID-PPG uses its predicted Gaussian standard deviation, $r_{0,t} = 1 + \text{softplus}(z_t)$, where $z_t$ is the model output controlling scale and $\text{softplus}(z) = \ln(1 + \exp(z))$. These standard deviations are measured in bpm and are retained from the fixed estimator outputs.

For the other eight estimators, the adapter computes spectral entropy: $r_{0,t} = -\Sigma_k s_{t,k} \ln(s_{t,k})$, where $s_{t,k} \geq 0$ and $\Sigma_k s_{t,k} = 1$. Natural logarithms are used, without division by the logarithm of the number of bins. Larger entropy indicates more diffuse spectral evidence. These supplementary indicators are not native calibrated error estimates.

For Welch, $s_{t,k}$ is the normalized PPG Welch power interpolated onto the 35–220 bpm grid. For ACC-NLMS, it is normalized spectral power of the ACC-filtered PPG residual. TROIKA uses its normalized sparse spectrum, JOSS its cleaned sparse spectrum after motion suppression, SpaMA its surviving candidate spectral mass, and WFPV its exported spectral evidence. For these six estimators, spectra are interpolated onto the 1 bpm grid, floored at $10^{-12}$, and divided by their sum. An all-zero finite spectrum consequently yields uniform weights. WFPV replaces nonfinite spectral rows by uniform evidence before normalization.

For TimePPG, the risk weights are obtained from the mean-centered input PPG: a 1024-point FFT at 32 Hz provides power within 35–220 bpm, which is floored at $10^{-12}$ and normalized by its sum. For Deep PPG, the weights are obtained by applying softmax to the last PPG spectral-magnitude input segment; logarithm arguments are floored at $10^{-300}$ for numerical stability. For these two scalar estimators, the adapted heart rate prior is a separate Gaussian centered on the predicted heart rate with a source-calibrated scale. Their baseline risk is therefore not the entropy of that Gaussian prior.

Seven output strategies are evaluated: original estimates, a 20 bpm correction cap, the cap with posterior-gain consistency, uncapped correction, uncapped correction with consistency, ridge rejection, and same-risk rejection. The proposed method is the complete M1–M4 framework with uncapped correction and the posterior-gain consistency check. All configurations were fixed before evaluation.

Ridge rejection predicts the natural logarithm of one plus the original absolute error from 17 source-domain features generated by participant cross-fitting. The ridge regression model uses participant-balanced weights and a fixed regularization strength of 1. At test time, predicted errors rank windows or are compared with a source-calibrated threshold. Accepted windows retain their original heart rates. This comparator evaluates selection using a separate, simple error predictor. Same-risk rejection requires no additional predictor. In risk-ranked comparisons, it uses the framework's risk scores and accepted windows but retains the original heart rates. Its MAE difference from the framework therefore measures numerical correction on the same windows. Thresholds targeting source MAE of 5 bpm are calibrated separately for each strategy; accepted sets can differ under that evaluation mode.

After freezing weights, each configuration compares original risk with learned-error risk from the uncapped variant without consistency checking on three calibration participants. Risk family is selected by participant-macro-averaged MAE at 80% coverage, with ties favoring original risk. The same family is used for four M4 decision variants. Calibration also yields a pooled-quantile threshold for 80% coverage and a threshold targeting source macro-averaged MAE no greater than 5 bpm. The latter requires at least 10% coverage for every calibration participant. Zero output is allowed when no threshold is feasible. Each strategy uses thresholds calibrated for its own outputs; rejection-only strategies do not inherit corrected-output error-budget thresholds. This empirical calibration provides no finite-sample risk guarantee.

**S7. Cross-Dataset Transfer With Fixed Auxiliary Models**

All 600 configurations, spanning estimators, five source folds, three seeds, and four trained variants, are applied separately to each external dataset. No additional auxiliary fitting is performed. Source-selected risk families, decision rules, and thresholds remain unchanged; external references do not select seeds, folds, or output variants. Each external participant thus has 15 complete primary source configurations. Their metrics are averaged, rather than averaging 15 predictions into an ensemble estimate.

External estimator caches retain the processing rules of their registered implementations. Supplementary spectral risk for scalar outputs is not described as native estimator probability. Input and history availability masks propagate downstream; interpolation is not used to create eligible predictions. Physiological evidence uses a common sampling representation, 8 s and 24 s contexts, and five aligned historical positions. KID-PPG retains 1000 steps of per-recording adaptation without heart rate labels. Thus, zero auxiliary retraining does not mean that every core estimator performs no optimization. Whole-record resampling and some estimator preprocessing are retrospective. These batch results do not establish strict streaming deployment of all ten methods; S20.3 separately specifies the smartphone configuration.

Inputs, reference-file hashes, configurations, recordings, and predictions for each external dataset were archived before the final scoring procedure. All datasets had been accessed during earlier development. Fixing the analysis configurations prevents subsequent model selection at this stage, but does not create an independent confirmatory dataset. The experiments therefore assess cross-dataset transfer under fixed configurations.

**S8. Metrics, Paired Decomposition, and Statistics**

Let h0 be the original estimate, h the framework output, and y the reference. Absolute error is $e(h)=|h-y|$, and paired error change is $\Delta e=e(h)-e(h0)$, in bpm. Coverage is accepted windows with finite predictions divided by all reference-eligible windows; MAE is calculated only over accepted windows. Within the same paired accepted set, beneficial corrections have $\Delta e<-1$, small error changes have $-1\leq\Delta e\leq1$, and harmful corrections have $\Delta e>1$. These mutually exclusive, exhaustive categories share a denominator. The middle category includes unchanged outputs and small error changes; it is not a retention-action rate. Fig. 3(b) shows their macro-averaged composition, with confidence intervals in Table S3. Severe added error counts accepted windows with $\Delta e>10$ relative to all reference-eligible windows. It is neither the proportion of absolute errors exceeding 10 bpm nor a measure of clinical harm. With zero output, coverage is zero, while MAE and paired proportions remain undefined.

Evaluation modes include all available outputs without additional rejection, a fixed original-estimator 80% accepted set, independently ranked 80% sets, and fixed source thresholds for 80% coverage or 5 bpm MAE. Risk–coverage curves span 5%–100% in 5-percentage-point steps. The 80% operating point was part of the existing protocol, not optimized for these results. Target accepted counts use each participant's reference-eligible window count and are rounded down. Missing finite predictions or risk scores can reduce actual coverage below target. Ranking is a retrospective evaluation; fixed-threshold transfer is reported separately.

At 80% target coverage, let A be the framework accepted set and B the set selected by original-estimator risk. Total change, $\mathrm{MAE}(h,A)-\mathrm{MAE}(h0,B)$, equals the correction term $\mathrm{MAE}(h,A)-\mathrm{MAE}(h0,A)$ plus the selection term $\mathrm{MAE}(h0,A)-\mathrm{MAE}(h0,B)$. Same-risk rejection preserves h0 and uses framework risk. Accepted indices were verified to match the framework for every configuration. This decomposition is a paired algebraic identity, not a causal effect estimate.

Participants are the primary aggregation units. DaLiA's three seeds or the 15 external source configurations are averaged within each participant before equal weighting across participants. Cross-estimator macro-averages also weight the ten implementations equally within a participant. They summarize the registered estimator set; estimators are not treated as independently sampled algorithms, and windows are not pooled across datasets. Recording duration does not alter participant weights. Percentile 95% confidence intervals use 10000 participant-bootstrap resamples with replacement and seed 260922. Paired differences are calculated before resampling. Intervals are conditional on fitted models and exclude full-retraining variability and dependence from shared training folds. No confirmatory p-values or multiplicity-adjusted significance claims are reported.

For source MAE5 thresholds, total, zero-output, nonempty, and above-5-bpm configuration counts are reported separately. Participants without eligible references are excluded from these failure denominators. Configuration counts describe runs and are not independent sample sizes. Changes in conditional MAE caused by zero outputs are accompanied by valid-participant counts and counts with complete model replicates. Exploratory leave-one-estimator-out analyses assess sensitivity to individual estimators; none is excluded from the primary analysis. Analyses use NumPy 2.2.5 and pandas 2.2.3, and all figures derive from actual scores or archived predictions.

**S9. Selection and Display of Signal Examples**

Fig. 2 uses GalaxyPPG with Deep PPG as the fixed estimator and the complete framework (FourthUnbounded+agreement). M3/M4 weights are fixed to PPG-DaLiA outer fold 0 and seed 219; Section S1 defines the folds. The same configuration supplies retention, correction, and rejection examples. Across 23 recordings, central windows require eligible references, finite original and framework estimates and risk, and 15 consecutive neighboring windows on each side at 4 s intervals. There were 16,175 centers satisfying common validity and 15,882 satisfying context requirements, from 20 participants.

Retention requires original absolute error no greater than 5 bpm, an unchanged estimate, and acceptance by the source 80% threshold. Correction requires original error above 5 bpm, error reduction of at least 5 bpm, and acceptance. Rejection requires threshold rejection; harmful correction requires acceptance and added error above 1 bpm. Equality uses absolute tolerance $10^{-6}$ bpm. Examples are selected closest to their category median for original error, correction gain, risk, or added error, respectively. Ties are ordered by participant identifier and time. Candidate counts are 2,415, 1,946, 9,740, and 145, respectively.

Selection rules were recorded in FIGURE_CONTRACT.md before drawing the examples, not preregistered before experimentation. ECG-based event identification serves retrospective explanation and changes neither inference nor thresholds. The four categories do not exhaustively partition all windows, and windows are not independent samples. Fig. 2 retains the selected retention, correction, and rejection events, showing raw PPG/ECG from −8 to 16 s and seven window-end heart rate points. This crop shortens the display without reselecting events or changing predictions or aggregate results. Full context for all four events, including harmful correction, remains in source data and Table S8. Lines break at invalid references, without smoothing or waveform reconstruction. CASE_AUDIT.json and case_selection.csv record selection; Fig3_signals contains raw segments and timestamp mappings.

The Retain, Correct, and Reject waveforms in the Abstract figure come from the corresponding real GalaxyPPG examples in Fig. 2, using Deep PPG. Each shows only raw PPG from the central 0–8 s decision window, preserving amplitudes and samples. No waveform repair or visual selection is applied. The waveforms represent evidence for the corresponding decisions. Correct refers to changing the heart rate estimate, not reconstructing PPG; $h_0$ and $h^*$ denote the original and candidate heart rates.

**S10. Scope of Prototype Wearable Evaluation**

The study includes static, walking, running, and recovery-rest recordings from one participant, scored offline against independent ECG. These recordings are separate from GalaxyPPG and the other public datasets. Fig. 5 shows prototype results, whereas Fig. 2 retains the GalaxyPPG Deep PPG examples. Section S20 details acquisition, input processing, frozen configurations, and evaluation.

The wearable acquires signals in real time and transmits them to a smartphone running a selected estimator and the correction framework. The computer backend analyzes saved recordings using standardized preprocessing and multiple estimators. Preprocessing was informed by earlier recordings from the same participant. Although auxiliary weights and thresholds were not refitted to the new ECG measurements, this remains a retrospective evaluation after development. Its batch statistics do not directly measure the end-to-end performance of the deployed smartphone configuration.

Overlapping windows and repeated source configurations are not independent participants and were not used to estimate population confidence intervals. The prototype recordings were collected from the first author for deployment and technical feasibility evaluation. No institutional ethics approval or exemption determination was obtained for these author self-recordings. These self-recordings alone do not establish generalization across participants or clinical validity.

## S11. M1 Pulse Periodicity and Harmonic Evidence

### S11.1 Structure and Purpose

M1 is a deterministic signal analysis module that extracts fundamental, half-frequency, and double-frequency evidence for each heart rate hypothesis. For example, a 120 bpm hypothesis corresponds to 2 Hz, with additional power checks at 1 and 4 Hz. This representation helps distinguish fundamental and harmonic components. Candidate selection also uses motion, history, and relative-error evidence.

At each 4 s update, M1/M2 extract evidence from the current 8 s signal segment and a continuous 24 s signal segment with the same endpoint. The longer segment covers the union of five consecutive 8 s windows separated by 4 s (8 + 4 × 4 = 24 s); it is not an average of five separately computed outputs. This raw-signal context differs from the five-position feature memory in M3. Because each stored feature position may itself use 24 s of signals, five feature positions span 16 s between endpoints and can require 40 s of raw-signal history, as described in S13.

### S11.2 Base Spectra and Autocorrelation

A signal of length $N$ is linearly detrended, multiplied by a Hann window $w[n]$, and transformed using a 4096-point fast Fourier transform (FFT). Power interpolated onto the heart rate grid is denoted by

$$P_L(g) = |FFT_{4096}\{w[n]detrend(s_L[n])\}(g/60)|^2 \qquad \text{(S1)}$$

$$\overline{P}_L(g) = \frac{P_L(g)}{max_{u\in\mathcal{G}}P_L(u)+\varepsilon}. \qquad \text{(S2)}$$

Here, $L \in \{8,24\}$. The implementation floors the denominator instead of explicitly adding $\varepsilon$; the equation is shorthand. Spectral features are then transformed by

$$\ell(z) = \frac{\log(1+100z)}{\log 101} \qquad \text{(S3)}$$

to compress their dynamic range. ACC base spectra are normalized separately by axis, then combined by taking the maximum across axes at each frequency.

After bandpass filtering and mean removal, normalized PPG autocorrelation is calculated as

$$R_L(\tau) = \frac{\sum_n \tilde{s}_L[n]\tilde{s}_L[n+\tau]}{\sum_n \tilde{s}_L[n]^2}, A_L(g) = Interp\left(R_L, \frac{64\times 60}{g}\right). \qquad \text{(S4)}$$

The common feature sampling rate is 64 Hz. Normalized autocorrelation measures waveform similarity at the lag corresponding to a heart rate period. The implementation uses an FFT and normalizes by zero-lag correlation.

### S11.3 Harmonic Observation Range

The posterior grid covers 35–220 bpm, while harmonic observations span 17.5–440 bpm. Thus, second harmonics of high heart rate candidates are not truncated at 220 bpm. Original estimates and midpoint candidates derived from them are not constrained to the posterior grid.

## S12. M2 Motion Coherence and Residual Spectral Evidence

### S12.1 Structure and Purpose

M2 is a deterministic spectral module. It uses ACC–PPG coherence to characterize motion-related interference and passes both original and residual spectra to subsequent models. It does not reconstruct time-domain artifact waveforms.

### S12.2 Coherence and Residual Power

Let segmented Fourier coefficients be $P_m(f)$ and $A_{m,d}(f)$, where $d = 1,2,3$ indexes ACC axes and $m$ indexes overlapping segments:

$$S_{pp}(f) = \frac{1}{M}\sum_m |P_m(f)|^2, S_{aa,d}(f) = \frac{1}{M}\sum_m |A_{m,d}(f)|^2, S_{pa,d}(f) = \frac{1}{M}\sum_m P_m(f)A_{m,d}(f)^*. \quad \text{(S5)}$$

$$C_d(f) = clip\left(\frac{|S_{pa,d}(f)|^2}{S_{pp}(f)S_{aa,d}(f)+10^{-20}}, 0, 1\right), C(f) = max_d C_d(f). \quad \text{(S6)}$$

$$P_{\mathrm{res}}(f) = S_{pp}(f)[1 - C(f)]. \quad \text{(S7)}$$

High coherence indicates a frequency-specific association between PPG and at least one acceleration axis. Residual power attenuates the contribution at that frequency. This quantity provides motion-related evidence rather than an artifact-free reference spectrum. Pulse components can also be attenuated when pulse and motion frequencies overlap.

For each $r \in \{1, 1/2, 2\}$, four features are extracted at the frequencies corresponding to $rg$:

$$\psi_{L,r}(g) = \left[\ell(u_{L,r}(g)),\ C_{L,r}(g),\ \ell(v_{L,r}(g)),\ [1 - C_{L,r}(g)]u_{L,r}(g)\right], \quad \text{(S8)}$$

Here, $u$ is PPG power normalized by its maximum over 17.5–440 bpm, and $v$ is residual power normalized by its own band maximum. The fourth feature uses $u$ without logarithmic compression.

The two time scales, three frequency ratios, and four features produce 24 channels. These channels include PPG power and interactions with motion, so M1 and M2 have partly overlapping roles.

### S12.3 Complete Ordering of the 30 Features

Table S1 lists the 30 feature channels in the zero-based order used by the implementation.

**Table S1. Physiological evidence channels**

| Channel | Definition |
|---|---|
| 0, 1 | 8 s PPG base spectrum and maximum ACC base spectrum across axes, compressed using $\ell$ |
| 2, 3 | Corresponding 24 s base spectra |
| 4, 5 | 8 s and 24 s PPG autocorrelation |
| 6–9 | 8 s, fundamental $g$: $\psi$ |
| 10–13 | 8 s, half-frequency $g/2$: $\psi$ |
| 14–17 | 8 s, double-frequency $2g$: $\psi$ |
| 18–21 | 24 s, fundamental $g$: $\psi$ |
| 22–25 | 24 s, half-frequency $g/2$: $\psi$ |
| 26–29 | 24 s, double-frequency $2g$: $\psi$ |

Thus, each window has $X_t \in \mathbb{R}^{186\times 30}$. The 4096-point full-window FFT for base spectra and the 2048-point segmented FFT for coherence are separate computations, not a single spectral estimator.

## S13. M3 Five-Position Memory and Posterior Network

### S13.1 Network Structure and Tensor Dimensions

The same small multilayer perceptron (MLP) is applied independently to each heart rate hypothesis. There is no convolution across frequencies; hypotheses compete through the final softmax. History uses channel-wise diagonal decay without additional gates or attention.

### S13.2 Exact Historical-State Recurrence

Let $X^{(0)}, \ldots, X^{(4)}$ index five positions from oldest to newest, including the current window. Channel decay is defined by

$$\alpha_c = \sigma(d_c), c = 1, \ldots, 30. \quad \text{(S9)}$$

$$S^{(0)} = X^{(0)}, S^{(m)} = \alpha \odot S^{(m-1)} + (1-\alpha) \odot X^{(m)}, m = 1, \ldots, 4. \quad \text{(S10)}$$

The final state is $S_t = S^{(4)}$. For a complete five-position history, the expanded recurrence is

$$\begin{aligned} S_t &= \alpha^4 \odot X_{t-4} \\ &+(1-\alpha) \odot (\alpha^3 \odot X_{t-3} + \alpha^2 \odot X_{t-2} \\ &+\alpha \odot X_{t-1} + X_t). \end{aligned} \quad \text{(S11)}$$

Weights sum to 1 within each channel. Initially, $d_c = \log 4$, so $\alpha_c = 0.8$, giving five-position weights $(0.4096, 0.1024, 0.128, 0.16, 0.2)$. The oldest position includes the initialization contribution; recurrence is limited to these five positions.

The historical state is recomputed from the most recent five positions. At the start of a recording, the earliest available features are repeated, and input validity is checked at every position. Positions are 4 s apart, spanning 16 s between endpoints. Combined with the 24 s evidence window at each position, the earliest signal is 40 s before the current endpoint. The smartphone retains the current 24 s of raw signals and features for five positions.

### S13.3 Energy Adjustment and Posterior

$$z_t(k) = [X_t(k), S_t(k), X_t(k) - S_t(k)] \in \mathbb{R}^{90}, \quad \text{(S12)}$$

$$u_t(k) = W_2 SiLU(W_1 z_t(k) + b_1) + b_2, E_t(k) = 6\tanh(u_t(k)/6), \quad \text{(S13)}$$

$$T = 0.5 + 3\sigma(\theta), q_t(k) = \frac{\exp\{\log[max(\overline{p}_t(k), 10^{-10})]/T + E_t(k)\}}{\sum_l \exp\{\log[max(\overline{p}_t(l), 10^{-10})]/T + E_t(l)\}}. \quad \text{(S14)}$$

$\overline{p}$ is the mean over prior channels. The evaluated input has only one channel, so $\overline{p} = p$. Each instance uses a single core estimator.

$T \in (0.5, 3.5)$ controls distribution sharpness, while $E \in (-6, 6)$ adjusts support for each heart rate hypothesis. Equivalently,

$$q_t(k) \propto p_t(k)^{1/T} \exp E_t(k). \quad \text{(S15)}$$

Initialization uses $\theta = \log 0.2$, giving $T = 1$. The final linear layer is initialized to zero, so the initial network approximately preserves the prior. The term posterior describes this distribution but does not imply guaranteed statistical calibration.

The parameter count is $(90 \times 24 + 24) + (24 + 1) + 30 + 1 = 2240$. Memory decay, the MLP, and temperature are learned jointly.

### S13.4 Construction of Nine Candidates

Let $Q_\beta$ be the first grid value at which the discrete posterior CDF reaches $\beta$. Candidates follow the implementation order:

$$\mathbf{a}_t = [h_{0,t}, \sum_k q_t(k) g_k, g_{argmax_k q_t(k)}, Q_{.1}, Q_{.25}, Q_{.5}, Q_{.75}, Q_{.9}, (h_{0,t} + Q_{.5})/2]. \quad \text{(S16)}$$

These are the original estimate, mean, mode, five percentiles, and the midpoint between the original estimate and median. Duplicates are allowed. The original estimate occupies position 0 to give retention priority in ties.

## S14. M4 Consistent Gain–Harm Decisions

### S14.1 Posterior Expectations and Learned Error Predictions

The posterior provides each candidate's expected absolute error, expected added error, and neighborhood support:

$$e_j^q = \sum_k q(k)|a_j - g_k|, \quad \text{(S17)}$$

$$h_j^q = \sum_k q(k) max\{|a_j - g_k| - |h_0 - g_k|, 0\}, r_j^q = \sum_k q(k)\mathbf{1}(|a_j - g_k| \le 5). \quad \text{(S18)}$$

Posterior expected net gain is $g_j^q = e_0^q - e_j^q$. These quantities are calculated under q. M4 separately learns $\hat{G}_j, \hat{H}_j, \hat{E}_j$ using actual source-participant errors as supervision. Eligibility requires both sufficient learned net gain and posterior expected error below that of the original estimate.

### S14.2 Complete 106-Dimensional Feature Vector

Let $\delta_j = a_j - h_0$, with posterior mean $\mu_q$, standard deviation $\sigma_q$, and entropy $\mathcal{H}_q$. The first 16 features are

$$\begin{aligned} f_j^{16} = [&a_j/200, h_0/200, \delta_j/40, |\delta_j|/40, \\ &e_j^q/20, h_j^q/20, r_j^q, e_0^q/20, r_0^q, \\ &(e_0^q - e_j^q)/20, r_j^q - r_0^q, \sigma_q/40, \mathcal{H}_q/6, \\ &\mu_q/200, \mathbf{1}(\delta_j < 0), \mathbf{1}(|\delta_j| > 20)]. \end{aligned} \quad \text{(S19)}$$

For candidate $a_j$, the nearest grid position is $k_j = argmin_k |a_j - g_k|$. The remaining 90 features are

$$f_j^{90} = [X_t(k_j), S_t(k_j), X_t(k_j) - X_t(k_0)]. \quad \text{(S20)}$$

The final vector is $f_j = [f_j^{16}, f_j^{90}] \in \mathbb{R}^{106}$. Its third 30-dimensional block is the current-evidence difference between candidate and original estimate. In contrast, M3 uses a temporal difference between current and historical evidence. For out-of-grid candidates, evidence comes from the nearest endpoint, but the candidate heart rate remains unchanged.

Features are standardized using source-training statistics:

$$\tilde{f}_{j,c} = clip\left(\frac{f_{j,c}-\mu_c}{max(\sigma_c,0.05)}, -5,5\right). \quad \text{(S21)}$$

Means and standard deviations are calculated across training windows and all nine candidates, then reused unchanged during external inference.

### S14.3 Two-Branch Network

Each branch consists of one affine transformation. Softmax, softplus, and constrained decoding introduce the nonlinear operations. The second branch predicts a single base error for each window, shared by all nine candidates.

The parameter count is $(106 \times 3 + 3) + (106 + 1) = 428$.

### S14.4 Exact Consistent Decoding

$$\boldsymbol{\pi}_j = softmax(W_h\tilde{f}_j + b_h), b = 20softplus(w_r^{\top}\tilde{f}_0 + b_r), d_j = |a_j - h_0|. \quad \text{(S22)}$$

Let $\pi_{j,1}, \pi_{j,2}, \pi_{j,3}$ denote the three allocations. The implementation decodes them as

$$\hat{B}_j = min(d_j\pi_{j,1}, b), \quad \text{(S23)}$$

$$\hat{H}_j = max\big(d_j\pi_{j,2},\ max(d_j - 2b + \hat{B}_j, 0)\big), \quad \text{(S24)}$$

$$\hat{G}_j = \hat{B}_j - \hat{H}_j, \hat{E}_j = b - \hat{G}_j. \quad \text{(S25)}$$

The three logits are allocation parameters, not independent retention, correction, and rejection probabilities. The third component participates through the softmax sum constraint but is not returned directly. $\hat{B}$ is an internal beneficial component; the returned gain is net gain after subtracting added error.

The decoding satisfies

$$\hat{E}_j \geq 0, |\hat{G}_j| \leq d_j, b + \hat{E}_j \geq d_j, \hat{E}_j - b = -\hat{G}_j. \quad \text{(S26)}$$

because absolute-error distances obey the triangle inequality:

$$||a_j - y| - |h_0 - y|| \leq |a_j - h_0|, |a_j - y| + |h_0 - y| \geq |a_j - h_0|. \quad \text{(S27)}$$

Embedding these geometric relationships prevents inconsistent predictions, such as claiming a 30 bpm improvement after changing an estimate by only 2 bpm. For the original estimate $d_0 = 0$, the identities $\hat{B}_0 = \hat{H}_0 = \hat{G}_0 = 0$ and $\hat{E}_0 = b$ necessarily hold.

Decoding does not enforce $\hat{H}_j = max(-\hat{G}_j, 0)$. Beneficial and harmful components can both be positive, with their difference giving net gain. This joint prediction parameterization differs from the definition of actual added error for an individual training sample.

### S14.5 Primary Decision Rule

For $j > 0$, eligibility requires

$$I_j = \mathbf{1}\big(\hat{G}_j \geq 1\big)\mathbf{1}\big(\hat{H}_j \leq 2\big)\mathbf{1}\big(e_0^q - e_j^q > 0\big). \quad \text{(S28)}$$

$$U_0 = 0, U_j = \begin{cases} \hat{G}_j - 2\hat{H}_j, & I_j = 1, \\ -\infty, & I_j = 0, \end{cases} j^* = argmax_{j=0,\dots,8} U_j. \quad \text{(S29)}$$

$$\hat{h} = a_{j^*}. \quad \text{(S30)}$$

The primary configuration, identified as FourthUnbounded+agreement in the analysis files, applies no 20 bpm correction cap. The maximization returns the first candidate attaining the highest utility. Because the original estimate is listed first, a tie at zero utility retains that estimate. A replacement must therefore satisfy all three eligibility conditions and have positive utility.

For illustration only, consider an original estimate of 140 bpm and a candidate of 100 bpm; these are not experimental observations. With $\hat{G} = 8, \hat{H} = 1, g^q = 5$, utility is 6 and the candidate can compete. With $\hat{G} = 2, \hat{H} = 1.5, g^q = 5$, all three conditions hold but utility is $-1$, so the original estimate is retained. The added-error limit of 2 bpm applies to a prediction; actual error can still increase by more than 2 bpm.

## S15. Independent Output Risk Gate

M4 selects $\hat{h}$, and the risk gate then determines acceptance. Each configuration compares original-estimator risk $r_0$ with learned candidate error $\hat{E}_{j^*}$ on three source-calibration participants. Selection uses participant-macro-averaged MAE80 for the uncapped variant without agreement checking, with ties favoring original risk. The selected family is then used by all four M4 decision strategies.

The rejection risk for the proposed method is therefore

$$r_t = \begin{cases} r_{0,t}, & \text{raw risk selected,} \\ \hat{E}_{t,j^*}, & \text{learned risk selected.} \end{cases} \qquad \text{(S31)}$$

The base error $b$ is not itself the final threshold, and not every configuration rejects according to learned error. When the selected family is raw, original-estimator risk remains in use even after heart rate correction.

For a frozen threshold $\tau$,

$$Accept_t = \mathbf{1}(\hat{h}_t \text{ finite})\mathbf{1}(r_t \text{ finite})\mathbf{1}(r_t \le \tau). \qquad \text{(S32)}$$

For the source 80% threshold, the implementation calculates $m = \lfloor 0.8 N_{\text{ref}} \rfloor$ from the complete reference-eligible count $N_{\text{ref}}$, sorts finite available risk scores, and selects value $min(m, N_{\text{avail}})$. If no position is available, a rejection-sentinel threshold is used. Ties, missing outputs, and distribution shifts can move actual coverage away from 80%. A separate source MAE≤5 bpm rule requires at least 10% coverage per calibration participant and allows zero output if no solution is feasible.

When external auxiliary history is unavailable, restoration to the full timeline may retain the core estimate but set auxiliary risk to NaN. This does not pass the rejection gate. All-available-output evaluation and fixed-risk-gate evaluation therefore use different output sets.

## S16. Two-Stage Training and Losses

### S16.1 M3 Distribution Target

Reference heart rate is converted to a Gaussian soft target with a standard deviation of 3 bpm:

$$\tilde{y}_k = \frac{\exp[-(g_k - y)^2/(2\cdot 3^2)]}{\sum_l \exp[-(g_l - y)^2/(2\cdot 3^2)]}. \qquad \text{(S33)}$$

Let $F_q(k) = \sum_{l \le k} q_l$ and $F_y(k) = \sum_{l \le k} \tilde{y}_l$. The implemented loss is

$$\mathcal{L}_{\text{post}} = \frac{1}{B} \sum_t \left[ -\sum_k \tilde{y}_{t,k} \log q_{t,k} + \frac{0.2}{20} \sum_k \left( F_{q_t}(k) - F_{y_t}(k) \right)^2 \right]. \qquad \text{(S34)}$$

The first term favors probability near the reference, while the cumulative-distribution term incorporates distance on the heart rate grid. This is not a raw-waveform reconstruction loss.

M3 is trained for 2,000 steps using AdamW, a learning rate of 0.002, weight decay of 0.03, and a batch size of 96. The gradient norm is clipped at 2, and sampling is balanced across participants. With probability 0.1, each sample's prior channel is replaced by the prior from another sampled window to reduce reliance on the prior.

### S16.2 M4 Supervision

For the original estimate and candidates in the same window,

$E^*_{t,j} = |a_{t,j} - y_t|, G^*_{t,j} = E^*_{t,0} - E^*_{t,j}, H^*_{t,j} = max(-G^*_{t,j}, 0)$.(S35)

Let $\hat{v} = (\hat{G}, \hat{H}, \hat{E})/20$ and $v^* = (G^*, H^*, E^*)/20$. Then

$$\mathcal{L}_{\text{decision}} = mean[\quad SmoothL1(\hat{v} - v^*)] + 2mean\left[max(v^*_H - \hat{v}_H, 0)^2\right]. \tag{S36}$$

PyTorch's default SmoothL1 is $\rho(u) = u^2/2$ when $|u| < 1$, and $|u| - 1/2$ otherwise. The first term averages over batches, candidates, and three outputs; the second averages over batches and candidates. The additional term penalizes underestimation of actual added error.

The weights and biases of both M4 linear layers are initialized to zero. Training uses AdamW with a learning rate of 0.003, weight decay of 0.1, batch size of 256, and participant-balanced sampling. The model is trained for 2,000 steps. M3 posteriors are computed and fixed beforehand; the M4 loss does not update M3 or the core estimator.

### S16.3 Rationale for Participant Cross-Fitting

Training M4 on predictions from the same participants used to fit M3 could underestimate replacement risk. Cross-fitting instead generates each participant's posteriors using an M3 model trained without that participant. Historical decay parameters are taken from the corresponding M3 model.

Each estimator, outer fold, seed, and trained variant requires three internal M3 models, one final M3 model, and one M4 model. Ten estimators, five folds, three seeds, and four variants yield 600 configurations, comprising 2,400 M3 fits and 600 M4 fits. Each external participant is evaluated with 15 source configurations. Their metrics are averaged; their predictions are not combined into an ensemble.

The weights from training step 2,000 are used without checkpoint selection on the evaluation set.

## S17. Decoding Constraints and Information Limits

### S17.1 Proof of Geometric Consistency

Omitting window and candidate indices, let $d = |a - h_0| \geq 0$ and $b \geq 0$. Softmax ensures $\pi_i \geq 0$ and $\sum_i \pi_i = 1$. From $B = min(d\pi_1, b)$, it follows that $0 \leq B \leq min(d, b)$. Therefore,

$$d\pi_2 \leq d - d\pi_1 \leq d - B, d - 2b + B \leq d - B. \tag{S37}$$

Because $0 \leq d - B$, maximum-based decoding gives $0 \leq H \leq d - B$. Hence,

$$-d \leq B - H = G \leq d, E = b - B + H \geq 0. \tag{S38}$$

Using $H \geq d - 2b + B$, we further obtain

$$b + E = 2b - B + H \geq d. \tag{S39}$$

For the original estimate, $d = 0$, so $B = H = G = 0$ and $E = b$. These relationships constrain the algebra of predicted quantities. They do not equate predictions with actual errors or guarantee calibration or a true harm bound. The harm branch is not forced to equal $max(-G, 0)$; it receives separate supervision from actual added error.

### S17.2 Why Estimators Using Spectral Information May Still Benefit

Deterministic spectral, harmonic, and difference features add no information to the raw signals, but they can make useful structure more accessible to finite-capacity models. The training objective also changes from a single heart rate or distribution to paired gain, added error, and absolute error relative to a fixed estimate. Candidate selection then applies an explicit cost. Spectral processing in a core estimator is therefore compatible with additional correction gains. Those gains cannot be attributed solely to recomputing spectra.

Complementarity is not guaranteed. Let $\mathcal{I}$ represent all information shared by the estimator and auxiliary layer. If the estimator is already Bayes-optimal for absolute error in the same target domain,

$$h^*(\mathcal{I}) \in argmin_h \mathbb{E}[|h - Y| \mid \mathcal{I}], \quad \text{(S40)}$$

deterministic post-processing using the same information cannot further reduce conditional expected absolute error. Practical finite models need not satisfy this ideal condition, and the study also considers trade-offs involving harmful correction and coverage. This statement defines a theoretical boundary; it does not establish Bayes optimality for any evaluated estimator.

The experiments show average error reductions, gains from correction on the same accepted windows, and effects of the specified ablations. They do not isolate the separate contributions of representation, paired supervision, and constrained decoding. Controlled comparisons would be needed to attribute these gains to individual mechanisms.

## S18. Roles and Interfaces in the Agent Architecture

The agent architecture organizes signal analysis, a limited historical state, and output decisions around a fixed heart rate estimator. Each of the ten estimators is evaluated separately in the core-model role. The system uses neither a language model nor collaboration among estimators. Auxiliary parameters are fitted separately for each estimator and remain fixed during external evaluation.

**Table S2. Module interfaces and outputs**

| Role | Implementation | Output or function |
|---|---|---|
| Core model | One registered heart rate estimator | Original estimate, adapted prior, and original risk |
| Tools | M1 pulse analysis and M2 motion analysis | 30-dimensional evidence at 186 heart rate hypotheses |
| Memory | M3 limited history and posterior network | Five-position evidence state, posterior, and nine candidates |
| Decision | M4 evaluation and explicit selection rules | Shared original error, candidate gains and harms, and estimate selection |
| Delivery gate | Independent risk gate | Release or reject the selected estimate |

Tools execute in a fixed sequence, and some M1/M2 interaction channels share PPG and ACC information. M3 states derive from input evidence, not released heart rates, ECG feedback, or generated text. M4 directly reads tool and memory features to compare candidates with the original estimate. Rejection follows selection, so both retained and corrected estimates can subsequently be rejected. Training uses no reinforcement learning. M4 loss does not propagate to M3 or the core estimator, and its three-way softmax does not represent action probabilities.

Fig. 1 shows the overall information flow through the fixed core estimator and M1–M4. Table S2 summarizes the module interfaces. M1/M2 use 8 s and 24 s contexts at 64 Hz and have no trainable weights. M3 uses five positions, including the current window, and shares network parameters across 186 heart rate hypotheses. M4 has 428 trainable parameters, with risk checking after candidate selection. The architecture includes no iterative feedback, dynamic tool routing, or additional training stage. Motion and history ablations support some tool and memory contributions. Prototype recordings also include an M4-removal comparison with the same candidates (S20; Fig. S16). Independent ablation of all harm constraints and online-feedback experiments remain unvalidated.

## S19. Mathematical Definitions of Outputs and Metrics

This section defines final outputs and evaluation metrics. The acceptance indicator $A_t$ is defined in main-text (6), and the selected heart rate in (5). The final output is

$$h_t^{\text{out}} = \begin{cases} \hat{h}_t, & A_t = 1, \\ \text{missing}, & A_t = 0. \end{cases} \qquad \text{(S41)}$$

When $A_t$=1, the selected estimate is released. When $A_t$=0, output is missing and no heart rate is reported. Missing outputs are excluded from the accepted set but remain reference-eligible. Let $\mathcal{R}$ contain all reference-eligible windows, $\mathcal{A}$ the accepted windows with finite outputs, and $\mathcal{A}_0$ those also having finite original estimates. MAE, coverage, and harmful correction rate are

$$MAE(\hat{h},\mathcal{A}) = \frac{1}{|\mathcal{A}|}\sum_{t\in\mathcal{A}} |\hat{h}_t - y_t|, Coverage = \frac{|\mathcal{A}|}{|\mathcal{R}|}. \qquad \text{(S42)}$$

$$HarmRate = \frac{1}{|\mathcal{A}_0|}\sum_{t\in\mathcal{A}_0} \mathbf{1}\{|\hat{h}_t - y_t| - |h_{0,t} - y_t| > 1\}. \text{(S43)}$$

MAE is measured in bpm. Coverage uses all reference-eligible windows as its denominator, whereas harmful correction uses paired accepted windows. The threshold in (S43) is an absolute-error increase exceeding 1 bpm, not a heart rate change exceeding 1 bpm. Severe added error counts accepted windows with an error increase exceeding 10 bpm, divided by all reference-eligible windows. With an empty accepted set, MAE remains undefined and coverage is zero. Section S8 describes aggregation, paired decomposition, and statistics.

## S20. Single-Participant Acquisition and Offline Evaluation

### S20.1 Recording Audit and Reference Timeline

On October 3, 2026, participant PILOT01 completed separate static, walk, run, and rest recordings; rest denotes recovery after exercise. Raw wrist logs contain dual-wavelength PPG, ACC, gyroscope signals, frame counts, and reception times. Independent upper-sternum ECG provides the reference. Fourteen raw files comprise wrist logs, ECG logs, and compressed exports for the four recordings, plus an earlier communication-test log pair. The earlier test is excluded from scoring. Archive integrity and decoded ECG consistency were audited; duplicate synchronization files are not independent recordings. Table S11 reports durations, reception rates, and eligible-window counts.

Repeated ECG timestamps are reconstructed from batch-end reception times at 512 Hz. Reference heart rates retain the XQRS/GQRS dual-detector quality checks and reference_qualified flags. Window-level detector outputs, valid RR fractions, sampling coverage, and exclusions are stored in ecg_reference_windows.csv. PPG/ACC use the same recording clock, without an ECG-error-minimizing delay search. Reconstruction does not validate fixed interdevice transmission delays. Evaluation uses 8 s windows with a 4 s stride, requiring endpoints at or after 40 s and eligible references. There are 324 eligible windows, with dependence due to overlap.

PPG reception gaps exceeding 0.4 s split recordings into blocks, with 1 s boundary guards at both ends. Memory does not cross recordings or gaps. Auxiliary evidence needs 24 s of context, and five complete historical positions need approximately 40 s. When auxiliary quantities are unavailable but the core estimate is finite, all-available-output evaluation retains the original estimate. The matched_aux_full and full_history_only analyses separately restrict auxiliary availability and complete history; both are included in configuration-level sensitivity results.

**S20.2 Dual-Channel Preprocessing and Input Units**

Infrared and red channels are processed separately. Each block undergoes linear detrending, root-mean-square (RMS) normalization, and third-order zero-phase bandpass filtering at 0.5–5 Hz. Triaxial ACC supplies motion references for continuous normalized least mean squares (NLMS) filtering, with 12 taps per axis, step size 0.1, and leakage $10^{-4}$. The residual is used when the motion-intensity ratio over up to 24 s exceeds 0.1; otherwise, the bandpass output is retained. Block baseline and RMS are then adjusted. Processing uses a 12.5 Hz grid before adaptation to source-model sampling rates; auxiliary evidence uses 64 Hz. Actual reception rates of 9.67–11.19 Hz differ from this grid, and upsampling cannot recover unrecorded bandwidth. Zero-phase filtering and whole-block normalization make the pipeline retrospective. Preprocessing was informed by earlier exploration of the same participant, so this is post-development retrospective evaluation.

Infrared is the prespecified primary channel, with red analyzed separately. ECG does not select channels by window, and the ten estimators are not fused. Each estimator and its framework receive identical processed signals. TimePPG input adaptation multiplies PPG by the source-training median amplitude 45.349861 and divides ACC by 2142.539615, the median magnitude during the first 60 s of the static recording. These constants apply to both channels and all four states, without ECG-error tuning. This is numerical scale adaptation, not physical calibration. Unadapted outputs are retained separately for diagnosis. Fig. S18 shows fixed pre- and post-processing segments; regular-looking waveforms do not establish complete recovery of physiological components.

**S20.3 Frozen Models and Figure Configurations**

Each registered estimator reuses five source folds and seeds 219, 220, and 221, giving 15 frozen auxiliary configurations. Infrared includes full, no_harmonic, no_motion, and no_memory; red includes full only. This gives 750 estimator/channel/configuration combinations corresponding to 600 unique auxiliary parameter sets. FROZEN_MODEL_BINDINGS.json and file hashes bind checkpoints, M4 parameters, and calibration thresholds. Metrics are calculated for each configuration before arithmetic averaging; predictions are not combined into an ensemble. KID-PPG retains unlabeled filter fitting for 1,000 steps per recording/channel, totaling 8,000 steps, without updating the heart rate network. Local implementations and input adapters follow the source protocol and do not represent the best official implementations.

Fig. 5(d) fixes Deep PPG, infrared input, source fold 0, seed 219, and the unchanged source80 threshold. It shows all evaluation windows from 40 s onward in static, walking, and running recordings. Rejected outputs remain missing; crosses mark internal candidates only. For Fig. S11, TimePPG is the first estimator in the prespecified order with all three actions. Using the same fold and seed, each action is illustrated by its first chronological window, without selection for favorable ECG errors. The deployed research version of PulseLink 0.3.0 uses Welch with the full framework, source fold 0, seed 219, and a source 80% coverage threshold of 9.860092 for real-time analysis. The backend evaluates all ten methods and source configurations, calculates paired statistics, and generates figures. Fig. 5(c) shows the smartphone interface and workflow. Smartphone configuration and backend multi-method results are recorded separately; ten predictions are not fused into a phone output.

### S20.4 Comparison Sets and Aggregation

The available_full mode applies no additional rejection and compares original and corrected outputs on the same available windows within each estimator. Most methods have 307/324 windows; BeliefPPG and Deep PPG have 289/324, and KID-PPG has 304/324. The common_ten_full mode uses 289 infrared windows available to all ten estimators. Eligible windows from all four states are pooled to calculate each estimator's MAE before equal weighting across estimators. State-specific results are calculated separately. Cross-estimator macro-averages describe neither a single deployed estimator nor population effects.

The own_rank80 mode ranks risk within target recordings and accepts at most 80% of reference-eligible windows, without ranking ECG errors. Each method accepts 259/324 windows overall. Same-risk rejection uses the full framework's accepted set but preserves original heart rates. Correction gain compares original and corrected errors on that set. Selection gain compares original-estimate errors on sets selected by original and framework risk. Their sum equals total gain within each configuration. The source80 and source_mae5 modes reuse source-calibrated thresholds without target-ECG reselection. Source_mae5 strategies are calibrated independently, so their accepted sets need not match.

A harmful correction increases absolute error by more than 1 bpm relative to the original estimate. Its rate uses paired output windows as the denominator. Initially accurate estimates have an original absolute error no greater than 5 bpm; deterioration rates are calculated within that subset. The severe added-error proportion counts error increases exceeding 10 bpm relative to all reference-eligible windows. Overall summaries weight estimator-specific rates equally. For Fig. S20(b), method–window counts are pooled within each configuration before calculating rates. This avoids undefined rates for individual state–method pairs with no initially accurate windows, but differs from averaging estimator-specific rates. A zero denominator is reported as NA.

Error bars and shading show the minimum–maximum range across 15 frozen source configurations, not participant-bootstrap confidence intervals. Folds, seeds, wavelengths, and correlated windows are not additional participants. MAE is undefined for zero-output configurations. Source-threshold coverage averages include all 15 configurations, while MAE averages include only nonempty configurations; zero-output counts are reported separately.

### S20.5 Results and Interpretation

Without additional rejection, paired macro-averaged MAE decreased from 9.087 to 4.445 bpm. Excluding TROIKA gave a decrease from 7.102 to 4.454 bpm. Restricting all ten estimators to 289 common windows gave a decrease from 9.235 to 4.332 bpm. Across ten estimators, MAE decreased from 14.811 to 4.241 bpm during static recording, from 8.544 to 5.210 bpm during walking, and from 9.997 to 5.968 bpm during running. The large original TROIKA errors contributed to the static result; excluding TROIKA gave a decrease from 7.716 to 4.148 bpm. Table S12 reports estimator-specific results. Fig. S15 presents both channels and all four states; Fig. S19 presents state-specific sensitivity and matched-coverage comparisons.

At 80% target coverage, macro-averaged MAE was 8.598 bpm for original-risk selection, 7.492 bpm for same-risk rejection, 3.298 bpm for the framework, and 6.826 bpm for ridge rejection. Correction and selection contributed approximately 4.195 and 1.105 bpm, respectively, to the total reduction of 5.300 bpm. The decomposition compares numerical corrections on fixed windows and selection effects on unchanged estimates. It is descriptive and does not identify independent causal effects of individual modules.

Adding M4 reduced the harmful correction rate from 2.175% to 0.416% and the deterioration rate among initially accurate estimates from 2.644% to 0.516%. However, MAE increased from 2.574 to 4.445 bpm. These comparisons used all available outputs without additional rejection. The effect varied by estimator: KID-PPG had a harmful correction rate of 0.395% with M4 and 0.132% without it. After pooling initially accurate method–window counts by state, deterioration rates decreased from 1.040%, 6.000%, and 3.022% to 0.776%, 0.324%, and 0.954% for static, walking, and running recordings, respectively. Some individual estimators showed changes in the opposite direction (Fig. S20). Thus, more conservative correction did not improve every metric or estimator.

After transfer of thresholds targeting source MAE of 5 bpm, mean MAE among configurations with outputs was 8.871 bpm for KID-PPG and 8.244 bpm for TimePPG. ACC-NLMS, JOSS, and SpaMA produced no outputs in 7/15, 6/15, and 3/15 configurations, respectively. Source-domain calibration therefore did not guarantee the target error on prototype recordings. This retrospective single-participant evaluation is further limited by noncausal preprocessing, reconstructed timestamps, and reception bandwidth. Independent evaluation requires a fixed protocol and causal online processing.

## S21. Estimator Composition and TROIKA Sensitivity

All ten registered implementations remain in the primary analysis. Leave-one-out analyses test whether unusually large original errors from one implementation dominate the macro-average; they do not selectively remove unfavorable results. The local TROIKA variant runs at 25 Hz with custom SSA pairing and spectral tracking and is not equivalent to the authors' implementation. Its median original output reaches 292.68 bpm on Galaxy and Wild, leaving substantial scope for correction. Excluding it, the other nine estimators still reduce macro-averaged MAE by 3.41, 9.17, 5.83, and 3.08 bpm on DaLiA, Wrist, Galaxy, and Wild. Fig. S21(a) compares all-ten and no-TROIKA gains; Fig. S7 reports the complete leave-one-out analysis using the same paired calculations and participant-bootstrap intervals.

For prototype infrared recordings, ten-estimator macro-averaged MAE changes from 9.087 to 4.445 bpm, compared with 7.102 to 4.454 bpm after excluding TROIKA [Fig. S21(b)]. Each estimator retains its paired available windows without additional rejection. The common 289-window result, 9.235 to 4.332 bpm, instead tests differences in window availability and is a distinct sensitivity condition. Fig. S19 reports state-specific analyses. Gains are therefore not solely attributable to TROIKA, but macro-averages must specify which implementations they include.

# SUPPLEMENTARY TABLES

TABLE S3
ERROR, CORRECTION BENEFITS, AND HARMS WITHOUT ADDITIONAL REJECTION WITH 95% CONFIDENCE INTERVALS

| Metric | **DaLiA** | **Wrist** | **Galaxy** | **Wild** |
|---|---|---|---|---|
| Original MAE (bpm) | 13.68 [11.43, 16.12] | 45.41 [36.55, 54.75] | 38.50 [36.52, 40.86] | 39.97 [38.15, 42.19] |
| Framework MAE (bpm) | 9.23 [7.39, 11.28] | 34.61 [26.47, 45.07] | 21.38 [19.18, 24.10] | 26.08 [23.70, 28.60] |
| Beneficial corrections (%) | 16.06 [14.78, 17.48] | 30.74 [24.00, 37.45] | 36.64 [35.08, 38.27] | 26.99 [24.43, 30.04] |
| Error change within ±1 bpm (%) | 82.26 [80.75, 83.67] | 62.49 [57.10, 67.46] | 59.84 [58.43, 61.12] | 69.88 [67.03, 72.28] |
| Harmful corrections (%) | 1.68 [1.28, 2.16] | 6.77 [3.97, 10.09] | 3.52 [2.84, 4.46] | 3.12 [2.80, 3.45] |
| Severe added error (%) | 0.37 [0.16, 0.66] | 3.89 [1.92, 6.14] | 1.61 [1.14, 2.26] | 1.35 [1.17, 1.54] |

Note: Values are macro-averaged point estimates with participant-bootstrap 95% confidence intervals. DaLiA, Wrist, Galaxy, and Wild include 15, 8, 20, and 14 participants with eligible references. Three DaLiA seeds or 15 external source configurations are averaged within participant and estimator, followed by equal weighting of ten estimators and participants. The three correction categories use the same paired available-window denominator and sum to 100% before rounding. Severe added error instead uses all reference-eligible windows. Bootstrap resampling uses 10000 draws and seed 260922 with fixed model weights. Intervals reflect participant composition, not retraining uncertainty.

TABLE S4
PAIRED MAE CHANGES RELATIVE TO ORIGINAL ESTIMATES WITH 95% CONFIDENCE INTERVALS

| Dataset | Framework − original MAE (bpm) |
|---|---|
| DaLiA | -4.45 [-5.47, -3.65] |
| Wrist | -10.79 [-15.33, -5.98] |
| Galaxy | -17.11 [-18.07, -16.25] |
| Wild | -13.89 [-15.04, -12.98] |

TABLE S5
MAE CHANGE DECOMPOSITION AT 80% TARGET COVERAGE WITH 95% CONFIDENCE INTERVALS

| Dataset | Total change (bpm) | Correction (bpm) | Selection (bpm) |
|---|---|---|---|
| DaLiA | -6.26 [-7.71, -5.12] | -3.88 [-4.87, -3.07] | -2.38 [-2.95, -1.90] |
| Wrist | -12.19 [-17.51, -6.51] | -11.24 [-16.05, -6.27] | -0.95 [-1.91, -0.08] |
| Galaxy | -19.29 [-20.40, -18.27] | -17.79 [-18.85, -16.85] | -1.50 [-1.97, -1.02] |
| Wild | -15.02 [-16.37, -13.89] | -14.70 [-15.85, -13.79] | -0.32 [-0.64, -0.00] |

TABLE S6
MAE CHANGES AFTER MODULE ABLATION WITH 95% CONFIDENCE INTERVALS

| Dataset | No harmonic (bpm) | No motion (bpm) | No memory (bpm) |
|---|---|---|---|
| DaLiA | 0.61 [0.36, 0.85] | 0.97 [0.74, 1.24] | 0.87 [0.71, 1.05] |
| Wrist | 0.93 [-0.23, 2.10] | 5.98 [3.02, 8.67] | 1.52 [0.52, 2.46] |
| Galaxy | -1.62 [-1.91, -1.31] | 6.08 [5.51, 6.63] | 1.17 [0.89, 1.48] |
| Wild | -0.94 [-1.35, -0.56] | 7.08 [6.71, 7.43] | 1.35 [1.10, 1.58] |

TABLE S7
ACTUAL COVERAGE AND CONFIGURATION FAILURES WITH FIXED SOURCE THRESHOLDS

| Dataset | Source80 coverage (%) | Source MAE5 coverage (%) | MAE5 zero-output / all | MAE5 above 5 / nonempty | Above 5 (%) |
|---|---|---|---|---|---|
| DaLiA | 79.07 | 69.61 | 48/450 | 159/402 | 39.55 |
| Wrist | 42.29 | 44.87 | 183/1200 | 930/1017 | 91.45 |
| Galaxy | 45.48 | 47.24 | 365/3000 | 1310/2635 | 49.72 |
| Wild | 30.50 | 36.38 | 227/2100 | 1376/1873 | 73.47 |

TABLE S8
CENTRAL-WINDOW ERRORS IN REPRESENTATIVE EXAMPLES

| Outcome | Participant | Window start (s) | Original error (bpm) | Framework error (bpm) |
|---|---|---|---|---|
| Retention | P21 | 1412 | 1.621 | 1.621 |
| Correction | P15 | 420 | 12.104 | 2.364 |
| Rejection | P12 | 2508 | 9.382 | 9.382 (before rejection) |
| Harmful correction | P15 | 168 | 1.513 | 4.000 |

TABLE S9
MEAN MAE REDUCTIONS FOR SELECTED CORE ESTIMATORS

| Core estimator | **DaLiA** | **Wrist** | **Galaxy** | **Wild** |
|---|---|---|---|---|
| Deep PPG [3] | 1.335 | 5.222 | 2.910 | 1.215 |
| JOSS [5] | 5.756 | 15.847 | 7.560 | 5.779 |
| SpaMA [6] | 7.071 | 10.991 | 13.902 | 5.256 |
| WFPV [7] | 1.134 | 19.611 | 14.571 | 4.720 |
| KID-PPG [10] | 0.348 | 8.562 | 1.093 | 0.120 |
| BeliefPPG [9] | <0.001 (near zero) | <0.001 (near zero) | <0.001 (near zero) | <0.001 (near zero) |

TABLE S10
PARTICIPANT COUNTS AND SHARED COVERAGE FOR FIG. 3

| Dataset | Participants | Shared coverage (%) |
|---|---|---|
| DaLiA | 15 | 100.00 |
| Wrist | 8 | 98.29 |
| Galaxy | 20 | 97.81 |
| Wild | 14 | 99.34 |

Note: Original and framework estimates are compared on the same available windows. Shared coverage is a separate macro-average for each dataset. The unrounded DaLiA value is 99.9973%, displayed as 100.00% to two decimal places. These are the Fig. 3 analysis windows, not fixed-source-threshold transfer results.

TABLE S11
PROTOTYPE RECORDINGS AND EVALUATION WINDOWS

| State | Common duration (s) | Mean reception rate (Hz) | Reference-eligible windows |
|---|---|---|---|
| static | 274.46 | 9.67 | 59 |
| walk | 432.56 | 9.89 | 54 |
| run | 347.49 | 11.19 | 77 |
| rest | 575.38 | 10.28 | 134 |

Note: n = 1 participant and four separate recordings. Durations are audited common recording intervals. Counts include reference-eligible 8 s windows with a 4 s stride and endpoints at or after 40 s, totaling 324. Reception rates differ from the 12.5 Hz processing grid and 64 Hz model input sampling rate.

TABLE S12
PAIRED ALL-AVAILABLE-OUTPUT COMPARISONS FOR PROTOTYPE INFRARED RECORDINGS

| Core estimator | Paired windows | Original MAE (bpm) | Framework MAE (bpm) | Reduction (bpm) |
|---|---|---|---|---|
| BeliefPPG | 289 | 2.256 | 2.256 | 0.000 |
| KID-PPG | 304 | 12.601 | 9.887 | 2.714 |
| TimePPG | 307 | 14.484 | 9.571 | 4.913 |
| Deep PPG | 289 | 8.231 | 4.235 | 3.996 |
| Welch | 307 | 3.364 | 2.733 | 0.631 |
| ACC-NLMS | 307 | 7.049 | 3.533 | 3.516 |
| TROIKA | 307 | 26.945 | 4.366 | 22.579 |
| JOSS | 307 | 2.145 | 1.691 | 0.454 |
| SpaMA | 307 | 12.704 | 5.073 | 7.631 |
| WFPV | 307 | 1.086 | 1.104 | -0.019 |

Note: The four states are pooled without additional rejection. Each estimator’s original and framework outputs use the same available windows. Values are means over 15 frozen source configurations. Negative reductions indicate increased error. n = 1; window counts are not independent participant counts.

TABLE S13
FIXED SOURCE-THRESHOLD TRANSFER TO PROTOTYPE INFRARED RECORDINGS

| Core estimator | Source80 coverage (%) | Source MAE5 coverage (%) | Source MAE5 nonempty MAE (bpm) | Zero-output configurations |
|---|---|---|---|---|
| BeliefPPG | 81.05 | 87.35 | 2.250 | 0/15 |
| KID-PPG | 58.21 | 79.86 | 8.871 | 0/15 |
| TimePPG | 72.82 | 82.96 | 8.244 | 0/15 |
| Deep PPG | 84.44 | 83.83 | 3.490 | 0/15 |
| Welch | 85.08 | 84.61 | 1.279 | 0/15 |
| ACC-NLMS | 83.83 | 42.51 | 1.910 | 7/15 |
| TROIKA | 85.74 | 69.47 | 1.918 | 0/15 |
| JOSS | 86.13 | 50.51 | 1.252 | 6/15 |
| SpaMA | 80.58 | 56.50 | 3.171 | 3/15 |
| WFPV | 86.85 | 84.67 | 1.125 | 0/15 |

Note: Thresholds are reused from source calibration without adjustment to prototype ECG. Coverage includes all 15 configurations; MAE averages only nonempty configurations. Zero output is not assigned zero error. The two thresholds serve different calibration objectives.

# SUPPLEMENTARY FIGURES

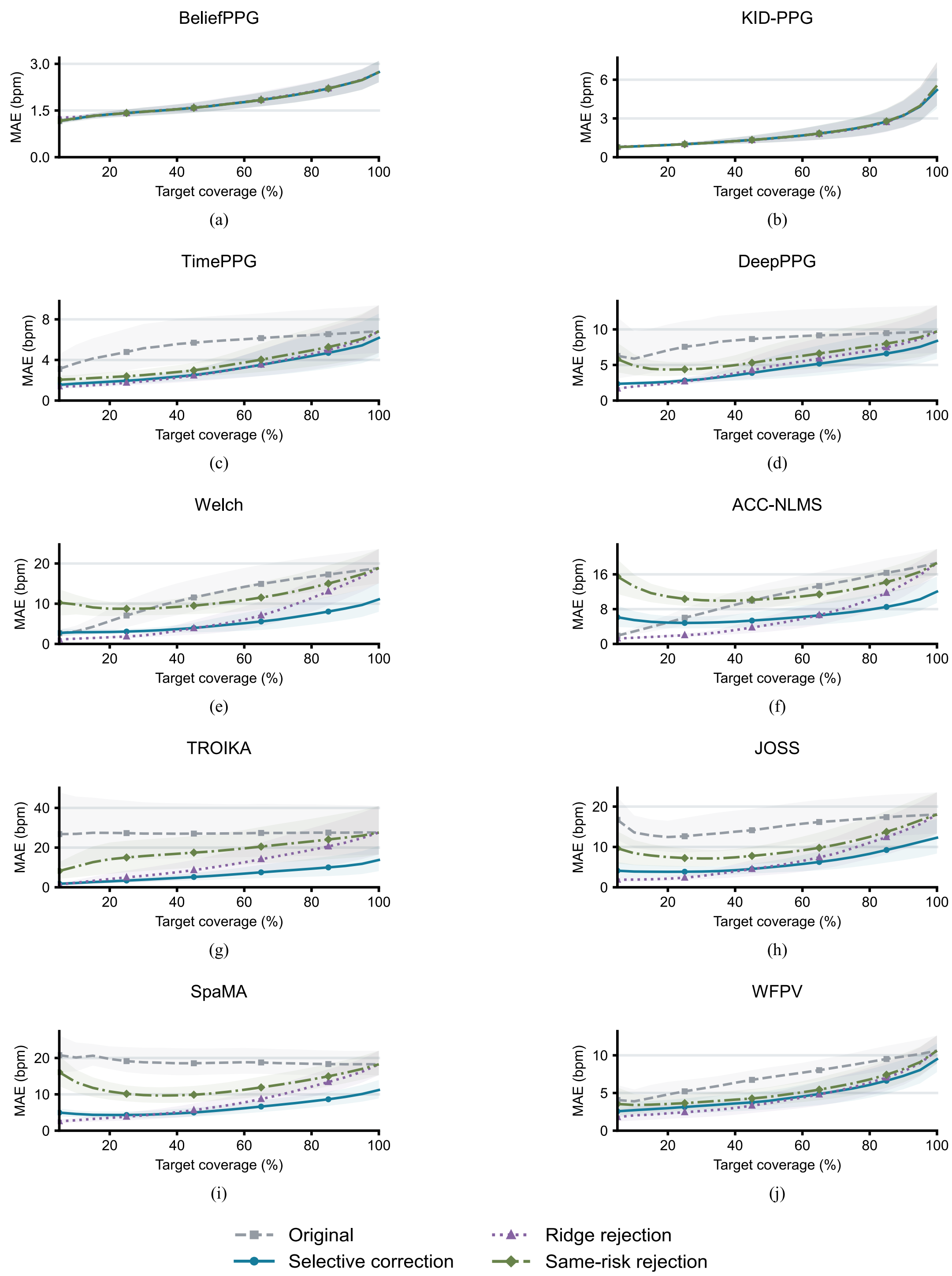

Fig. S1. Estimator-specific risk–coverage curves on DaLiA. (a)–(j) BeliefPPG, KID-PPG, TimePPG, Deep PPG, Welch, ACC-NLMS, TROIKA, JOSS, SpaMA, and WFPV. Curves compare original estimates, the framework, ridge rejection, and same-risk rejection, showing accepted-set MAE at different target coverage levels. Vertical scales differ by panel. Shading indicates participant-bootstrap 95% confidence intervals.

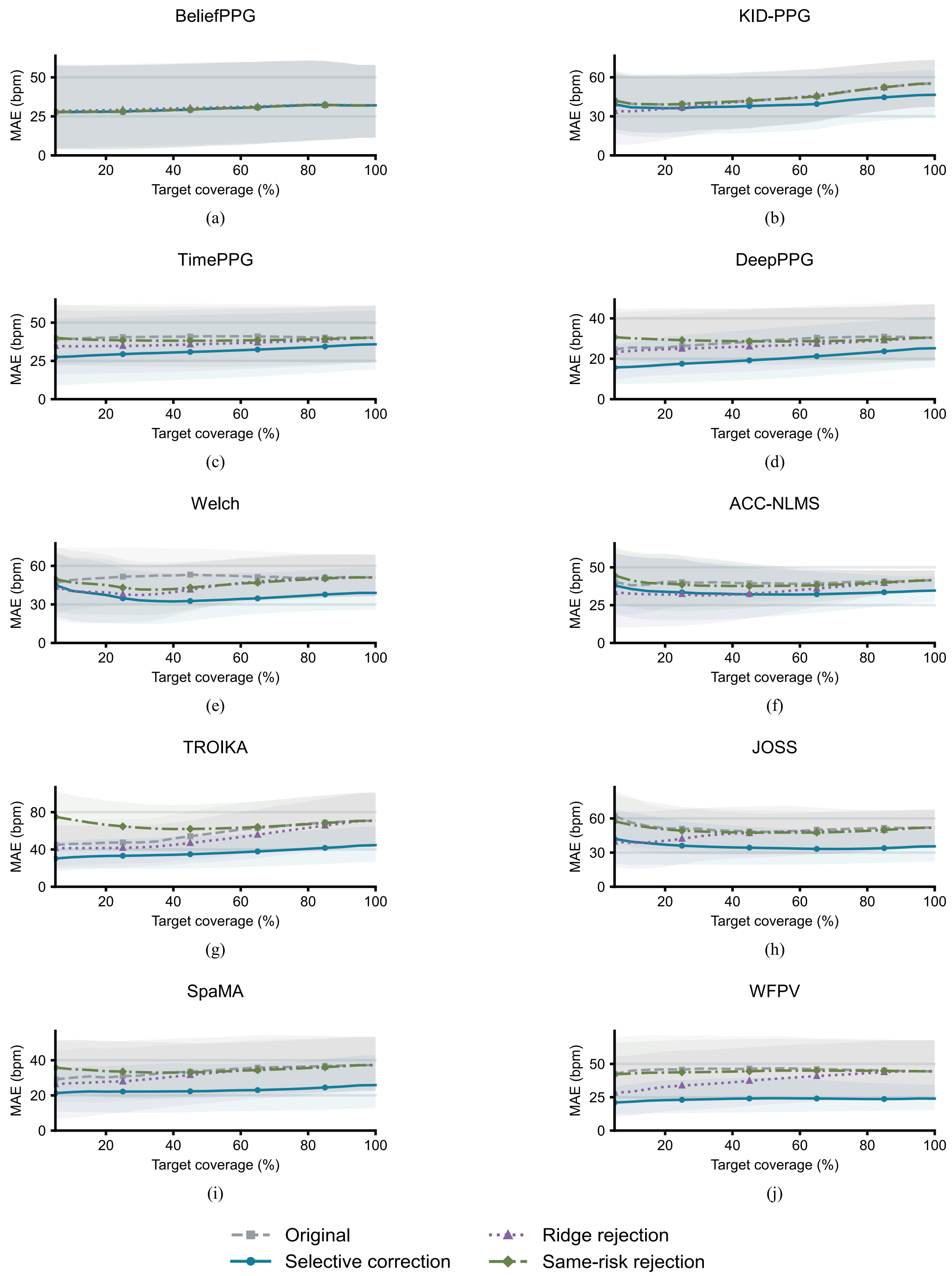

Fig. S2. Estimator-specific risk–coverage curves on Wrist. (a)–(j) BeliefPPG, KID-PPG, TimePPG, Deep PPG, Welch, ACC-NLMS, TROIKA, JOSS, SpaMA, and WFPV. Curves compare original estimates, the framework, ridge rejection, and same-risk rejection, showing accepted-set MAE at different target coverage levels. Vertical scales differ by panel. Shading indicates participant-bootstrap 95% confidence intervals.

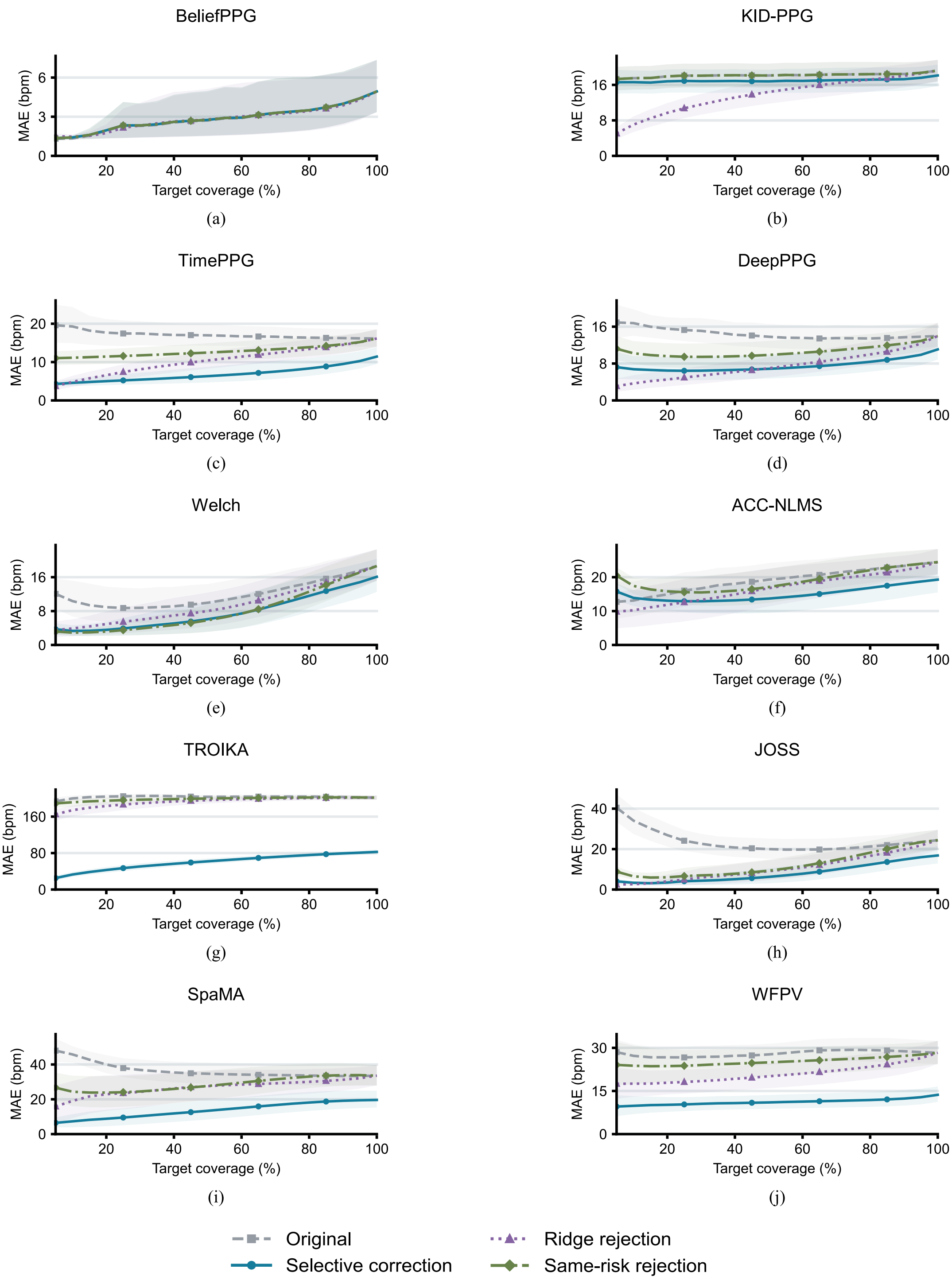

Fig. S3. Estimator-specific risk–coverage curves on Galaxy. (a)–(j) BeliefPPG, KID-PPG, TimePPG, Deep PPG, Welch, ACC-NLMS, TROIKA, JOSS, SpaMA, and WFPV. Curves compare original estimates, the framework, ridge rejection, and same-risk rejection, showing accepted-set MAE at different target coverage levels. Vertical scales differ by panel. Shading indicates participant-bootstrap 95% confidence intervals.

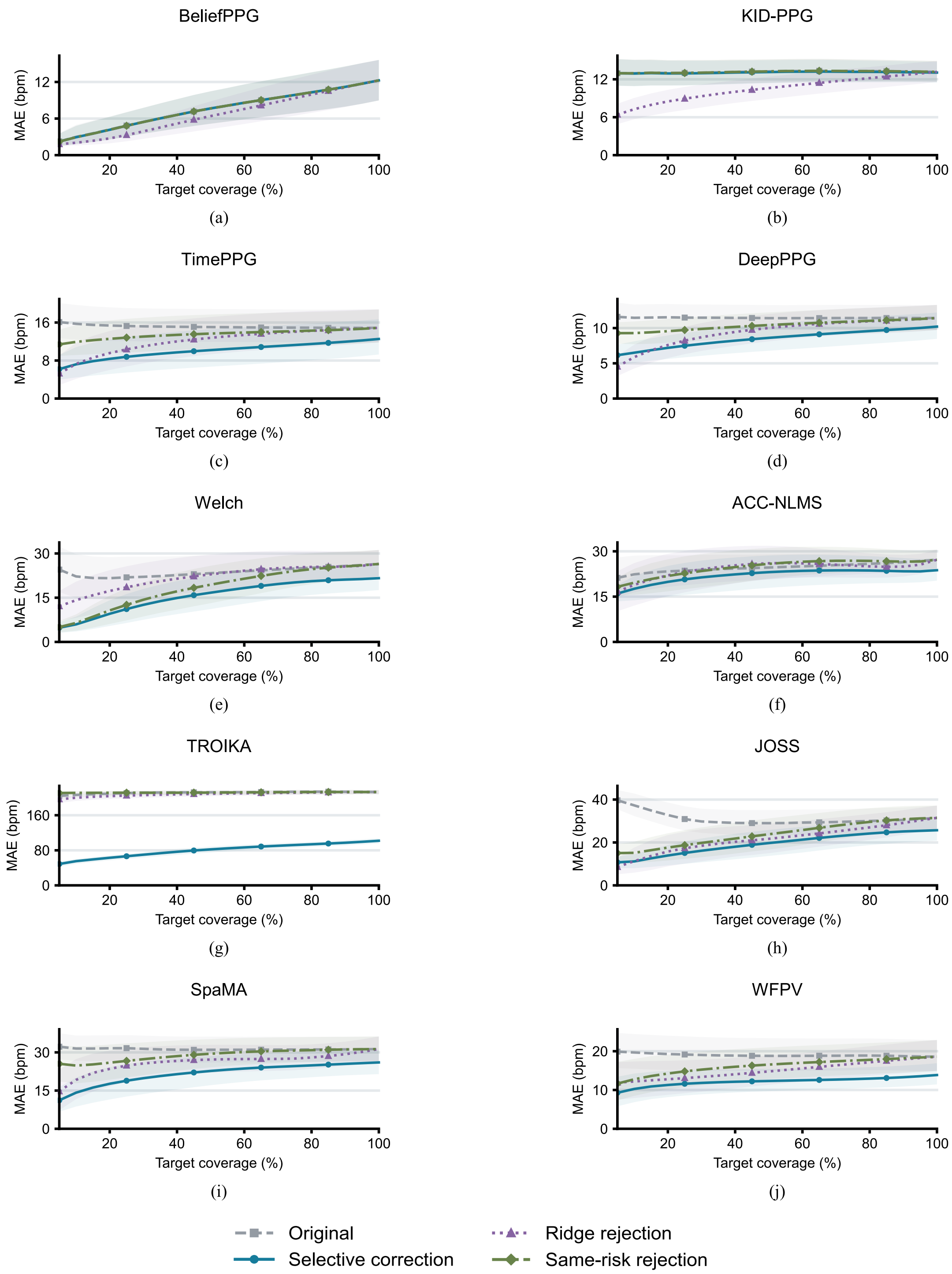

Fig. S4. Estimator-specific risk–coverage curves on Wild. (a)–(j) BeliefPPG, KID-PPG, TimePPG, Deep PPG, Welch, ACC-NLMS, TROIKA, JOSS, SpaMA, and WFPV. Curves compare original estimates, the framework, ridge rejection, and same-risk rejection, showing accepted-set MAE at different target coverage levels. Vertical scales differ by panel. Shading indicates participant-bootstrap 95% confidence intervals.

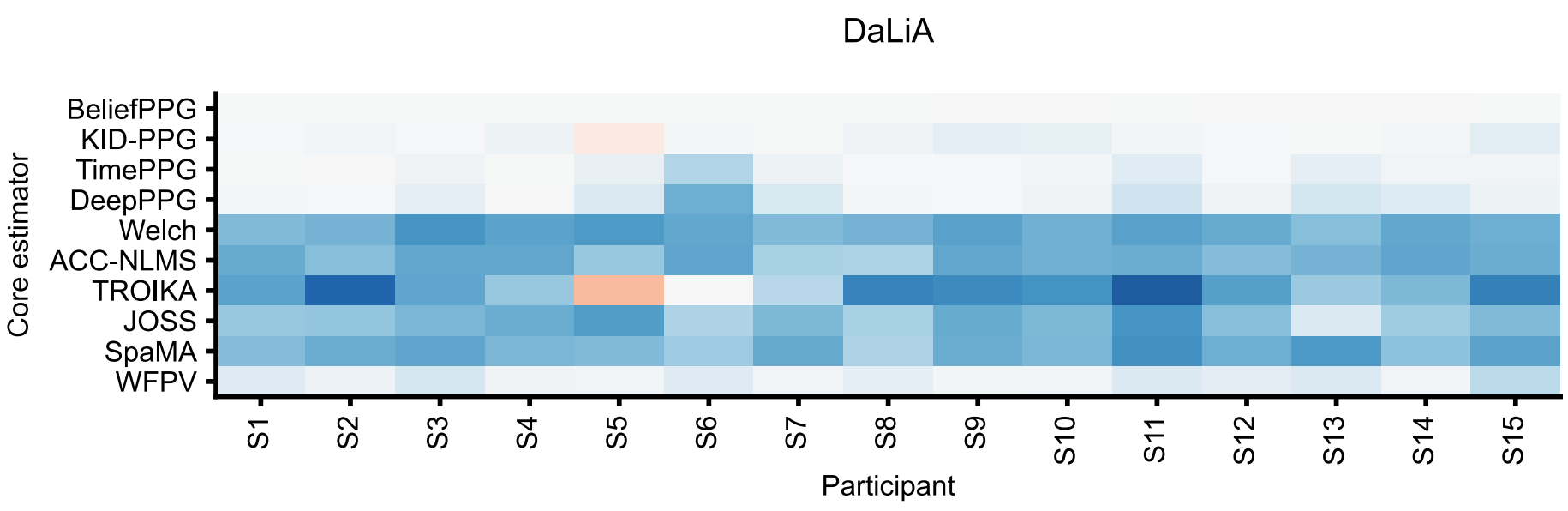

(a)

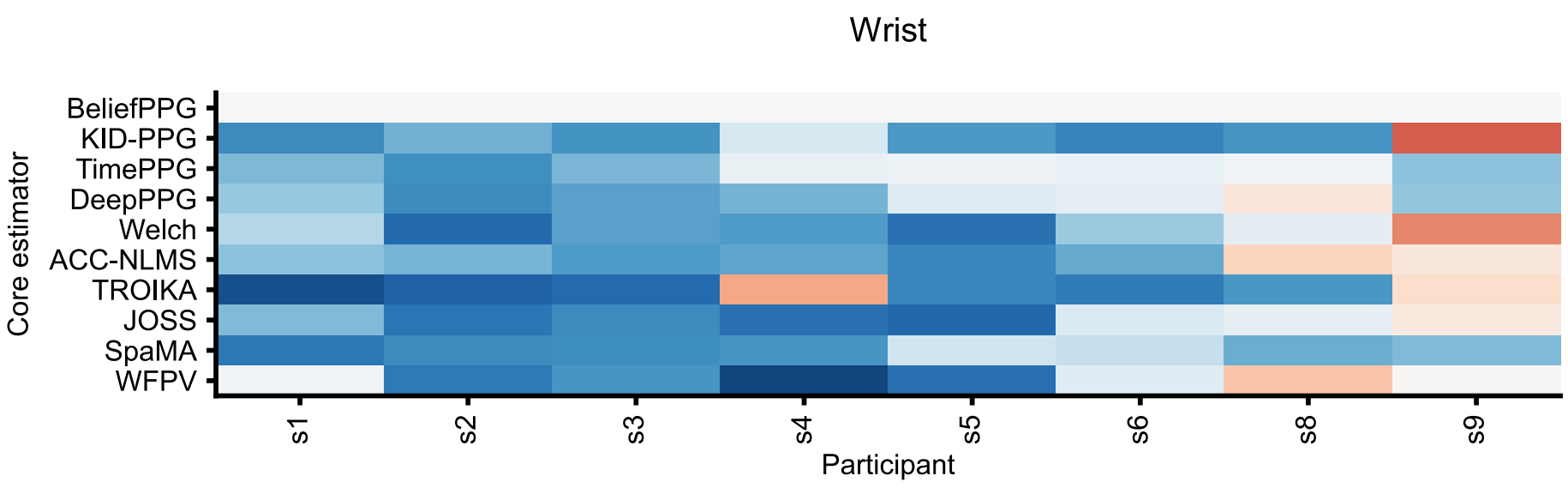

(b)

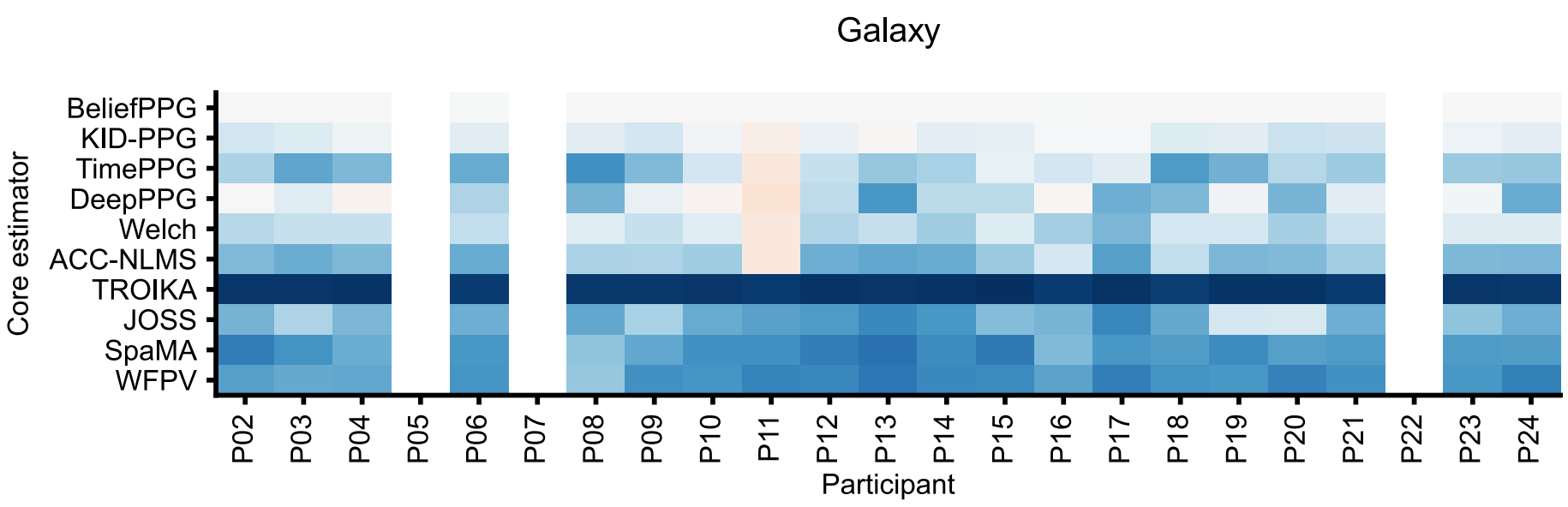

(c)

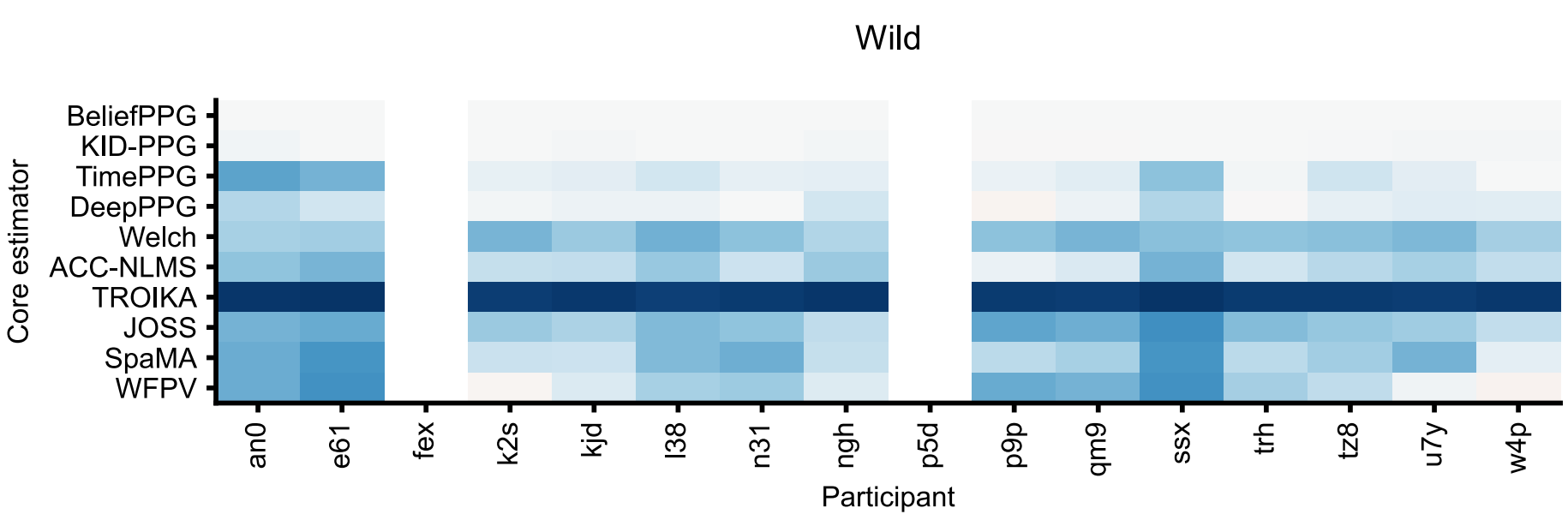

(d)

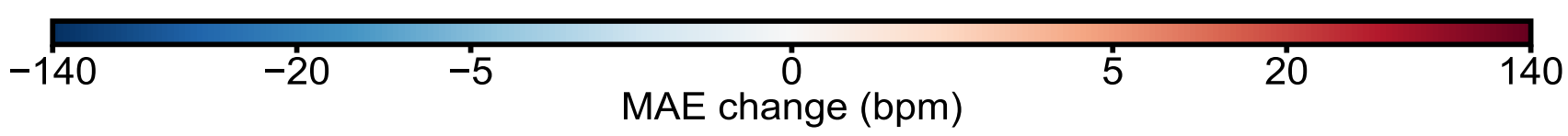

Fig. S5. MAE changes by participant and estimator. (a)–(d) DaLiA, Wrist, Galaxy, and Wild. Cells show framework minus original MAE; blue indicates decreases and red increases. The symmetric logarithmic scale is linear within ±5 bpm and spans ±140 bpm. Blank columns indicate participants without eligible references. Gains vary across participants and estimators.

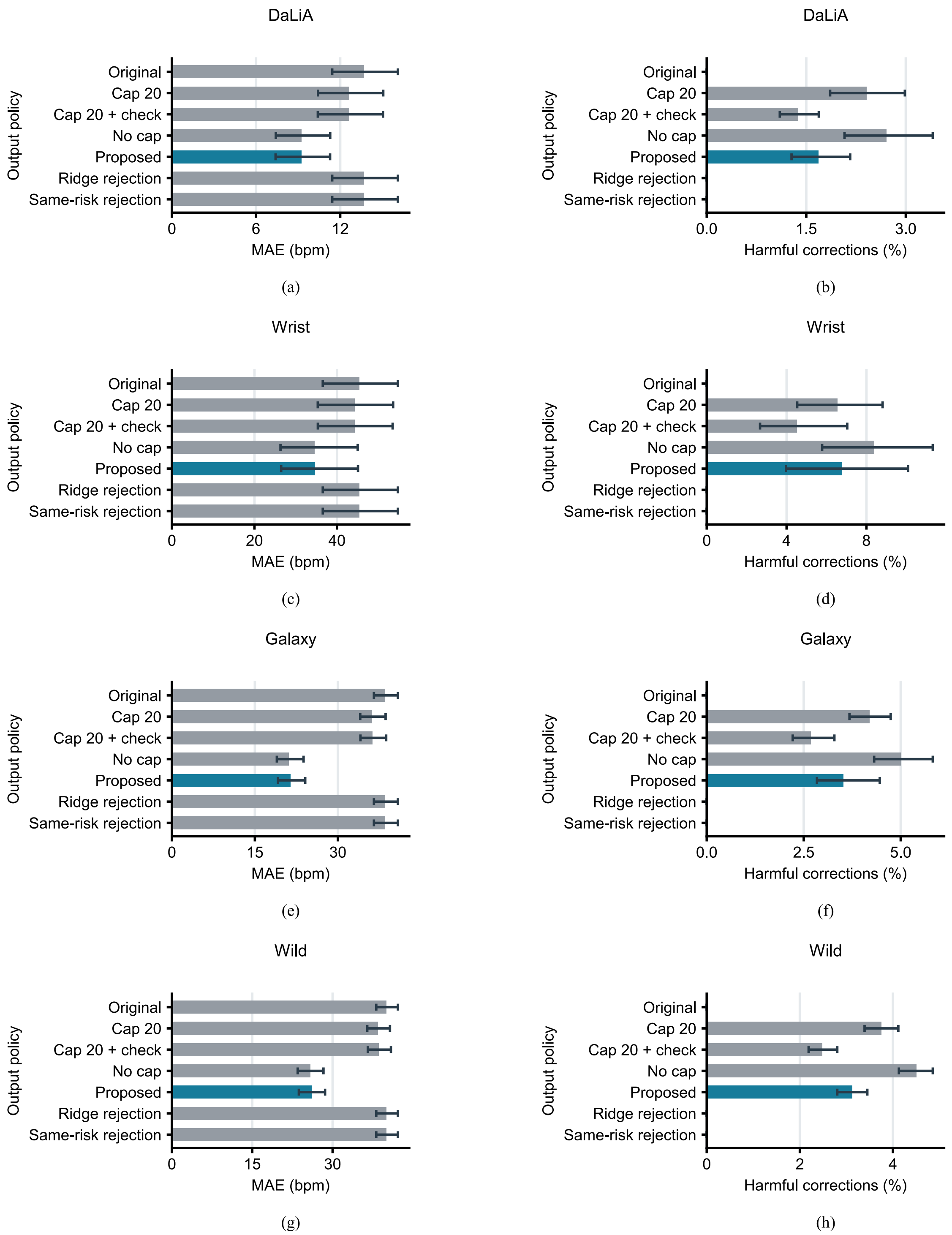

Fig. S6. Error and harmful correction rates for seven output strategies. Rows correspond to DaLiA, Wrist, Galaxy, and Wild. Left: MAE. Right: harmful correction rate. Cap 20 denotes a 20 bpm correction cap, check denotes posterior-gain consistency checking, and No cap denotes uncapped correction. Blue identifies the proposed method. Error bars indicate participant-bootstrap 95% confidence intervals.

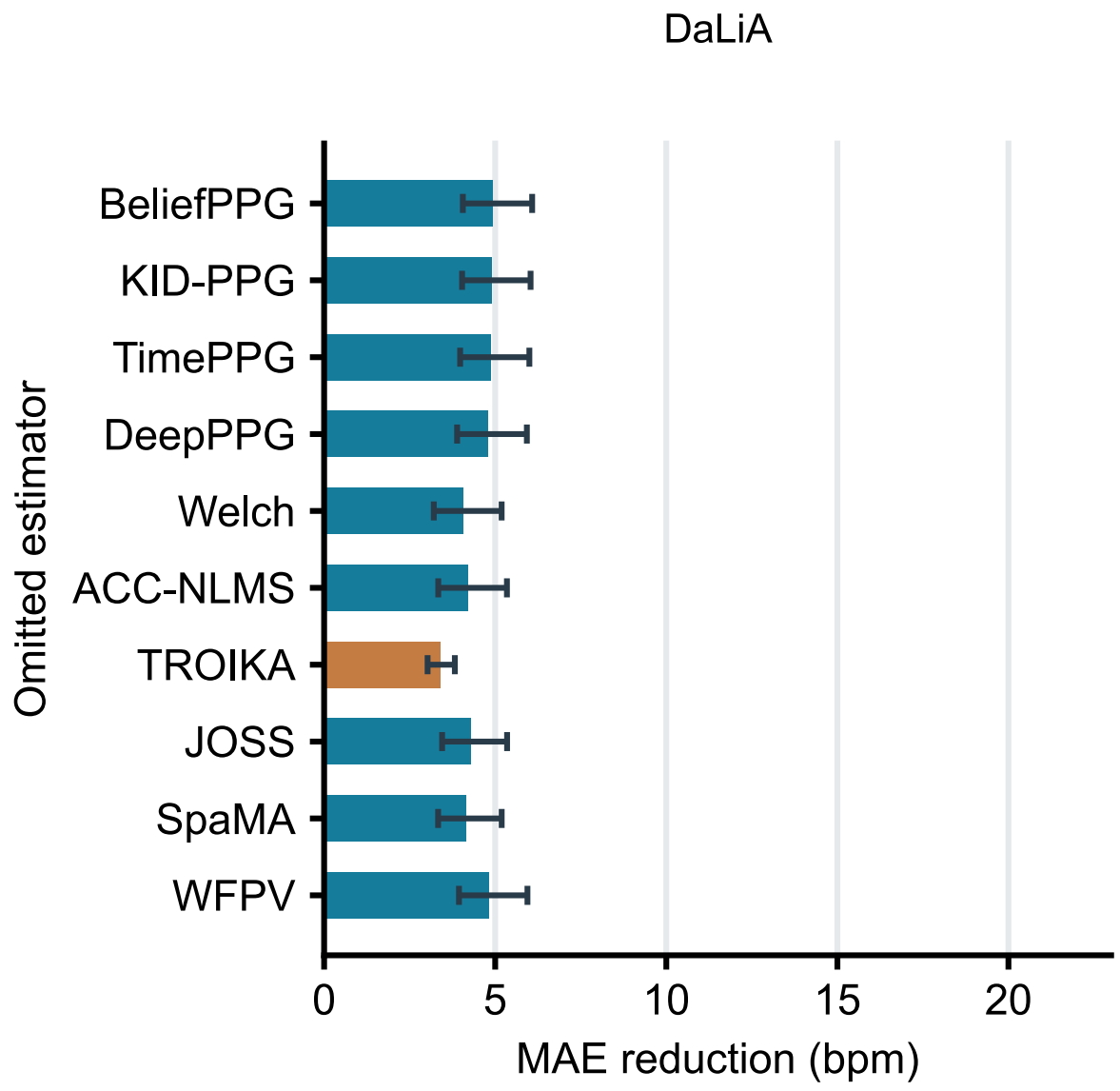

(a)

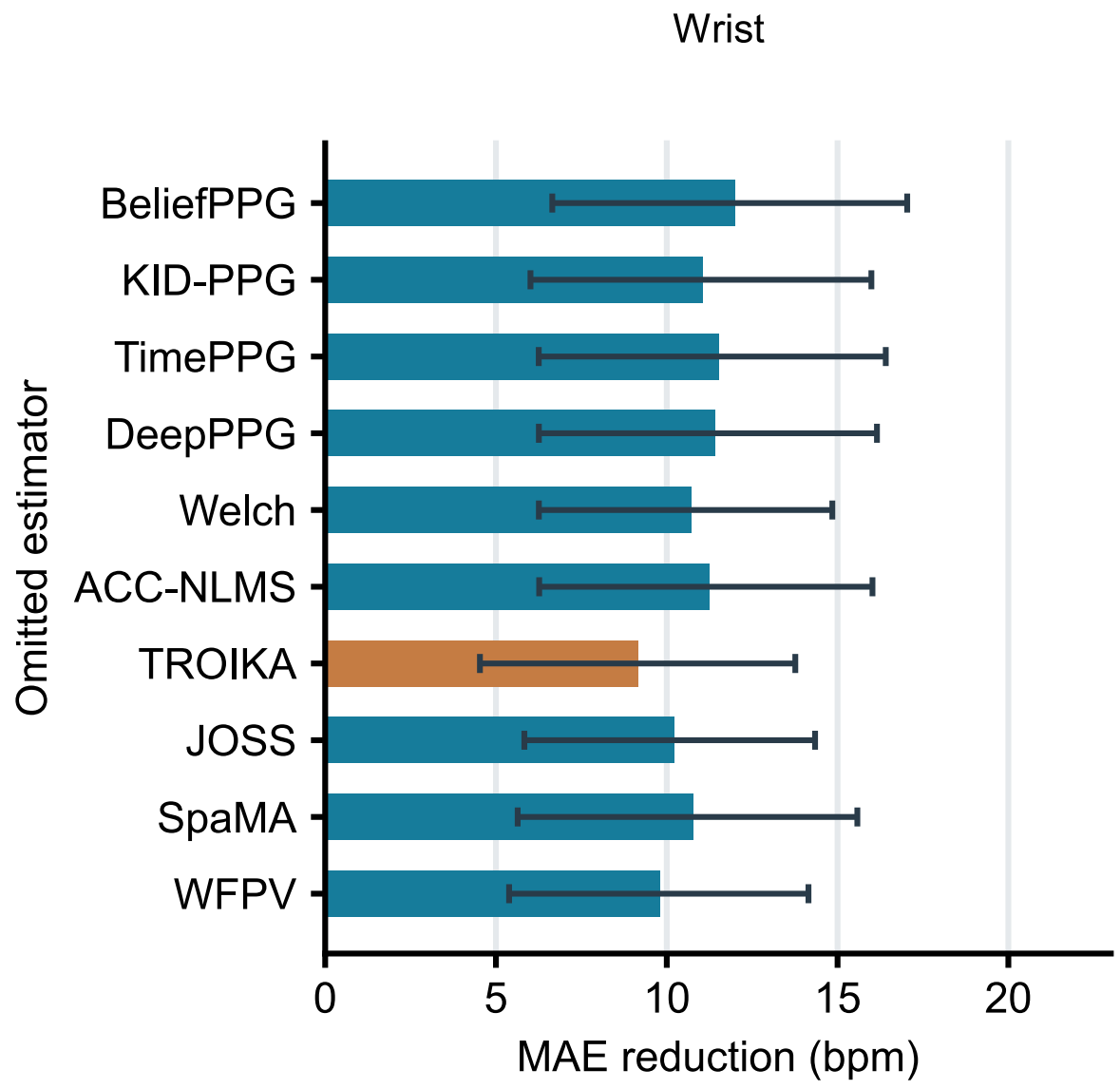

(b)

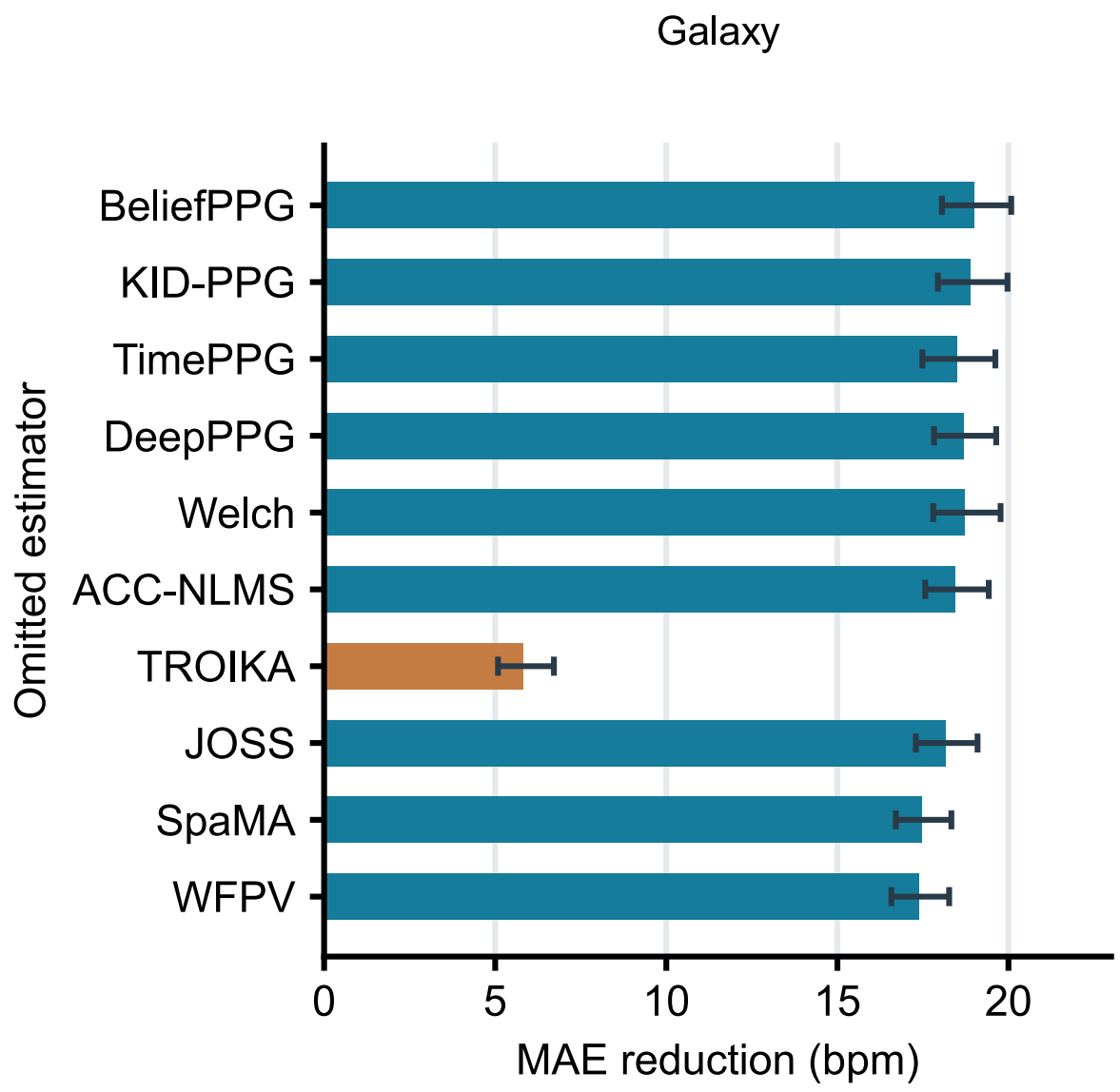

(c)

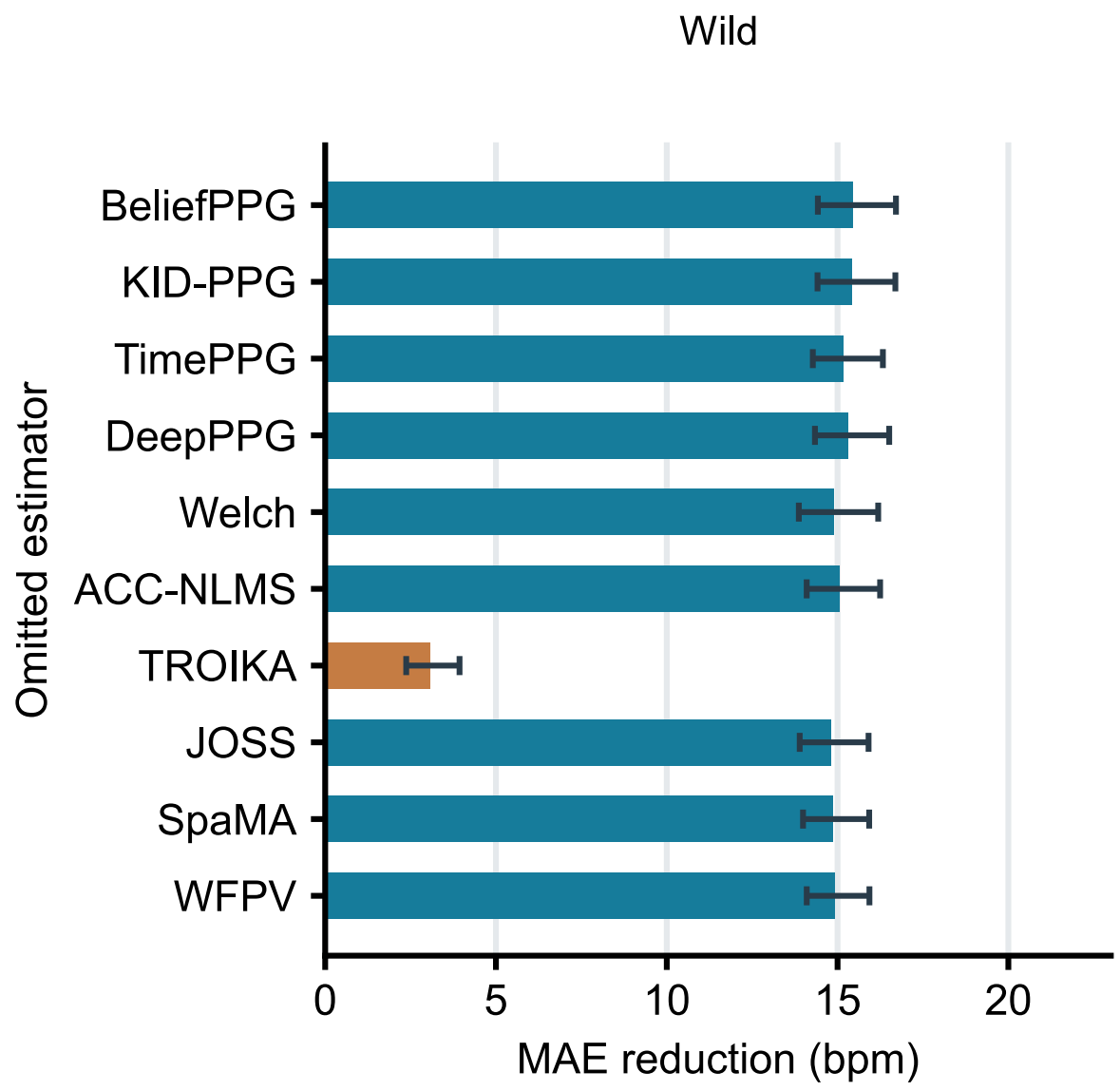

(d)

Fig. S7. Macro-averaged MAE reductions after excluding one estimator. (a)–(d) DaLiA, Wrist, Galaxy, and Wild. The vertical axis identifies the excluded estimator. Positive values favor the framework. Orange marks exclusion of local TROIKA; other exclusions are blue. Average improvements remain under every exclusion, although their magnitudes differ. Error bars indicate participant-bootstrap 95% confidence intervals.

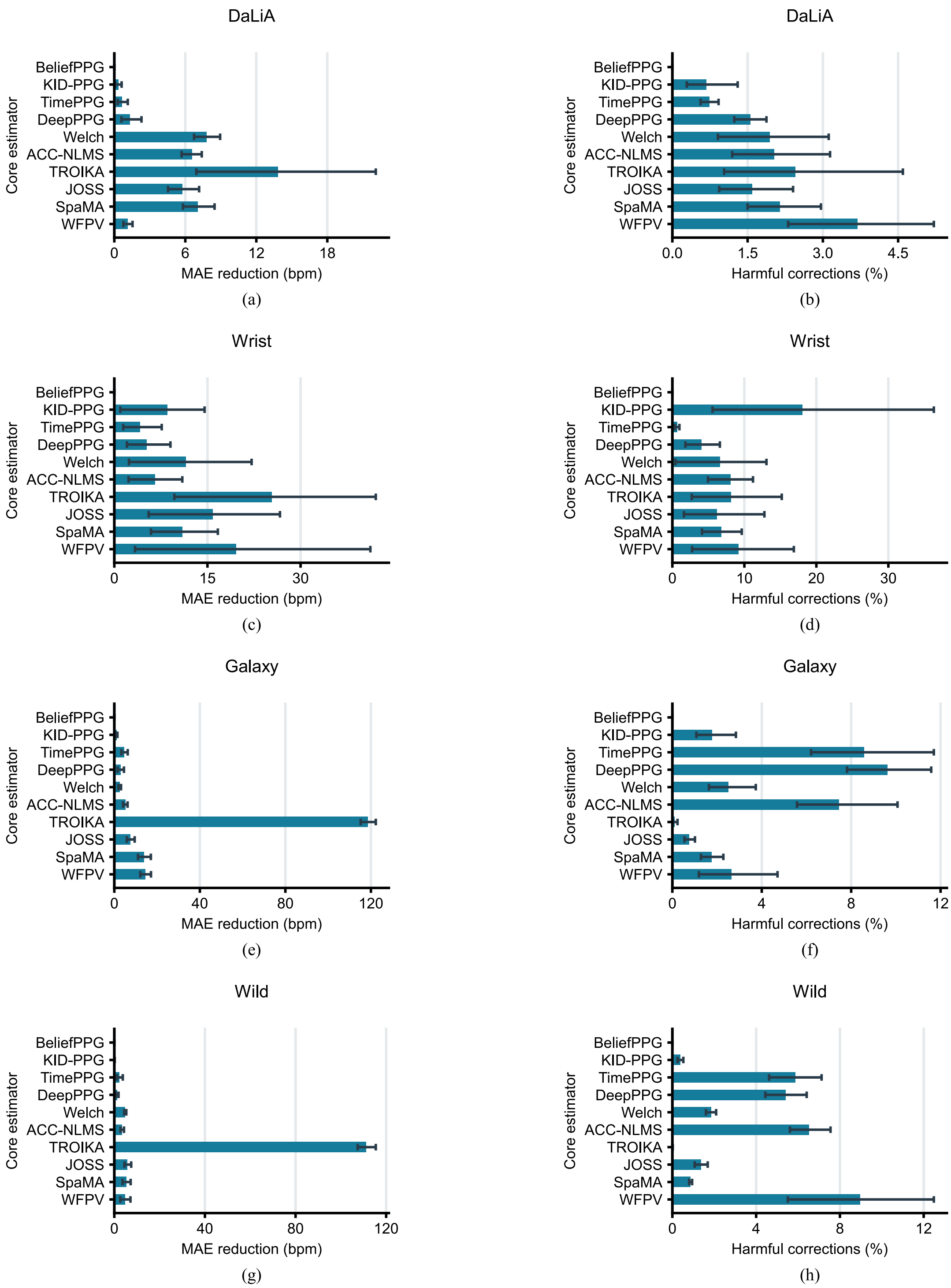

Fig. S8. Correction outcomes by estimator. Rows correspond to DaLiA, Wrist, Galaxy, and Wild. Left: MAE reduction. Right: harmful correction rate. Gains differ across estimators, with BeliefPPG nearly unchanged. Horizontal scales are selected for each dataset. Error bars indicate participant-bootstrap 95% confidence intervals.

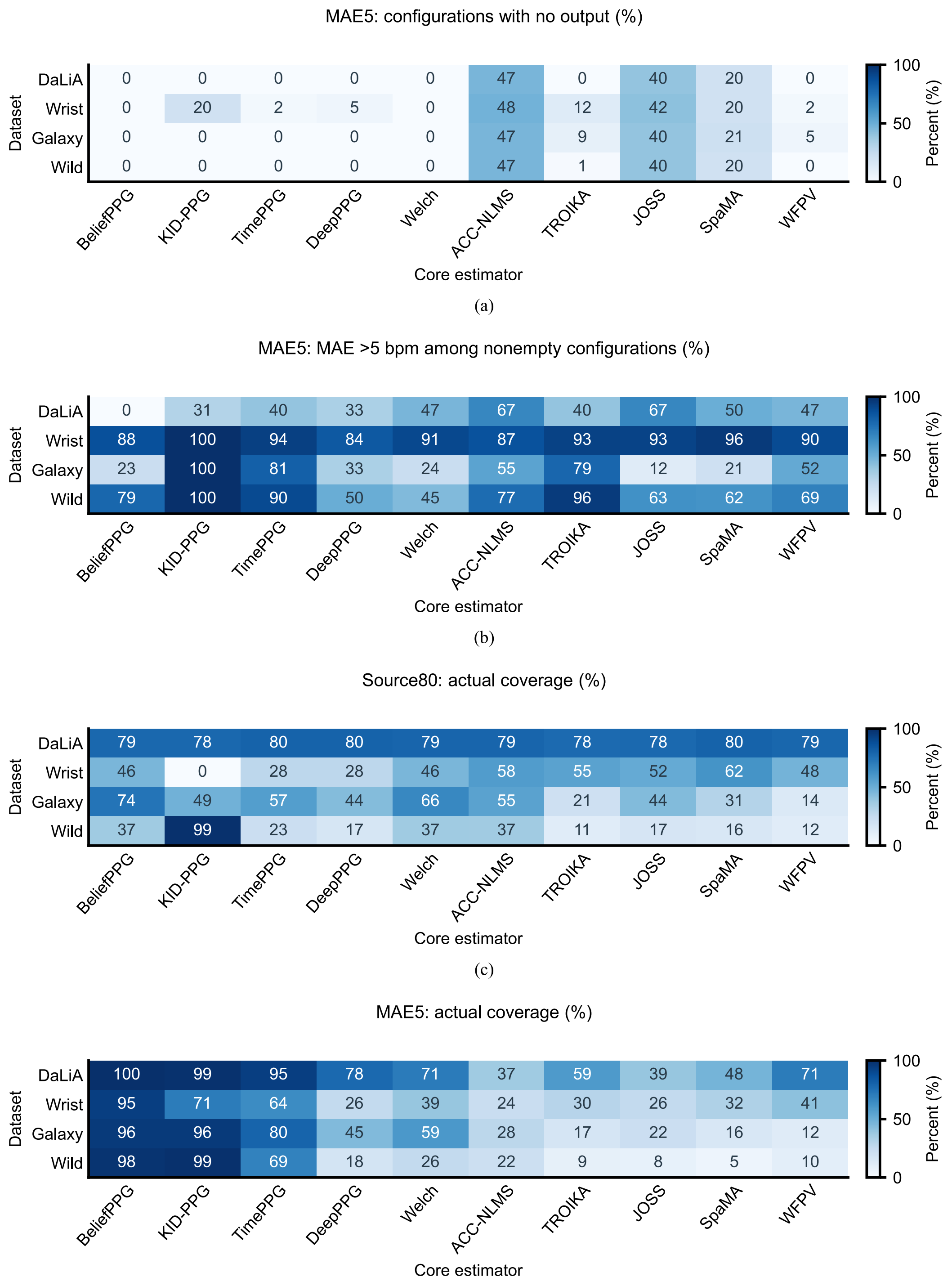

Fig. S9. Transfer of fixed source thresholds. (a) Zero-output configuration proportion under the source MAE 5 bpm threshold. (b) Proportion of nonempty configurations exceeding 5 bpm MAE. (c) Actual coverage under the source 80% threshold. (d) Actual coverage under the source MAE 5 bpm threshold. Cells are percentages on a common 0%–100% color scale. External datasets do not retain the source coverage or error targets.

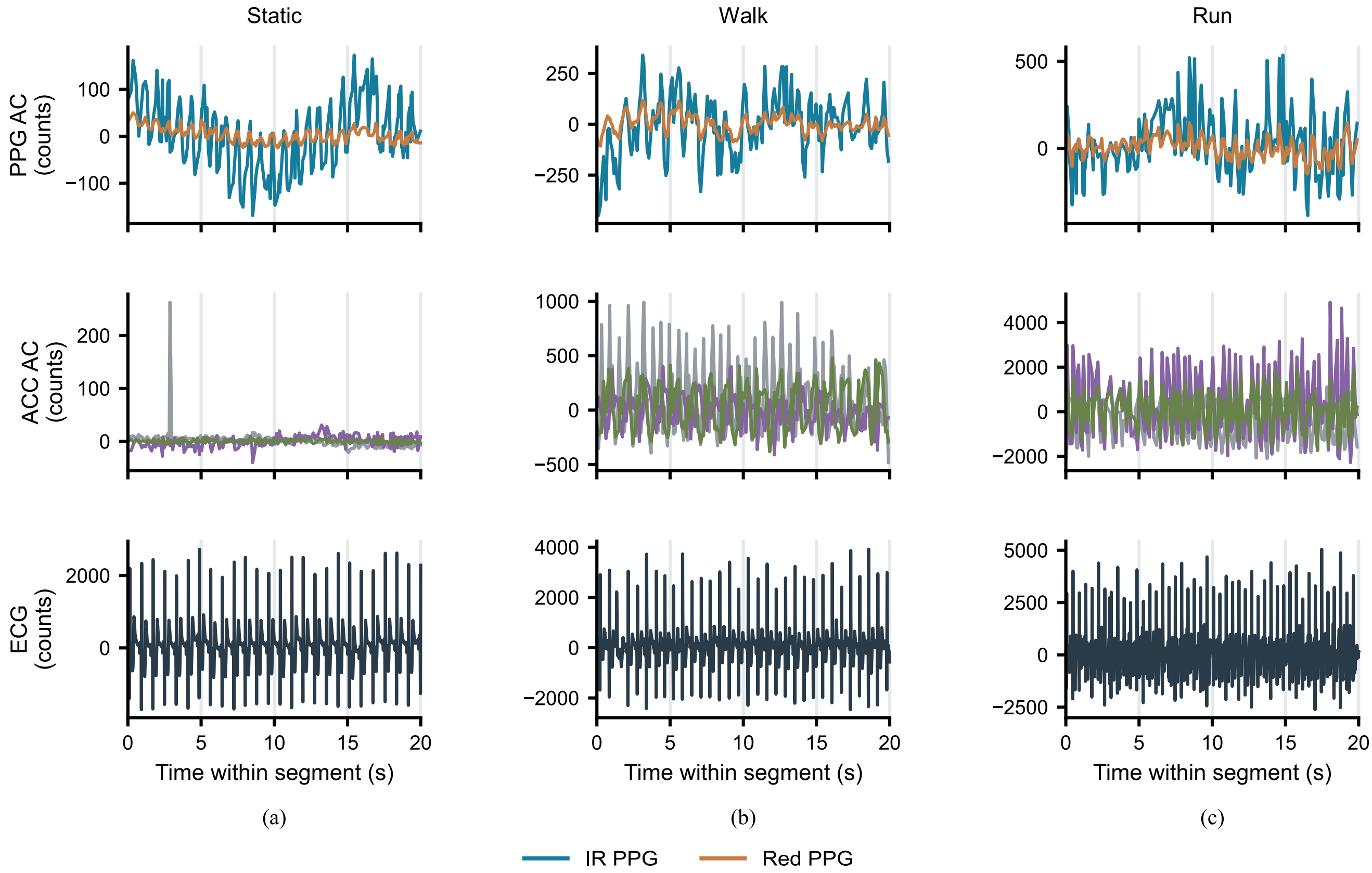

Fig. S10. Fixed 20 s examples of synchronized raw prototype signals. Dual-wavelength PPG, acceleration, and independent ECG are shown for static, walking, and running recordings. Units follow panel labels; offsets are removed only for display. Separate recordings are not concatenated into one continuous acquisition. These are acquired signals; model inputs after preprocessing are shown in Fig. S18. (a)–(c) Static, walking, and running. Gray, purple, and green acceleration curves denote x, y, and z axes.

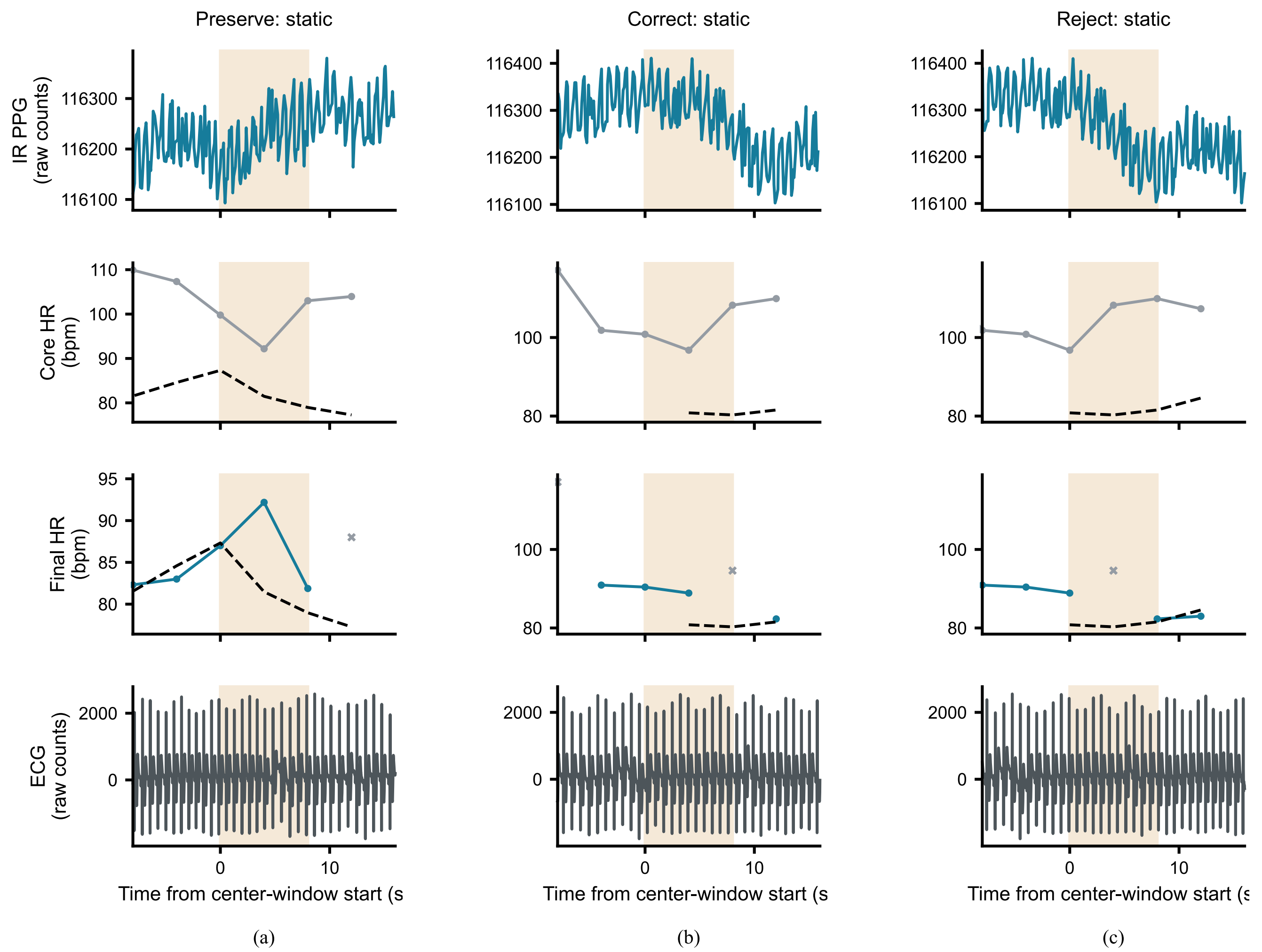

Fig. S11. Retention, correction, and rejection examples in prototype recordings using TimePPG. Each column aligns raw signals, heart rates, and references. The original estimator and framework receive identical processed inputs. Examples use source fold 0 and seed 219 and the first-eligible-window rule in S20.3, without selection for error reduction. Rejected windows have no final heart rate. Examples do not estimate overall action frequencies.

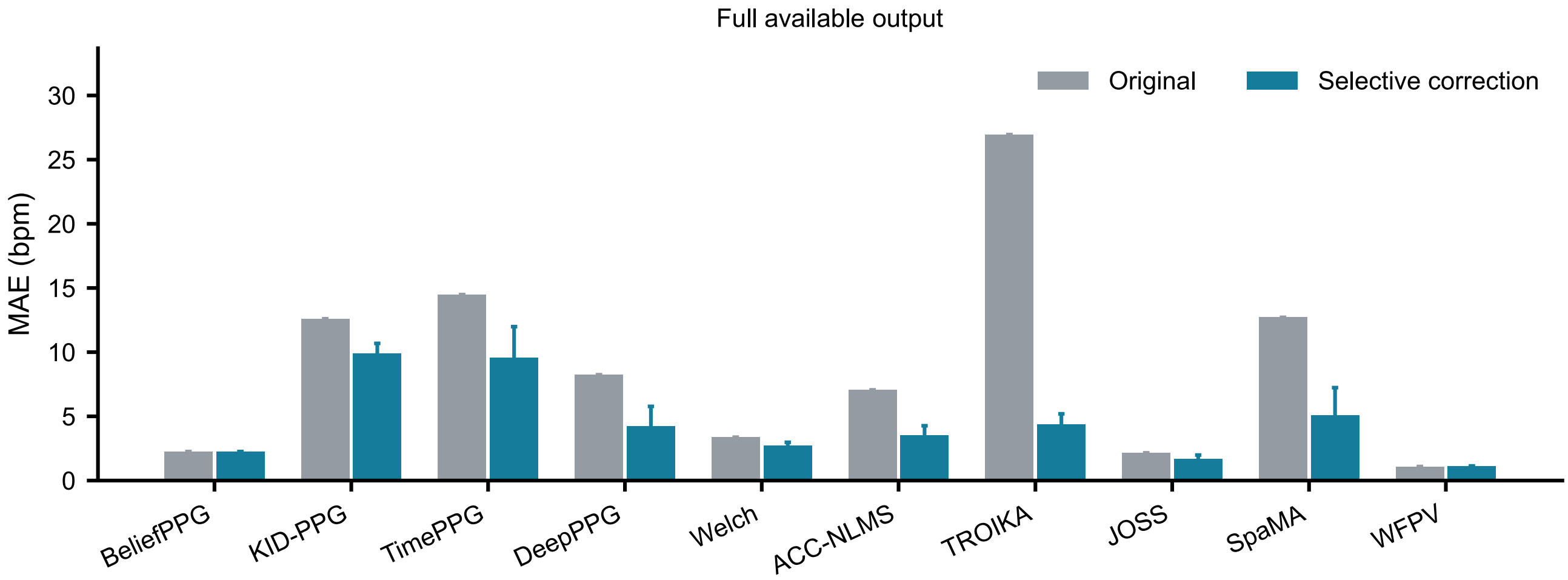

(a)

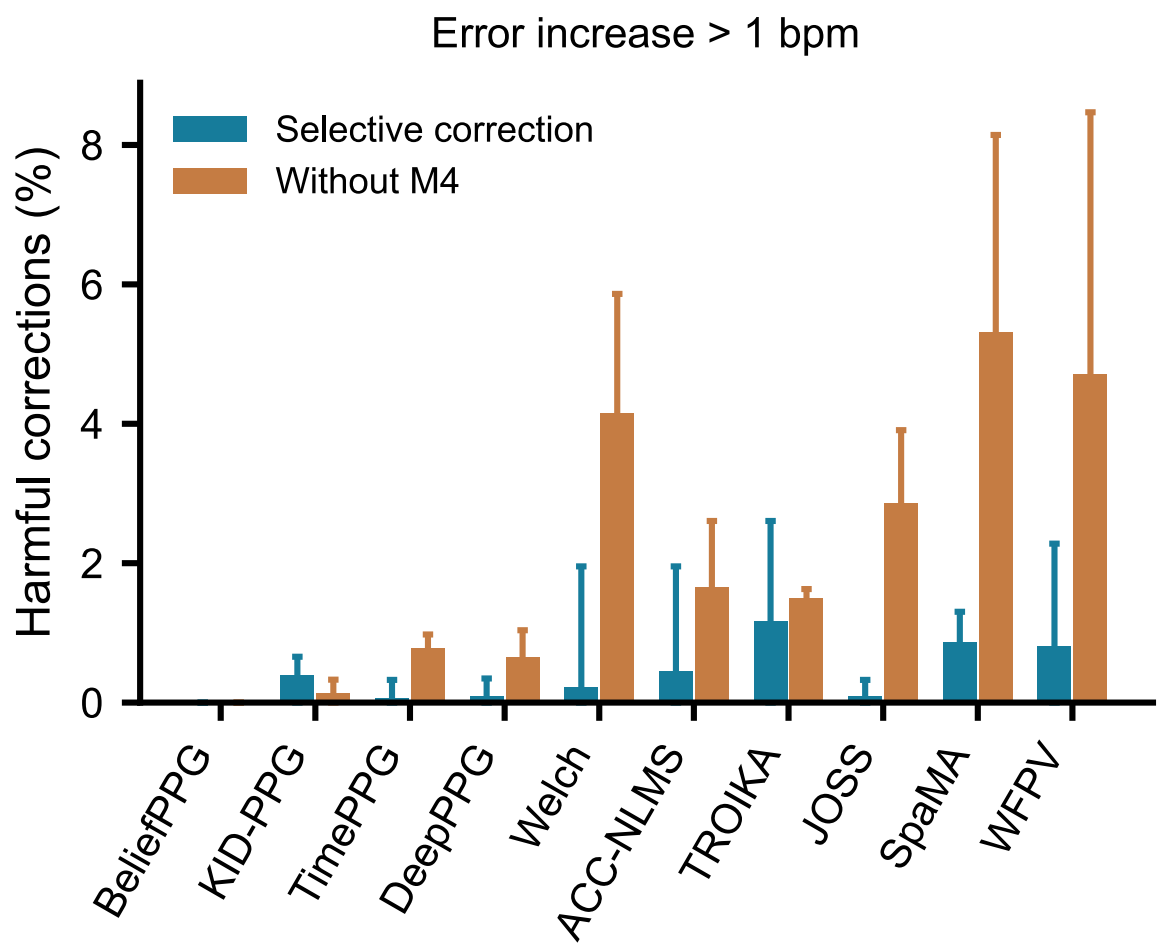

(b)

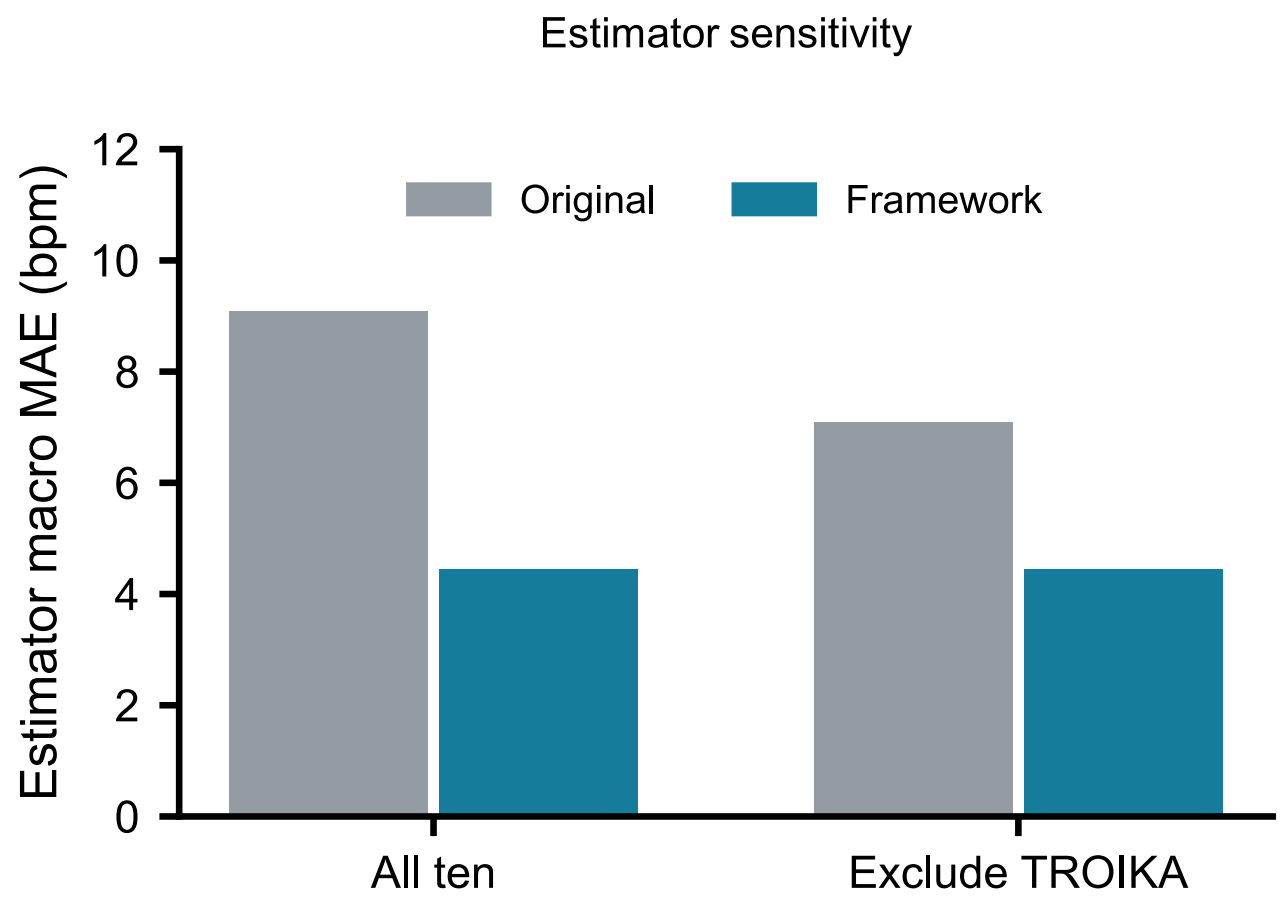

(c)

Fig. S12. All-available-output accuracy and harmful correction in prototype infrared recordings. (a) Paired MAE for ten estimators and their frameworks. (b) Proportions with absolute-error increases above 1 bpm, with and without M4. (c) Macro-averages for all ten estimators and after excluding TROIKA. Four states are pooled, with estimator-specific paired available windows and no additional rejection. Bars are means over 15 frozen source configurations; error bars show configuration ranges. n = 1.

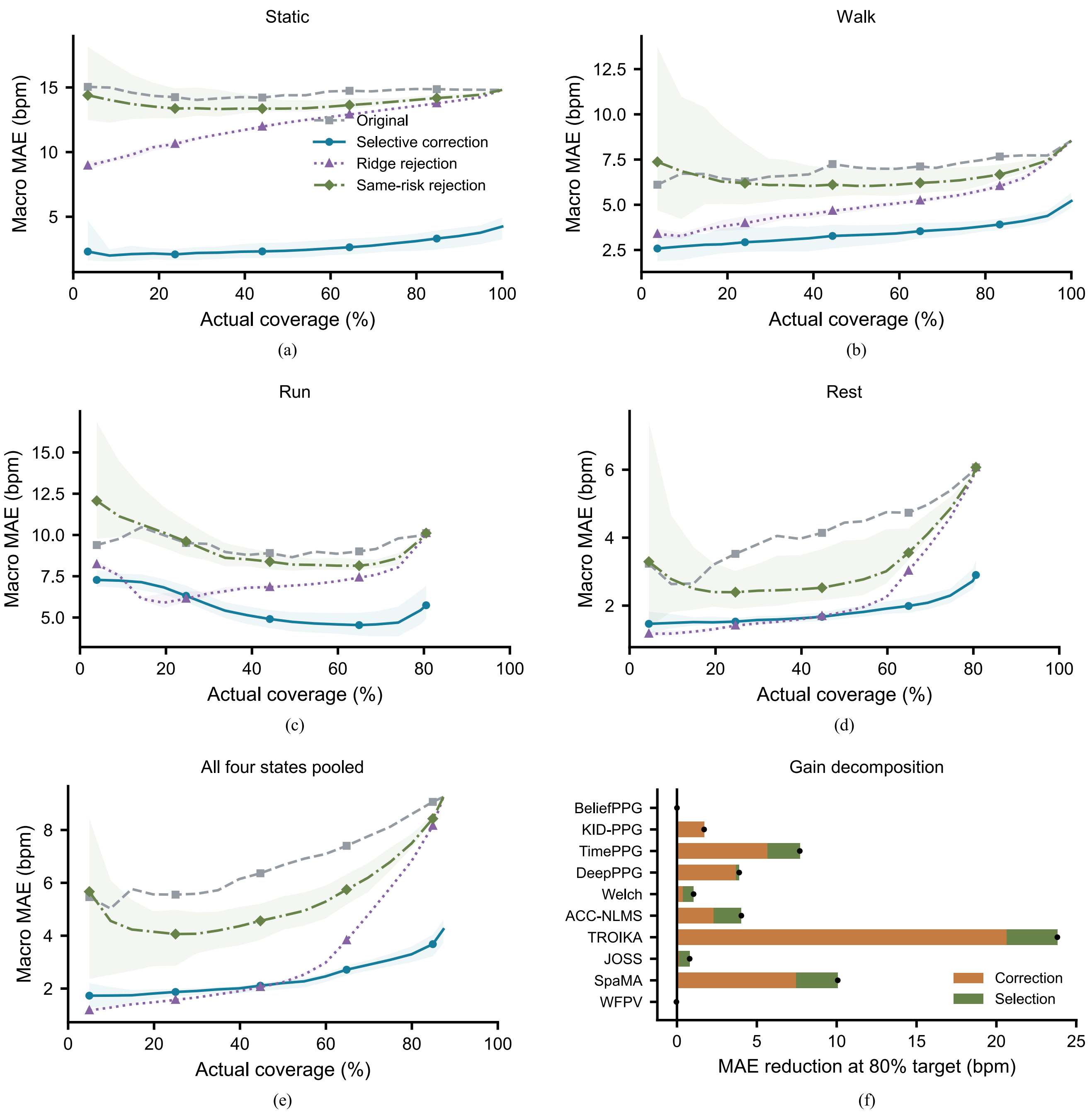

Fig. S13. Risk–coverage curves and gain decomposition in prototype infrared recordings. (a)–(d) Static, walking, running, and recovery rest. (e) Four states pooled. (f) Estimator-specific correction and selection gains at 80% target coverage. Curves show original-risk selection, the framework, ridge rejection, and same-risk rejection. The framework and same-risk rejection share accepted windows. Shading shows the range of 15 source configurations, not population confidence intervals.

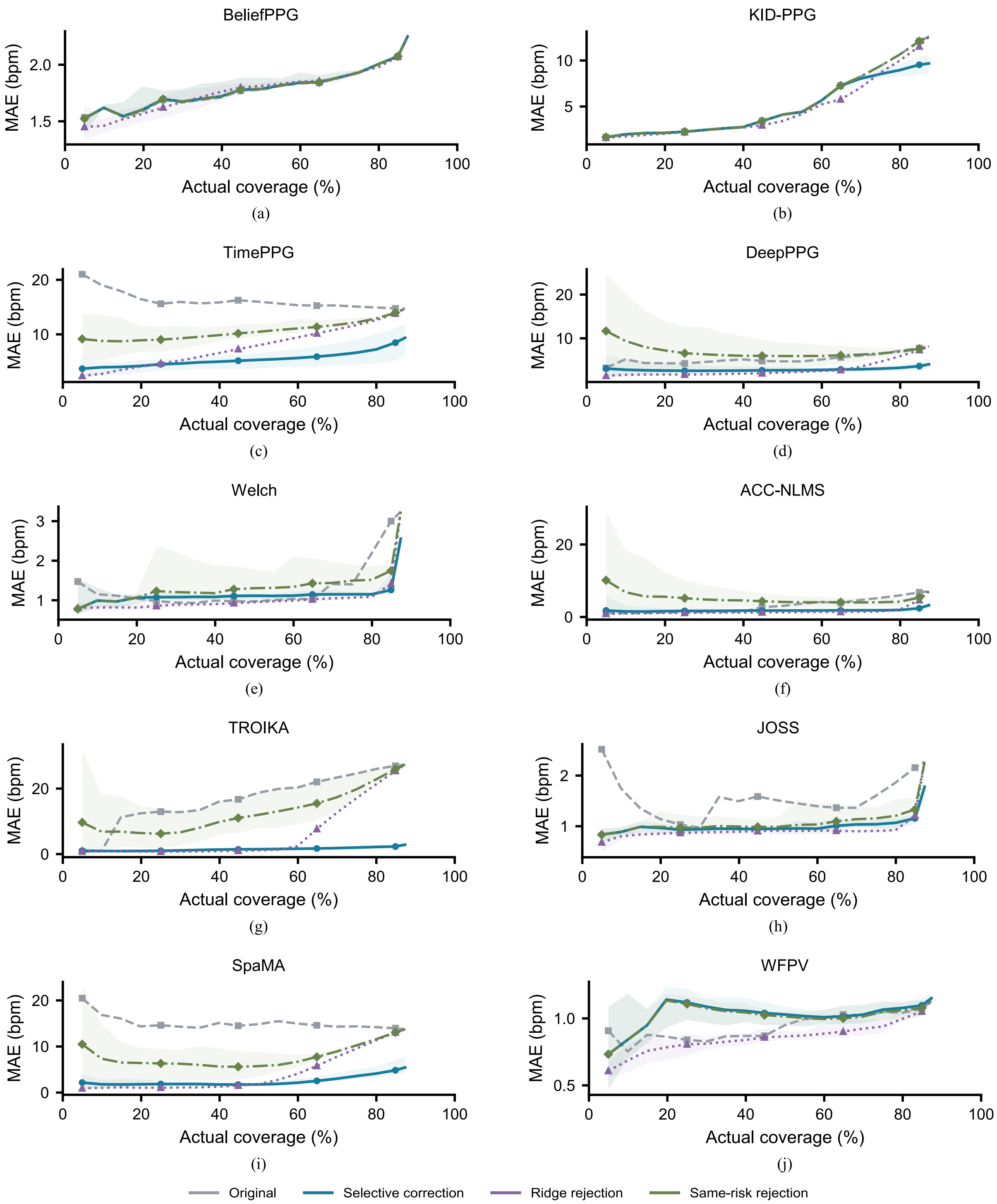

Fig. S14. Risk–coverage curves for ten estimators in prototype infrared recordings, pooled across four states. Panel-specific vertical scales show original-risk selection, the framework, ridge rejection, and same-risk rejection. The horizontal axis is actual coverage. Curves rank risk within target recordings and differ from fixed-source-threshold transfer. Shading indicates frozen-source-configuration ranges. n = 1.

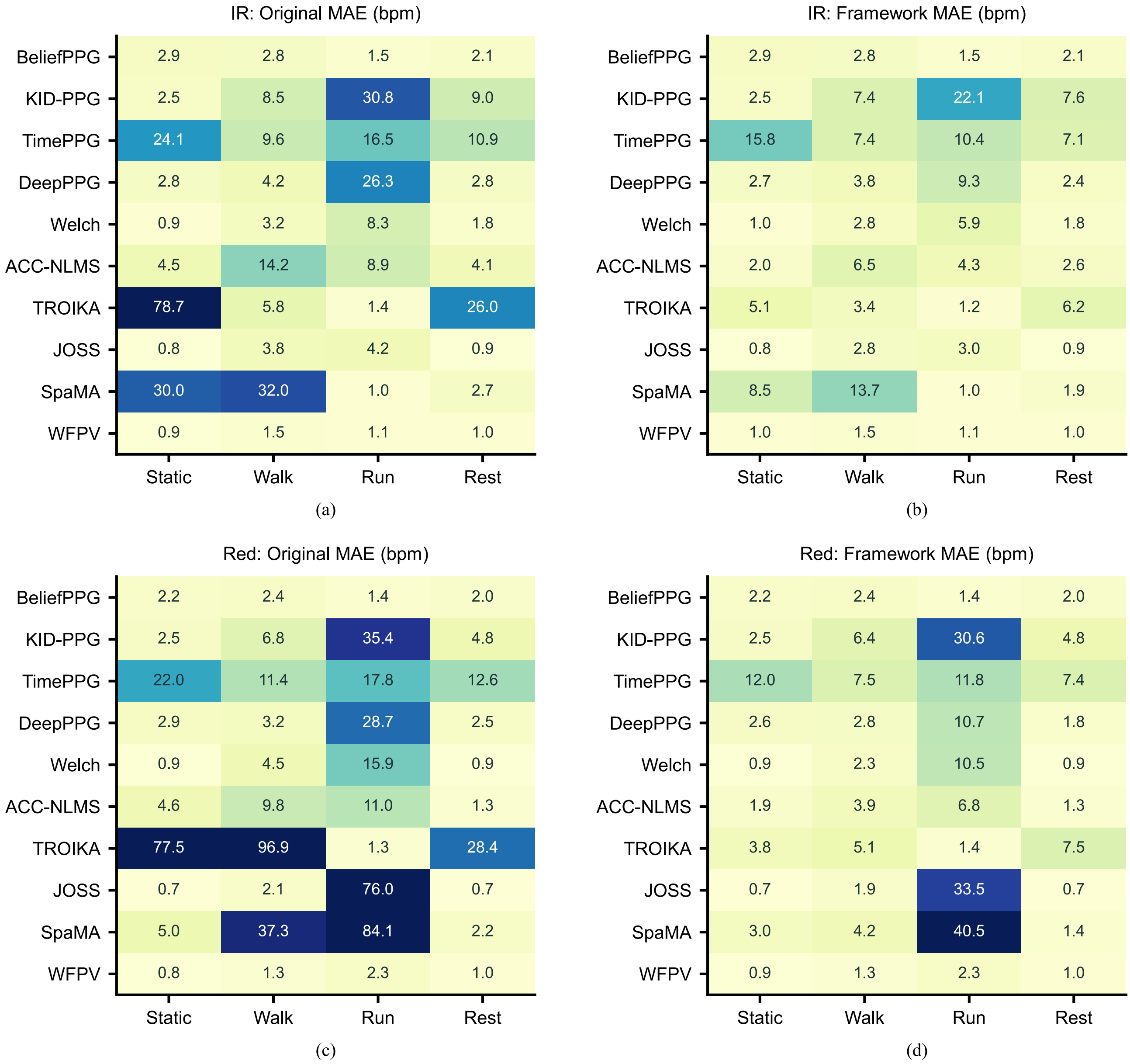

Fig. S15. Estimator-specific MAE across two channels and four prototype states. (a), (b) Original and framework infrared results. (c), (d) Original and framework red-channel results. Each method uses its own paired available windows. Values average 15 frozen source configurations. Color saturates at 40 bpm; labels retain actual values. Infrared is primary, and red is a separate sensitivity analysis. Channels are not separate participants.

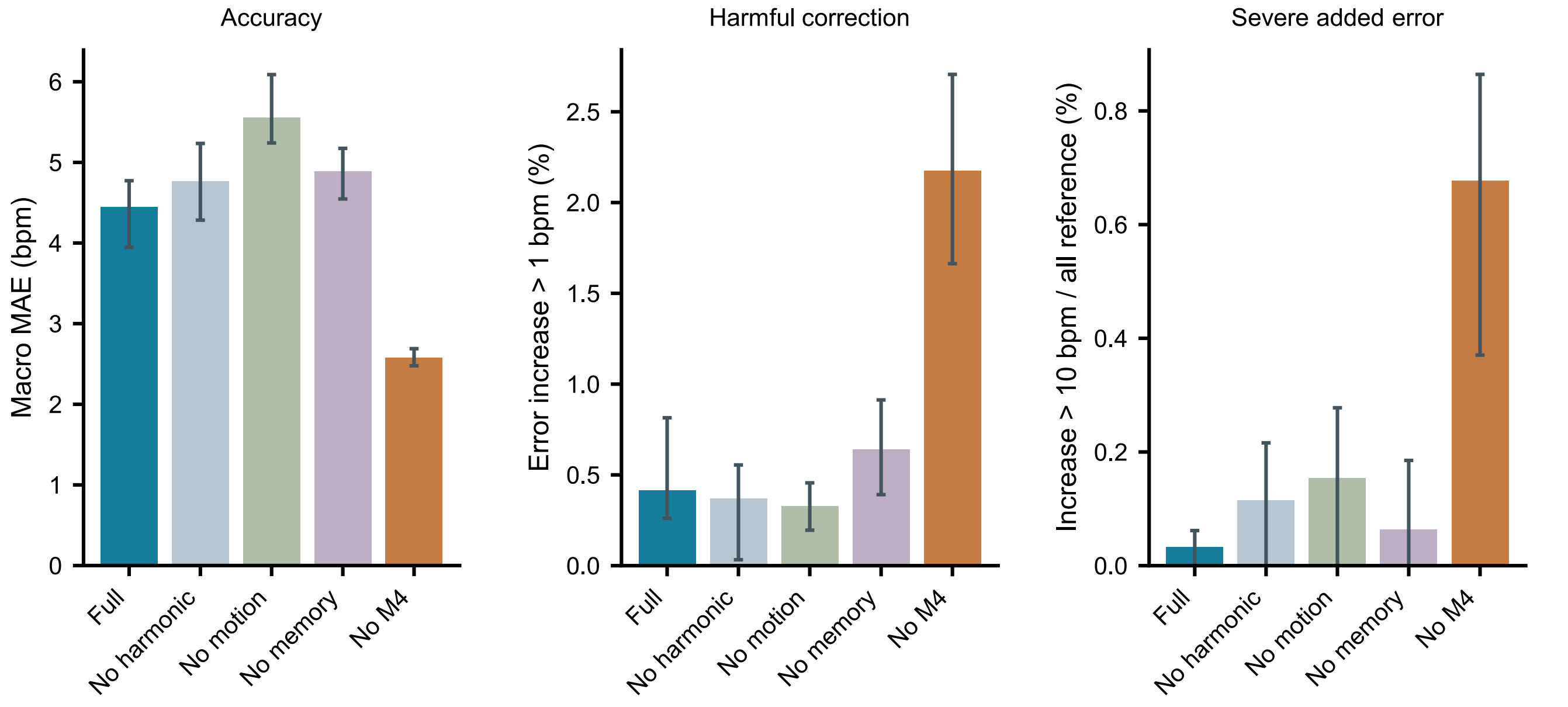

(a) (b) (c)

Fig. S16. Module-ablation trade-offs in prototype infrared recordings. (a) Macro-averaged MAE. (b) Harmful correction rate. (c) Severe added-error proportion relative to all reference-eligible windows. Full, No harmonic, No motion, and No memory reuse their corresponding source-trained weights. No M4 removes candidate gain–harm selection. All comparisons pool four states without additional rejection. Error bars show ranges across 15 frozen source configurations.

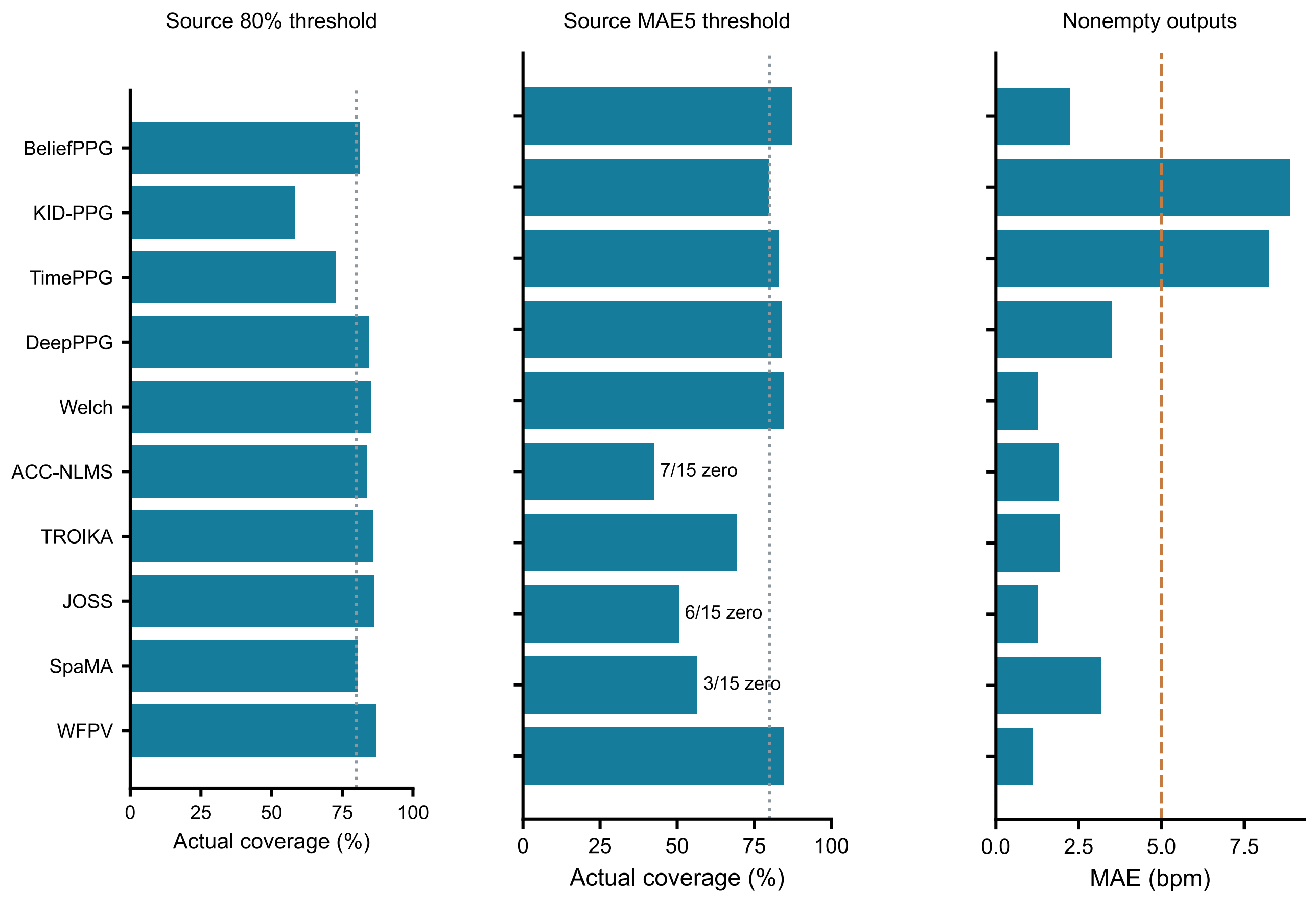

(a) (b) (c)

Fig. S17. Fixed-source-threshold transfer to prototype infrared recordings. (a) Actual coverage under source80. (b) Actual coverage under source MAE 5 bpm thresholds. (c) MAE for nonempty configurations under the latter rule. Thresholds are not adjusted to prototype ECG. Coverage includes all 15 configurations; MAE averages only nonempty configurations, with zero-output counts labeled separately. Missing output is not zero error.

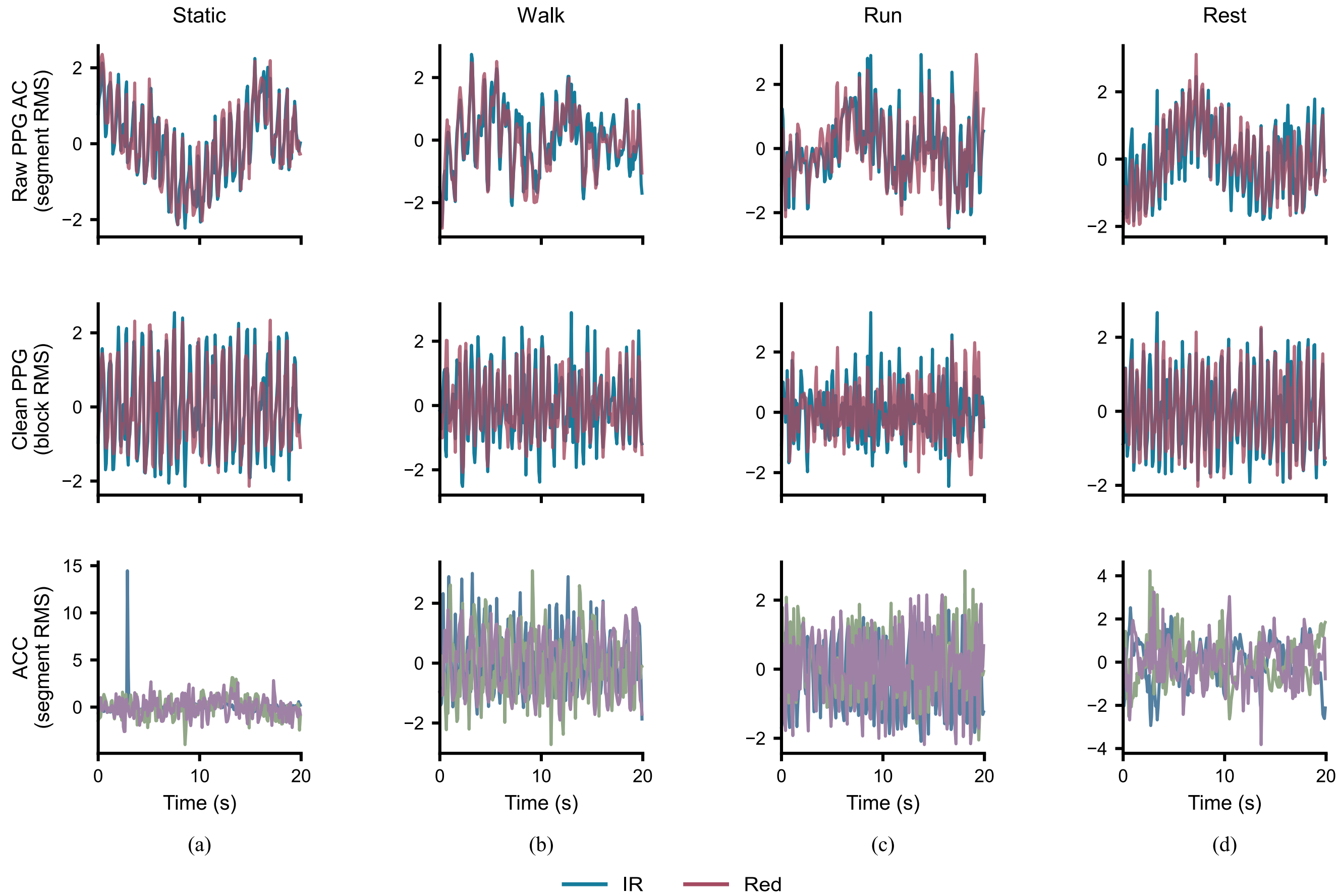

Fig. S18. Fixed examples of prototype signal preprocessing before and after bandpass filtering and continuous acceleration-assisted processing. Segments are 20 s long; Section S20.2 specifies parameters. Each estimator and framework receive identical processed signals. Zero-phase filtering and whole-block normalization are retrospective. No clean PPG reference is available, so waveform appearance cannot establish physiological recovery. (a)–(d) Static, walking, running, and recovery rest. Rows show raw PPG, processed PPG, and acceleration.

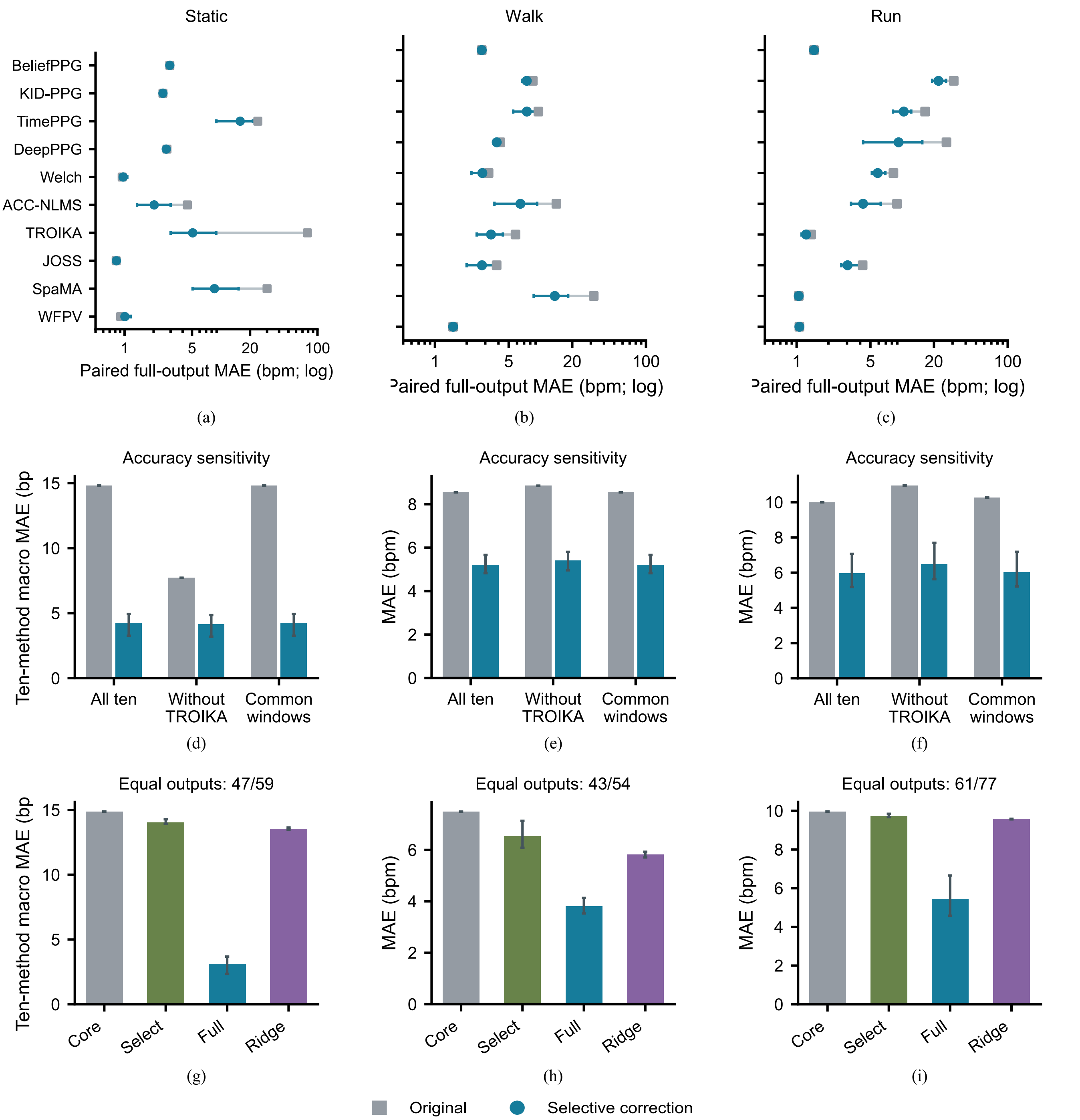

Fig. S19. State-specific gains for static, walking, and running recordings. (a)–(c) Estimator-specific paired all-output MAE on logarithmic horizontal axes. (d)–(f) Macro-averages for all ten estimators, excluding TROIKA, and strictly common windows. (g)–(i) MAE at 80% target coverage for original-risk selection, same-risk rejection, the framework, and ridge rejection. Accepted counts are 47/59, 43/54, and 61/77 reference-eligible windows, respectively. Ranking is performed separately by state, unlike pooled four-state ranking. Bars and points are configuration means, with configuration-range error bars. n = 1.

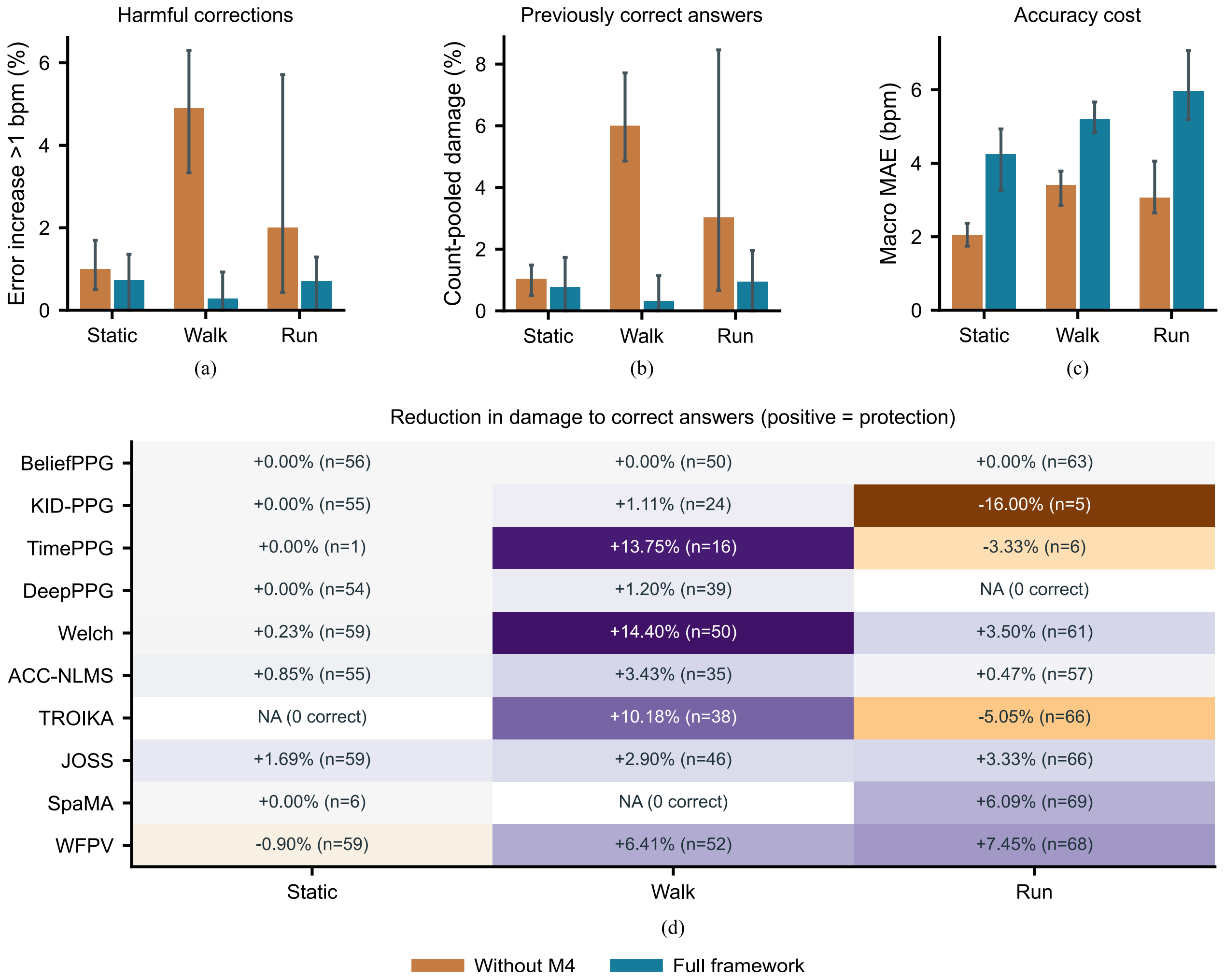

Fig. S20. Protection and accuracy costs of M4 in static, walking, and running recordings. (a) Harmful correction rate. (b) Deterioration among initially accurate estimates, calculated from pooled method–window counts within each configuration. (c) Macro-averaged MAE. (d) Deterioration rate without M4 minus the full-framework rate; positive differences indicate protection and negative differences indicate worsening. Initially accurate means original absolute error ≤5 bpm; deterioration means an absolute-error increase >1 bpm. n denotes initially accurate windows per method; NA denotes an undefined rate when none exist. Means and ranges use 15 frozen source configurations, not population confidence intervals. Percent symbols in (d) denote absolute percentage-point differences, not relative percentage reductions.

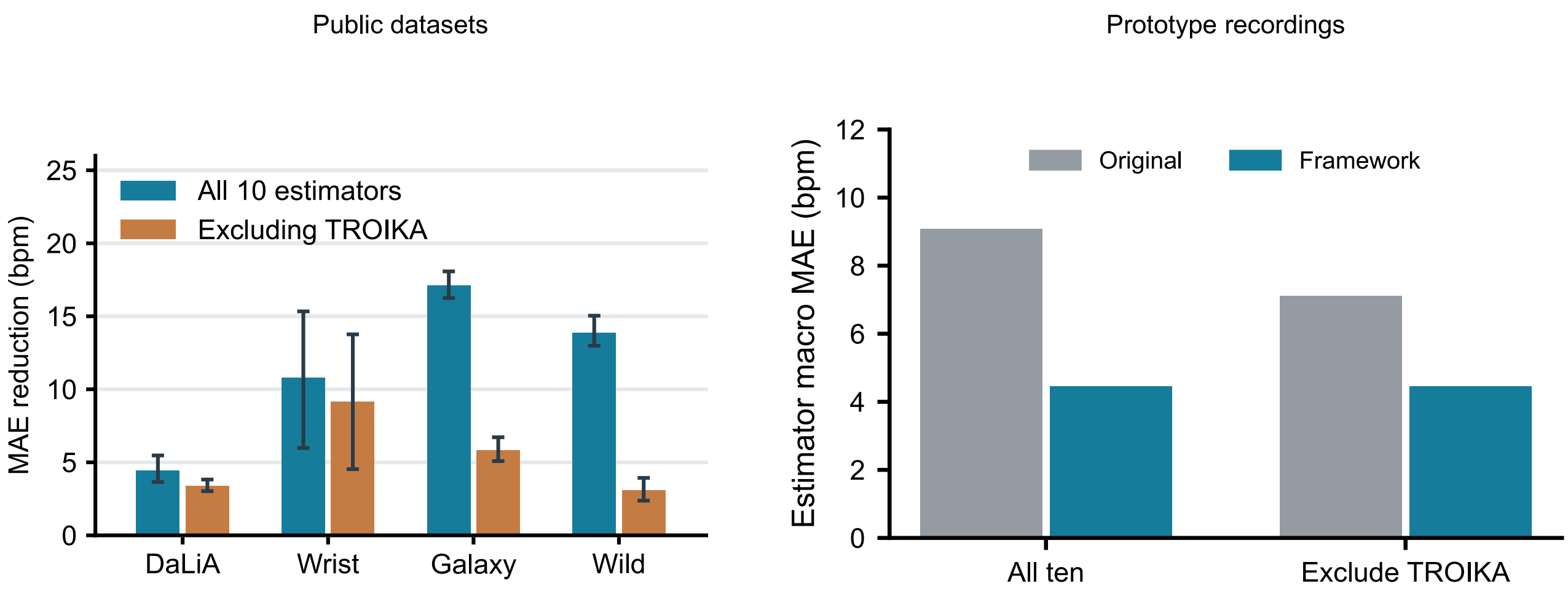

(a) (b)

Fig. S21. Sensitivity to estimator composition. (a) Macro-averaged MAE reductions on four public datasets for all ten estimators and after excluding local TROIKA, with participant-bootstrap 95% confidence intervals. (b) Original and framework macro-averaged MAE in prototype infrared recordings for all ten estimators and without TROIKA. Bars average frozen source configurations; n = 1. Both comparisons retain estimator-specific paired windows without additional rejection. All ten implementations remain in the primary analysis. Section S21 describes the motivation, implementation differences, and interpretation limits.